\pdfoutput=1
\documentclass[11pt]{article}

\usepackage{acl}

\usepackage{times}
\usepackage{tabu}
\usepackage{latexsym}

\usepackage[T1]{fontenc}
\usepackage[utf8]{inputenc}

\usepackage{microtype}

\usepackage{xspace,mfirstuc,tabulary}
\usepackage{booktabs}
\usepackage{amssymb}
\usepackage{amsmath}
\usepackage{multirow,booktabs, hhline}
\usepackage[ruled,noend]{algorithm2e}
\usepackage{amsmath, bm}
\newcommand\ourmodel{ELLE\xspace}
\usepackage{graphicx}
\usepackage{color}
\usepackage{bbm}
\usepackage{bbding}

\usepackage{cleveref}
\crefname{section}{§}{§§}
\Crefname{section}{§}{§§}

\title{\ourmodel: Efficient Lifelong Pre-training for Emerging Data}

\author{
 Yujia~Qin$^{1,2,3}$\thanks{\ \ Indicates equal contribution.}\hspace{0.3em}, Jiajie~Zhang$^{1,2,3*}$, Yankai~Lin$^{4}$, Zhiyuan~Liu$^{1,2,3,5,6}$\thanks{\ \  Corresponding author.}, Peng~Li$^7$\thanks{\ \ Part of the work was done while Peng Li was working at Tencent.}, \\ \textbf{Maosong~Sun$^{1,2,3,5,6,8\dag}$, Jie~Zhou$^4$} \\
 $^1$Department of Computer Science and Technology, Tsinghua University, Beijing, China \\
 $^2$Beijing National Research Center for Information Science and Technology \\
 $^3$Institute for Artificial Intelligence, Tsinghua University, Beijing, China\\
 $^4$Pattern Recognition Center, WeChat AI, Tencent Inc. \\
 $^5$International Innovation Center of Tsinghua University, Shanghai, China \\
 $^6$Beijing Academy of Artificial Intelligence \\
 $^7$Institute for AI Industry Research (AIR), Tsinghua University, China. \\
 $^8$Jiangsu Collaborative Innovation Center for Language Ability, Xuzhou, China \\
\texttt{\{qyj20,jiajie-z19\}@mails.tsinghua.edu.cn}\\
}

\begin{document}
\maketitle

\begin{abstract}
Current pre-trained language models (PLM) are typically trained with static data, ignoring that in real-world scenarios, streaming data of various sources may continuously grow. This requires PLMs to integrate the information from all the sources in a lifelong manner. Although this goal could be achieved by exhaustive pre-training on all the existing data, such a process is known to be computationally expensive. To this end, we propose \ourmodel, aiming at efficient lifelong pre-training for emerging data. Specifically, \ourmodel consists of (1) function preserved model expansion, which flexibly expands an existing PLM's width and depth to improve the efficiency of knowledge acquisition; and (2) pre-trained domain prompts, which disentangle the versatile knowledge learned during pre-training and stimulate the proper knowledge for downstream tasks. We experiment \ourmodel with streaming data from $5$ domains on BERT and GPT. The results show the superiority of \ourmodel over various lifelong learning baselines in both pre-training efficiency and downstream performances. The codes are publicly available at \url{https://github.com/thunlp/ELLE}.

% In the era of information explosion, streaming data of various sources may continuously grow. To integrate the information from all the sources, people tend to instantiate a new larger PLM, then pre-train it from scratch on all the data. However, this ignores the reusability of an existing well-trained PLM on old data. To this end, we propose \ourmodel, aiming to efficiently pre-train larger PLMs for growing data in a lifelong manner. Specifically, \ourmodel consists of (1) compound model expansion, which flexibly expands an existing PLM's depth and width, and efficiently inherits its knowledge; (2) dynamic memory replay, which efficiently mitigate the catastrophic forgetting on old knowledge when learning on new data; and (3) pre-trained domain prompts, which is implanted during pre-training and utilized to stimulate the proper knowledge learned during pre-training for each downstream task. We experiment with streaming data from $5$ domains, and continually expand two typical PLMs (BERT and GPT) for $4$ times. The results show the superiority of \ourmodel over various baselines in both pre-training efficiency and downstream performances. We also analyze the effectiveness of each component in \ourmodel. Besides, we find that the expanded PLM by \ourmodel exhibits more similar functionality to the original PLM than baselines. All the data, model parameters and codes used will be available upon publication.
\end{abstract}
\section{Introduction}
%  By conducting large-scale self-supervised pre-training on the large-scale raw corpus, PLMs acquire versatile structural and semantic knowledge to boost the performance in the downstream tasks. 
% \begin{figure}[!t]
% \centering
% \includegraphics[width=0.45\textwidth]{figs/motivation.pdf}
% \caption{An illustration of our efficient lifelong pre-training setup. Specifically, given limited computational resources, each time when a newly collected large corpus is available, we continually pre-train an existing PLM on the union of old data conserved in the memory and the new data to integrate all the information.}
% \label{fig:motivation}
% \end{figure}
% In the setting of limited computational costs, the priority for PLMs is to review the ``most forgettable knowledge''. However, the forgetting levels may vary a lot across different kinds of old knowledge. Worse still, such forgetting levels may also change over time. In this regard, one should judge the forgetting levels dynamically and allocate the computational resources for reviewing different old knowledge accordingly;
% (1) to ensure PLMs focus more on the ``forgettable knowledge'', we propose \textbf{dynamic memory replay} mechanism, which constantly adjusts the proportion of data used for reviewing different old knowledge;
Pre-trained language models (PLM) have broken the glass ceiling for various natural language processing (NLP) tasks~\citep{radford2018improving,devlin2018bert,Han2021PreTrainedMP}. However, most of the existing PLMs are typically trained with a static snapshot of the web information, ignoring that in real-world scenarios, streaming data from various sources may continuously grow, e.g., the gatherings of literary works~\citep{zhu2015aligning}, news articles~\citep{zellers2019defending} and science papers~\citep{lo-etal-2020-s2orc}. In addition, the distribution of incoming data may also vary over time. This requires PLMs to continually integrate the information from all the sources to grasp the versatile structural and semantic knowledge comprehensively, so that PLMs could utilize the proper knowledge to boost the performance in various downstream tasks. 

A simple yet effective way to integrate all the information is to pre-train PLMs on all the existing data exhaustively. However, such a process is computationally expensive~\citep{schwartz2019green}, especially under the information explosion era when tremendous data is continually collected. This leaves us an important question: with limited computational resources, how can we efficiently adapt PLMs in a lifelong manner? We formulate it as the \textit{efficient lifelong pre-training} problem. Similar to conventional lifelong learning, PLMs are expected to continually absorb knowledge from emerging data, and in the meantime, mitigate the catastrophic forgetting~\citep{mccloskey1989catastrophic} of previously learned knowledge.

% Currently, most of existing PLMs are typically trained with a static snapshot of the web information, ignoring that in real-world scenarios, streaming data of various sources may continuously grow, e.g., the gatherings of literary works~\citep{zhu2015aligning}, news articles~\citep{zellers2019defending} and science papers~\citep{lo-etal-2020-s2orc}, and the distributions of incoming data may also vary over time. When the newly collected data accumulates to a certain magnitude, to continual integrate the information from all the sources, the common practice is to instantiate a new larger PLM and then pre-train it with both newly collected data and old data from scratch, due to the superior sample-efficiency of large PLMs over their smaller counterparts~\citep{kaplan2020scaling}, which is also demonstrated in our experiments. However, ignoring the availability of existing trained PLMs with old data inevitably leads to environmental concerns on the prohibitive computational costs. This leaves us an important question: with limited computational resources, how can we leverage the existing PLMs to train larger ones efficiently to both absorb fresh knowledge and organize the old knowledge in a lifelong manner? We formulate it as the \textit{efficient lifelong pre-training} problem, which is illustrated in Figure \ref{fig:motivation}.
In addition, efficient lifelong pre-training poses two new challenges: (1) \textbf{efficient knowledge growth}. When the overall data scale accumulates to a certain magnitude, packing more knowledge into a fixed-sized PLM becomes increasingly hard, which significantly impacts the efficiency of PLM's knowledge growth. This is because larger PLMs show superior sample efficiency and training efficiency over their smaller counterparts~\citep{kaplan2020scaling,li2020train} due to overparameterization~\citep{arora2018optimization}. That is, larger PLMs learn knowledge in a more efficient way. Therefore, timely model expansions are essential for efficient knowledge growth;
% To minimize the cost of mastering old knowledge, the expanded larger PLM should efficiently inherit the knowledge from its ``ancestor'' instead of learning it from scratch;
(2) \textbf{proper knowledge stimulation}. During pre-training, various knowledge from all domains is packed into PLMs hastily. However, a certain downstream task may largely require the knowledge from a specific domain. Thus it is essential for PLMs to disentangle different kinds of knowledge and properly stimulate the needed knowledge for each task.

In this paper, we propose \ourmodel, targeting at \textbf{E}fficient \textbf{L}ife\textbf{L}ong pre-training for \textbf{E}merging data. Specifically, (1) to facilitate the efficiency of knowledge growth, we propose the \textbf{function preserved model expansion} to flexibly expand an existing PLM's width and depth. In this way, we increase PLM's model size and thus improve its training efficiency. Before being adapted to a new domain, the expanded PLM performs a function recovering warmup to regain the functionality of the original PLM; (2) for proper knowledge stimulation, we pre-implant \textbf{domain prompts} during pre-training to prime the PLM which kind of knowledge it is learning. Therefore, versatile knowledge from multiple sources can be disentangled. During downstream fine-tuning, we could further utilize these implanted prompts and manipulate the PLM to stimulate the proper knowledge for a specific task.

To demonstrate the effectiveness of \ourmodel, we simulate the scenario where streaming data from $5$ domains sequentially comes. We pre-train two typical PLMs (BERT and GPT) and expand their model sizes each time when the new data is available. We experiment when the number of parameters is sequentially grown from both $30$M to $125$M and $125$M to $355$M. The experimental results show the superiority of \ourmodel over multiple lifelong learning baselines in both pre-training efficiency and downstream task performances. In addition, we conduct sufficient experiments to verify the effectiveness of each component of \ourmodel. In general, we provide a promising research direction and hope this work could inspire more future attempts towards efficient lifelong pre-training.

% We also find that the expanded PLM by \ourmodel exhibits similar functionality to the original PLM, demonstrating that it successfully inherits the knowledge from its ``ancestor''.
\section{Related Work}
\paragraph{Lifelong Learning for PLMs.}
Lifelong learning aims at incrementally acquiring new knowledge, and in the meantime, mitigating the catastrophic forgetting issue. Numerous efforts have been spent towards this goal, including (1) memory-based methods~\citep{rebuffi2017icarl,rolnick2018experience}, which perform experience replay with authentic data~\citep{NEURIPS2019_f8d2e80c}, automatically generated data~\citep{sun2019lamol}, or previously computed gradients~\citep{lopez2017gradient} conserved in the memory, (2) consolidation-based methods~\citep{kirkpatrick2017overcoming,Aljundi_2018_ECCV}, which introduce additional regularization terms to consolidate the model parameters that are important to previous tasks, and (3) dynamic architecture methods~\citep{rusu2016progressive,yoon2017lifelong}, which fix trained network architectures in old tasks and dynamically grow branches for new tasks. Lifelong learning is also a hot topic for PLMs. Some target at domain adaptation through continual pre-training~\citep{gururangan2020don}, parameter-efficient adapters~\citep{he-etal-2021-effectiveness} and sparse expert models~\citep{gururangan2021demix}. Others focus on the incremental acquisition of factual knowledge that changes over time~\citep{dhingra2021time,jang2021towards}. However, the existing works seldom consider our lifelong learning setting where streaming data from multiple sources is sequentially gathered. Recently, researchers have also conducted a series of empirical studies on the continual learning of PLMs~\citep{wu2021pretrained,jin2021lifelong}.

% \citet{jin2021lifelong} conduct empirical studies on conventional continual learning algorithms for PLM adaptation. However, they do not focus on PLM's training efficiency, which is different from our setting. % More detailed comparisons are left in \cref{sec:comparison}.

\paragraph{Efficient Pre-training in NLP.}
Many attempts have been made towards improving the efficiency of pre-training, such as designing novel pre-training tasks~\citep{clark2020electra}, model architectures~\citep{zhang2020accelerating}, optimization algorithms~\citep{you2019large} and parallel architectures~\citep{shoeybi2019megatron,shazeer2018mesh}. Until recently, researchers propose to ``back distill'' the knowledge from existing PLMs to accelerate large PLMs' pre-training~\citep{qin2021knowledge}. Another line of work proposes \textit{progressive training} to dynamically expand an existing PLM's size through parameter recycling~\citep{gong2019efficient,gu-etal-2021-transformer,chen2021bert2bert}. However, these methods typically focus on training PLMs on one static corpus, and thus cannot be directly applied to our lifelong pre-training setting.
% In this section, we first provide some preliminaries for PLMs and the task formulation in \cref{sec:prelim}. Then we present our methods for \textbf{L}if\textbf{E}long \textbf{P}rE-\textbf{T}r\textbf{A}ining with emerging data in \cref{sec:lepta}, which is dubbed as \ourmodel.
\begin{figure*}[!t]
\centering
\includegraphics[width=1\textwidth]{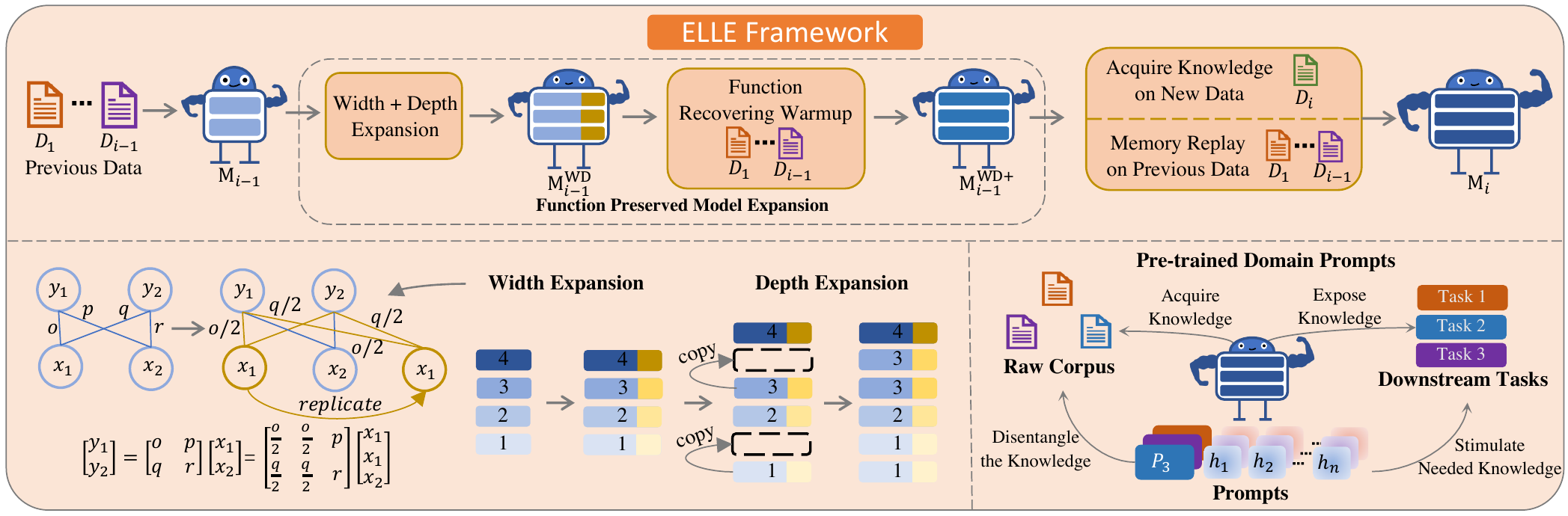}
\caption{Illustration of \ourmodel when adapting an existing PLM $\mathcal{M}_{i-1}$ trained on previous data $\overline{\mathcal{D}}_{i-1}$ to a new corpus $\mathcal{D}_i$. We also visualize the mechanism of width / depth expansion and pre-trained domain prompts.}
\label{fig:method}
\end{figure*}

\section{Methodology}
\subsection{Preliminaries}
% \label{sec:prelim}
\paragraph{Background for PLM.}
A PLM $\mathcal{M}$ generally consists of an embedding layer and $L$ Transformer~\citep{vaswani2017attention} layers. Given an input $\mathbf{x}$ consisting of a series of tokens, i.e., $\mathbf{x} = \{w_1, \dots, w_{|\mathbf{x}|}\}$, $\mathcal{M}$ first converts the input into embeddings $\{\mathbf{h}_1^0, \dots, \mathbf{h}_{|\mathbf{x}|}^0\}$, which are sequentially processed by each Transformer layer into contextualized hidden representations $\mathbf{H}^l = \{\mathbf{h}_1^l, \dots, \mathbf{h}_{|\mathbf{x}|}^l\}$, where $1 \! \le \! l \! \le \! L$. % The final representations $\mathbf{H}^L$ are used to calculate a self-supervised pre-training objective, e.g., masked language modeling~\citep{devlin2018bert}.% A classifier is further applied upon the final representation $\mathbf{H}^L$, resulting in a probability distribution for each token. Then the probability is compared with the label $\mathbf{y}$ to calculate the self-supervised objective. % $\mathcal{M}$ is then pre-trained with a self-supervised objective $\mathcal{L} = \mathcal{H}(\mathcal{P}(\mathbf{x}), \mathbf{y})$, where $\mathcal{H}$ denotes the loss function, e.g., cross-entropy for MLM~\citep{devlin2018bert} to acquire knowledge on the corpora.

\paragraph{Task Definition.}
% \label{sec:task_formulation}
Assume a stream of corpus $\overline{\mathcal{D}}_N$ from $N$ domains (e.g., news articles, web content and literary works) is sequentially gathered, i.e., $\overline{\mathcal{D}}_N = \{\mathcal{D}_1, \dots, \mathcal{D}_N\}$, where $\mathcal{D}_i = \{\mathbf{x}_i^j\}_{j=1}^{|\mathcal{D}_i|}$. The whole training process can be partitioned into several stages. Initially, we have a PLM $\mathcal{M}_1$, which has been well trained on $\mathcal{D}_1$, and for the $i$-th stage ($i > 1$), we obtain a new collection of data $\mathcal{D}_i$. Assume in this stage, we only have limited computational resources $\mathcal{R}_{i}$, our goal is to continually pre-train the existing PLM $\mathcal{M}_{i-1}$ to learn new knowledge on $\mathcal{D}_i$, and obtain a new PLM $\mathcal{M}_{i}$. Meanwhile, we expect the adapted PLM $\mathcal{M}_{i}$ should not forget the previously learned knowledge of $\overline{\mathcal{D}}_{i-1}$.

\paragraph{Overall Framework.}
As illustrated in Figure~\ref{fig:method}, starting from $\mathcal{M}_{i-1}$, which is trained on previous data $\overline{\mathcal{D}}_{i-1}$, we first expand $\mathcal{M}_{i-1}$'s width and depth and construct an enlarged PLM $\mathcal{M}_{i-1}^{\text{WD}}$ to improve its training efficiency. Then we perform function recovering warmup and train $\mathcal{M}_{i-1}^{\text{WD}}$ to inherit the knowledge of $\mathcal{M}_{i-1}$ to obtain $\mathcal{M}_{i-1}^{\text{WD+}}$. The above procedures are dubbed as \textbf{function preserved model expansion} (\cref{sec:expand}). After that, we continually pre-train $\mathcal{M}_{i-1}^{\text{WD+}}$ to gain new knowledge on $\mathcal{D}_i$. To mitigate the catastrophic forgetting on the previously learned knowledge, we employ data-based memory replay on a subset of previously gathered data $\overline{\mathcal{D}}_{i-1}^{sub} = \{\mathcal{D}_1^{sub}, \dots, \mathcal{D}_{i-1}^{sub}\}$ conserved in the memory, where $\mathcal{D}^{sub}_k = \{x_k^1, \dots, x_k^B\} \in \mathcal{D}_k$ ($1 \le k \le i-1$) and $B$ is the constrained memory size for each domain. To help PLMs disentangle the knowledge during pre-training and also stimulate the needed knowledge for each downstream task, we implant \textbf{domain prompts} into PLMs during the whole training process (\cref{sec:prompt}).

% As illustrated in Figure~\ref{fig:method}, \ourmodel consists of: \textbf{compound model expansion} (\cref{sec:expand}), \textbf{dynamic memory replay} (\cref{sec:replay}) and \textbf{pre-trained domain prompts} (\cref{sec:prompt}). In the following sub-sections, we introduce these components in detail.

\subsection{Function Preserved Model Expansion}
\label{sec:expand}
To accumulate knowledge more efficiently, each time when a new corpus $\mathcal{D}_i$ comes, we expand both $\mathcal{M}_{i-1}$'s width and depth to attain the superior sample efficiency and fast convergence brought by larger model capacity~\citep{li2020train}.

\paragraph{Width Expansion.} For width expansion, we borrow the function preserving initialization (FPI) from \citet{chen2021bert2bert}. For a brief introduction, FPI expands the matrices of all modules of a Transformer layer to arbitrary larger sizes and constructs an enlarged PLM $\mathcal{M}_{i-1}^\text{W}$. $\mathcal{M}_{i-1}^\text{W}$ is initialized using the corresponding matrices of the original $\mathcal{M}_{i-1}$ through parameter replication. For example, as visualized in Figure~\ref{fig:method}, the core principle of FPI is to divide the product of $o \times x_1$ into multiple partitions, e.g. $\frac{o}{2} \times x_1 + \frac{o}{2} \times x_1$. Formally, FPI expands a matrix $\bm{W} \in \mathbb{R}^{h_1 \times h_2}$ of $\mathcal{M}_{i-1}$ to an enlarged matrix $\bm{W}' \in \mathbb{R}^{(h_1+\Delta_{h_1}) \times h_2}$ of $\mathcal{M}_{i-1}^\text{W}$ as follows:
\vspace{-0.6em}
\begin{equation}
\small
\begin{aligned}
m(i) &=
\begin{cases}
i& i\in [1, h_1] \\
U(\{1, \dots, h_1\}) & i\in (h_1, h_1+\Delta_{h_1}],
\end{cases} \\
C_{i} &= \sum_{i'=1}^{h_1+\Delta_{h_1}}\mathbbm{I}({m(i')= m(i)}), \\
\bm{W}'_{(i, *)} &= \frac{1}{C_{i}} \cdot \bm{W}_{(m(i), *)} + \mathbbm{I}(C_{i} > 1) \cdot \bm{\delta}_i,
\end{aligned}
\end{equation}
where $U(\cdot)$ denotes a uniform sampling function, $m(\cdot)$ denotes the mapping function between two matrices, $\mathbbm{I}(\cdot)$ is an indicator function, $C_i$ counts how many partitions a specific neuron is splitted and $\bm{\delta}_i \in \mathbb{R}^{h_2}$ is a random gaussian noise. FPI ensures that both $\mathcal{M}_{i-1}^\text{W}$ and $\mathcal{M}_{i-1}$ have approximately the same functionality, i.e., both models have almost the same output given the same input. Besides function preservation, the initialized model could serve as a good starting point for further optimization. We refer readers to \citet{chen2021bert2bert} for more details about width expansion. Different from \citet{chen2021bert2bert}, we additionally introduce random noises $\bm{\delta}_i$ into the newly copied parameters of $\bm{W}'$ during initialization. These slight noises would break the symmetry after the replication and accelerate later pre-training.

\paragraph{Depth Expansion.} For depth expansion, previous works generally resort to stacking all the original PLM layers into $2\times$ layers through parameter replication~\citep{gong2019efficient}. Such initialization is demonstrated to improve training efficiency.

% i.e., for a $L$-layer PLM $\mathcal{M}_{i-1}$, they construct a $2L$-layer PLM $\mathcal{M}_{i-1}^\text{H}$ where the $i$-th and the ($i+L$)-th layer of $\mathcal{M}_{i-1}^\text{H}$ have exactly the same parameters with the $i$-th layer of $\mathcal{M}_{i-1}$ ($1 \le i \le L$).
However, the above \textit{layer stacking} method restricts the number of layers of the enlarged PLM $\mathcal{M}_{i-1}^\text{D}$ to be integer multiples of that of the original PLM $\mathcal{M}_{i-1}$, which is not flexible for practical uses. To improve the expansion flexibility so that $\mathcal{M}_{i-1}$ could be expanded with arbitrary number of layers, we propose a novel \textit{layer insertion} method to construct a new PLM $\mathcal{M}_{i-1}^\text{D}$ with $L+L'$ layers, where $1 \le L' \le L$. Specifically, we randomly select $L'$ layers from $\mathcal{M}_{i-1}$, copy each layer's parameters and insert the replication layer right before / after the original layer. We found empirically that inserting the copied layer into other positions would cause a performance drop, and the reason is that it will violate the processing order of the original layer sequence and break the PLM's original functionality. At each expansion stage when new data comes, since different layers have different functionalities, we always choose those layers that have not been copied before to help PLMs develop in an all-around way, instead of just developing a certain kind of functionality. Since both width expansion and depth expansion are compatible with each other, we simultaneously expand both of them to construct an enlarged model $\mathcal{M}_{i-1}^{\text{WD}}$, which inherits $\mathcal{M}_{i-1}$'s knowledge contained in the parameters.

\paragraph{Function Recovering Warmup.}
\label{sec:warmup}
Since the above model expansion cannot ensure exact function preservation and inevitably results in functionality loss and performance drops, we pre-train the initialized PLM $\mathcal{M}_{i-1}^{\text{WD}}$ on the previous corpora $\overline{\mathcal{D}}_{i-1}^{sub}$ conserved in the memory to recover the language abilities lost during model expansion, which is dubbed as function recovering warmup (FRW). After the warmup, we obtain $\mathcal{M}_{i-1}^{\text{WD+}}$, which successfully inherits the knowledge from $\mathcal{M}_{i-1}$ and is also well-prepared for the next training stage. % To ensure that $\mathcal{M}_{i-1}^{\text{WD+}}$ could converge during the warmup and recover comparable performance than $\mathcal{M}_{i-1}$, given limited training computations, the expanded size cannot be too large.

\subsection{Pre-trained Domain Prompt}
\label{sec:prompt}
% which pre-implants domain prompts into PLMs during pre-training so that the knowledge learned from different domain could be better disentangled and exposed properly in downstream fine-tuning
% In addition, which may also affect the proper knowledge exposure during downstream fine-tuning. In remedy for this, we propose to pre-implant \textbf{soft domain prompts} during pre-training, 
% When performing downstream fine-tuning, these prompts could serve as stimulators to expose the proper knowledge learned during pre-training.
Instead of training a separate model for each domain, we expect a single compact PLM to integrate the knowledge from all the sources. When confronted with a downstream task from a specific domain, the PLM needs to expose the proper knowledge learned during pre-training. To facilitate both knowledge acquisition during pre-training and knowledge exposure during fine-tuning, we resort to prompts as domain indicators and condition the PLM's behavior on these prompts. Soft prompts have been demonstrated as excellent task indicators~\citep{qin2021exploring} and have non-trivial transferability among tasks~\citep{su2021transferability}.

Specifically, during pre-training, to disentangle the knowledge from different sources, we implant a soft prompt token into the input to prime the PLM which kind of knowledge it is learning. The prompt of domain $i$ is a tunable vector $\mathbf{p}_i$. We prepend $\mathbf{p}_i$ before the original token embeddings $\mathbf{H}^0 = \{\mathbf{h}^0_1, \dots, \mathbf{h}^0_{|\mathbf{x}|}\}$ for an input $\mathbf{x} \in \mathcal{D}_i$, resulting in the modified input $\mathbf{H}^{0*} = \{\mathbf{p}_i; \mathbf{h}^0_1, \dots, \mathbf{h}^0_{|\mathbf{x}|}\}$, which is then processed by all the Transformer layers. Each $\mathbf{p}_i$ is optimized together with other parameters of the PLM during pre-training. During fine-tuning, when applying the PLM on a similar domain of data seen before, we could leverage the trained domain prompt and prepend it before the input of downstream data. In this way, we manually manipulate the PLM to stimulate the most relevant knowledge learned during pre-training.

\section{Experiments}
\label{sec:main_exp}

\begin{table*}[!t]
\small
\centering
\begin{tabu}{l@{~~~~}r@{~~~~}c@{~~~~}r@{~~~~}r@{~~~~}r@{~~~~}r@{~~~~}r@{~~~~}r@{~~~~}r@{~~~~}r}
\toprule
% \multicolumn{2}{c}{\textbf{Evaluated After}} & \multicolumn{2}{c}{Domain 1} & \multicolumn{2}{c}{Domain 2}    & \multicolumn{2}{c}{Domain 3}   & \multicolumn{2}{c}{Domain 4}   & \multicolumn{2}{c}{Domain 5}   \\ \hline
\multicolumn{1}{c}{\textbf{Domain}}          & \multicolumn{2}{c}{\textsc{WB}}       & \multicolumn{2}{c}{\textsc{Ns}}        & \multicolumn{2}{c}{\textsc{Rev}}     & \multicolumn{2}{c}{\textsc{Bio}}        & \multicolumn{2}{c}{\textsc{CS}}         \\ \hline
\multicolumn{1}{c}{\textbf{Metrics}}         & AP             & AP+       & AP            & AP+           & AP            & AP+          & AP            & AP+          & AP            & AP+          \\ \hline
\multicolumn{11}{l}{\textit{Growing from} $\text{BERT}_\text{L6\_D384}$ \textit{to} $\text{BERT}_\text{L12\_D768}$} \\
Naive (Lower Bound)  & $7.96$            & -          & $8.03$           & $5.54$           & $13.52$          & $21.42$         & $13.86$          & $17.67$         & $9.93$           & $9.81$          \\
                                 EWC       & $7.96$            & -          & $8.09$           & $5.65$           & $13.40$          & $20.98$         & $13.92$          & $17.75$         & $9.94$           & $9.82$          \\
                                 MAS       & $7.96$            & -          & $8.08$           & $5.65$           & $13.44$          & $21.17$         & $13.87$          & $17.67$         & $9.91$           & $9.75$          \\
                                 A-GEM       & $7.96$            & -          & $8.82$           & $6.72$           & $13.31$          & $20.06$         & $14.73$          & $18.89$         & $10.56$          & $10.58$         \\
                                 ER    & $7.96$            & -          & $6.85$           & $1.59$           & $6.99$           & $4.09$          & $6.66$           & $3.62$          & $6.39$           & $3.16$          \\
                Logit-KD  & $7.96$            & -          & $7.60$           & $0.99$           & $7.19$           & $1.95$          & $7.08$           & $2.02$          & $6.92$           & $1.92$          \\
                                 PNN       & $7.96$            & -          & $6.52$           & $0.00$           & $5.29$           & $\textbf{0.00}$ & $4.84$           & $\textbf{0.00}$ & $4.76$           & $\textbf{0.00}$ \\
                \ourmodel(ours)      & $\textbf{7.92}$   & -          & $\textbf{5.62}$  & $\textbf{-0.20}$ & $\textbf{4.81}$  & $0.64$          & $\textbf{4.41}$  & $0.64$          & $\textbf{4.06}$  & $0.44$          \\ \hline
\multicolumn{11}{l}{\textit{Growing from} $\text{BERT}_\text{L12\_D768}$ \textit{to} $\text{BERT}_\text{L24\_D1024}$} \\
ER    & $4.54$            & -          & $4.33$           & $1.31$           & $4.02$           & $1.46$          & $3.73$           & $1.15$          & $3.82$           & $1.28$          \\
                \ourmodel(ours)      & $\textbf{4.52}$   & -          & $\textbf{3.89}$  & $\textbf{0.47}$  & $\textbf{3.61}$  & $\textbf{0.75}$ & $\textbf{3.66}$  & $\textbf{0.97}$ & $\textbf{3.29}$  & $\textbf{0.54}$ \\ \hline
\multicolumn{11}{l}{\textit{Growing from} $\text{GPT}_\text{L6\_D384}$ \textit{to} $\text{GPT}_\text{L12\_D768}$} \\
 Naive (Lower Bound)  & $46.54$           & -          & $52.91$          & $37.96$          & $81.28$          & $177.22$        & $94.44$          & $160.51$        & $60.64$          & $80.48$         \\
                                 MAS       & $46.54$           & -          & $53.12$          & $38.44$          & $81.23$          & $177.20$        & $93.21$         & $157.93$        & $60.62$          & $80.28$         \\
                                 ER    & $46.54$           & -          & $44.49$          & $12.42$          & $35.46$          & $21.78$         & $33.24$          & $23.38$         & $31.94$         & $19.83$         \\
                 Logit-KD  & $46.54$           & -          & $48.93$          & $5.41$           & $37.60$          & $9.97$          & $34.60$          & $11.74$         & $33.67$          & $11.19$         \\
                                 PNN       & $46.54$           & -          & $39.90$          & $\textbf{0.00}$  & $26.84$          & $\textbf{0.00}$ & $\textbf{22.19}$ & $\textbf{0.00}$ & $21.43$          & $\textbf{0.00}$ \\
                    \ourmodel(ours)      & $\textbf{46.50}$  & - & $\textbf{36.84}$ & $2.25$           & $\textbf{25.60}$ & $4.38$          & $22.29$          & $5.88$          & $\textbf{20.49}$ & $4.31$   \\ \bottomrule       
\end{tabu}
\caption{Average perplexity (AP) and average increased perplexity ($\text{AP}^+$) of PLMs trained by different lifelong learning methods with the same train wall time. PLMs are trained with streaming data from \textsc{WB}, \textsc{Ns}, \textsc{Rev}, \textsc{Bio} and \textsc{CS} domain sequentially. We evaluate the performance each time when PLMs finish training on one domain.}
\label{table:table1}
\end{table*}

\subsection{Experimental Setting}
\paragraph{Data Streams.} We simulate the scenario where streaming data from $5$ domains is gathered sequentially, i.e., the concatenation of \textsc{Wikipedia} and \textsc{BookCorpus} (\textsc{WB})~\citep{zhu2015aligning}, \textsc{News Articles} (\textsc{Ns})~\citep{zellers2019defending}, \textsc{Amazon Reviews} (\textsc{Rev})~\citep{he2016ups}, \textsc{Biomedical Papers} (\textsc{Bio})~\citep{lo-etal-2020-s2orc} and \textsc{Computer Science Papers} (\textsc{CS})~\citep{lo-etal-2020-s2orc}. For each corpus $\mathcal{D}_i$, we roughly sample $3,400$M tokens, and the quantity for each $\mathcal{D}_i$ ($1 \le i \le 5$) is comparable to the pre-training data of BERT~\citep{devlin2018bert}. In addition, considering that in practice, the expense of storage is far cheaper than the computational resources for pre-training, we maintain a relatively large memory compared with conventional lifelong learning settings by randomly sampling $200$M tokens ($\mathcal{D}_i^{sub}$) for each corpus $\mathcal{D}_i$.

\paragraph{Evaluated Models.} We mainly follow the model architectures of BERT and GPT~\citep{radford2018improving}. We use byte-level BPE vocabulary to ensure there are few unknown tokens in each corpus. We experiment with the initial PLM $\mathcal{M}_1$ of $6$ layers and hidden size of $384$ (around $30$M parameters, denoted as $\text{BERT}_\text{L6\_D384}$ / $\text{GPT}_\text{L6\_D384}$), and linearly enlarge the PLM's number of parameters for $4$ times, to the final PLM $\mathcal{M}_5$ of $12$ layers and hidden size of $768$ (around $125$M parameters, denoted as $\text{BERT}_\text{L12\_D768}$ / $\text{GPT}_\text{L12\_D768}$). We also experiment on a larger model size, i.e., growing the PLM from $\text{BERT}_\text{L12\_D768}$ ($125$M) to $\text{BERT}_\text{L24\_D1024}$ ($355$M). Details of each $\mathcal{M}_i$'s architecture are listed in \cref{sec:pretrain_hyper}. We also discuss the effect of expanded model size at each stage in \cref{sec:model_expand}. % \footnote{Being the first work towards efficient lifelong pre-training, this paper experiments on an ideal setting that the corpus size of each domain is the same, and the number of parameters is grown linearly. We encourage future work to explore the effect of the size of streaming data and optimal expanded model size.}.
% , e.g., how may the optimal expanded size change under different data sizes

\paragraph{Training Details.} We train our model for $62,500$ steps for the first corpus. For the following domain $i$ ($i > 1$), after the model expansion, we perform function recovering warmup for $5,000$ steps, then train the resulting PLM for $20,000$ steps on the new data together with memory replay. Following \citet{chaudhry2019tiny}, we jointly train PLMs on a mixture samples from both $\mathcal{D}_i$ and $\overline{\mathcal{D}}_{i-1}^{sub}$ in each batch, and the sampling ratio of $\mathcal{D}_i$ and $\overline{\mathcal{D}}_{i-1}^{sub}$ is set to $9:1$ in every batch. Adam~\citep{kingma2014adam} is chosen as the optimizer. All the experiments are conducted under the same environment of $8$ V100 GPUs with a batch size of $2,048$. More training details of pre-training are left in \cref{sec:pretrain_hyper}. We also experiment with fewer computational budgets and memory budgets in \cref{sec:budgets}, and find that within a reasonable range, both of the two factors would not significantly influence the performance of \ourmodel.

\paragraph{Evaluation Metrics.}
We deem one algorithm to be more efficient if it could achieve the same performance as other methods utilizing fewer computations. For PLM, this is equivalent to achieving better performance using the same computations since pre-training with more computations almost always results in better performance~\citep{clark2020electra}. We evaluate the PLM's performance during both pre-training and downstream fine-tuning.

Specifically, for pre-training, we propose two metrics to evaluate how PLMs perform on the learned domains following~\citet{chaudhry2018efficient}: (1) average perplexity ($\text{AP}$) and (2) average increased perplexity ($\text{AP}^+$). We record the train wall time~\citep{li2020train} during pre-training. For a model checkpoint at time step $T$ when learning the $j$-th domain, we measure the checkpoint's perplexity $\text{PPL}_{T,i}$ on the validation set of each domain $i$. Let $\text{PPL}_{i,i}^f$ be the perplexity on the $i$-th domain when the PLM finishes training on the $i$-th domain, the above metrics are calculated as follows:
\begin{equation}
\small
\begin{aligned}
  \text{AP} &= \exp{\big(\frac{1}{j}\sum\limits_{i=1}^{j}\log\text{PPL}_{T,i}\big)}, \\
  \text{AP}^+ &= \frac{1}{j-1}\sum\limits_{i=1}^{j-1}(\text{PPL}_{T,i} - \text{PPL}_{i,i}^f),
 \end{aligned}
\end{equation}
where $\text{AP}$ measures the average performance on all the seen data $\{\mathcal{D}_1, \ldots, \mathcal{D}_j\}$. Lower $\text{AP}$ indicates the PLM generally learns more knowledge from existing domains; $\text{AP}^+$ measures the influence of current data $\mathcal{D}_j$ on previous data $\overline{\mathcal{D}}_{j-1}$. Lower $\text{AP}^+$ means PLMs forget less knowledge learned before.

To evaluate PLMs' performance in downstream tasks, for each domain, we select a representative task that is relatively stable, i.e., \textsc{MNLI}~\citep{williams2017broad}, \textsc{HyperPartisan}~\citep{kiesel2019semeval}, \textsc{Helpfullness}~\citep{mcauley2015image}, \textsc{ChemProt}~\citep{kringelum2016chemprot} and \textsc{ACL-ARC}~\citep{jurgens2018measuring} for \textsc{WB}, \textsc{Ns}, \textsc{Rev}, \textsc{Bio} and \textsc{CS}, respectively. Training details for fine-tuning are left in \cref{sec:finetune_detail}.

\begin{figure}[!t]
    \centering
    \includegraphics[width=0.45\textwidth]{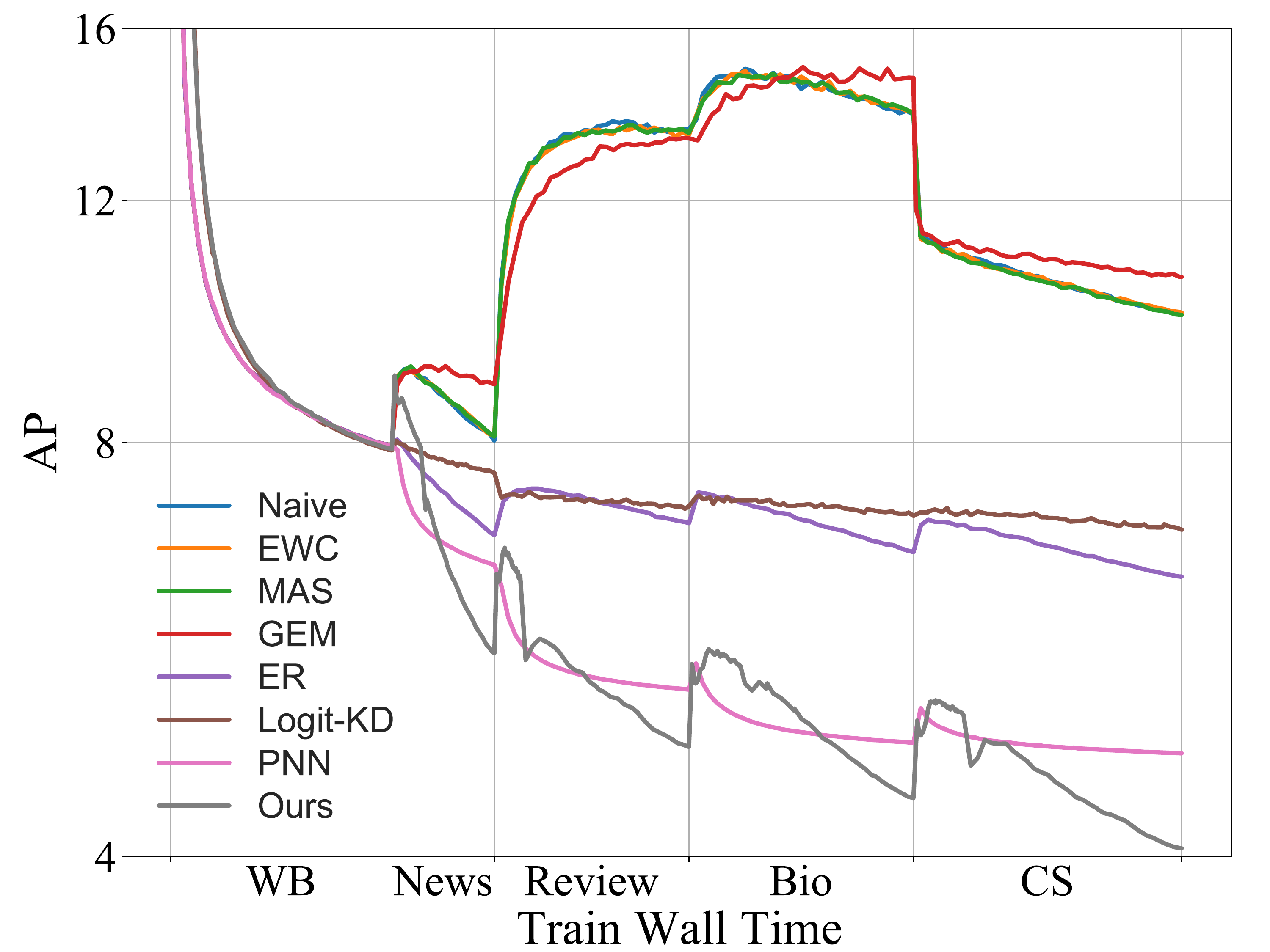}
    \caption{Average perplexity (AP) of different lifelong learning methods with $\text{BERT}_\text{L6\_D384}$ as the initial PLM. The trend curves for $\text{AP}^+$ and other PLMs are left in \cref{sec:trend_curve}.}
    \label{fig:main_exp}
\end{figure}

\paragraph{Baselines.} Keeping most of the experimental settings the same, we choose the following baselines for comparison:
(1) \textbf{Naive}, which is a naive extension of \citet{gururangan2020don} to continually adapt PLMs for each domain and can be seen as the lower bound; (2) \textbf{EWC}~\citep{schwarz2018progress}, which adopts elastic weight consolidation to add $L_2$ regularization on parameter changes; % It estimates the importance of parameters using a Fisher information matrix; % We apply Online EWC~\citep{schwarz2018progress} as baseline, a transformed and nearly effective version of EWC which simply accumulates the importance of parameters along the stream of domain corpora.
(3) \textbf{MAS}~\citep{Aljundi_2018_ECCV}, which estimates parameter importance via the gradients of the model outputs; (4) \textbf{ER}~\citep{chaudhry2019tiny}, which alleviates forgetting by jointly training models on a mixture samples from new data $\mathcal{D}_i$ and the memory $\overline{\mathcal{D}}_{i-1}^{sub}$. \ourmodel is based on ER and additionally introduces the model expansion and pre-trained domain prompts. For ER, we set the sampling ratio of $\mathcal{D}_i$ and $\overline{\mathcal{D}}_{i-1}^{sub}$ to be $9:1$ in every batch same as \ourmodel; (5) \textbf{A-GEM}~\citep{chaudhry2018efficient}, which constrains the new parameter gradients to make sure that optimization directions do not conflict with gradients on old domains;
% utilizing the gradients $g_{ref}$ calculated with a mini-batch from the memory. Averaged-GEM (A-GEM)~\citep{chaudhry2018efficient} is a more efficient version of GEM which optimizes the model with the nearest gradients $g^*$ of $g$ s.t. $g_{ref}^T g^* \ge 0$. A-GEM can ahieves the same or even better performance than the original GEM and thus we choose it as a baseline\\
(6) \textbf{Logit-KD}, which prevents forgetting by distilling knowledge from the previous model $\mathcal{M}_{i-1}$ using the old data in the memory; % According to \citet{jin2021lifelong}, logit distillation (Logit-KD)~\citep{hinton2015distilling} is the most effective distillation method for lifelong pre-training. It collects the output logits of $\mathcal{M}_i$ and $\mathcal{M}_{i-1}$, noted as $y_i$ and $y_{i-1}$ respectively. Then the pre-training loss is jointly optimized with distillation loss, i.e., the KL divergence between $y_i$ and $y_{i-1}$\\
(7) \textbf{PNN}~\citep{rusu2016progressive}, which fixes the old PLM $\mathcal{M}_{i-1}$ to completely avoid knowledge forgetting and grows new branches for learning new knowledge. For a fair comparison, we control the total train wall time of \ourmodel and all the baselines to be the same at each training stage, so that each method consumes the same computational costs.% During training of following corpora, old sub-models that have been frozen are not allowed to change but are connected to the new sub-model between corresponding layers to transfer knowledge from old corpora.

% There are also other progressive lifelong learning algorithms. However, they either can't be easily applied on BERT-structure models, such as Net2Net~\citep{chen2015net2net} which is only expanded for single-layer recurrent neural networks~\citep{sodhani2020toward}, or have obvious drawbacks, such as adapter approach~\cite{pfeiffer2021adapterfusion} where too few learnable parameters become a bottleneck of performance accordint to \citet{jin2021lifelong}.

\begin{table}[!t]
\centering
\small
\begin{tabu}{l@{~~~~}c@{~~~~}c@{~~~~}c@{~~~~}c@{~~~~}c@{~~~~}c}
\toprule
\textbf{Domain}        & \textsc{WB}        & \textsc{Ns}      & \textsc{Rev}    & \textsc{Bio}       & \textsc{CS}        & \textsc{AVG}       \\ \hline
\multicolumn{7}{l}{\textit{Growing from} $\text{BERT}_\text{L6\_D384}$ \textit{to} $\text{BERT}_\text{L12\_D768}$} \\
 Naive & $77.2$          & $72.8$          & $60.6$          & $77.1$          & $64.8$          & $70.5$         \\ 
                                EWC      & $77.4$           & $72.8$          & $61.6$          & $77.5$          & $59.6$           & $69.8$   \\
                            MAS      & $77.1$          & $73.7$          & $60.7$          & $77.5$           & $68.2$          & $71.5$          \\
                            A-GEM      & $76.6$          & $71.4$          & $61.5$          & $76.9$           & $67.5$          & $70.8$          \\
                            ER   & $77.6$         & $72.2$       & $61.9$         & $78.3$          & $63.5$          & $70.7$   \\
                            Logit-KD & $77.2$          & $69.5$          & $63.9$          & $76.8$           & $58.9$          & $69.2$          \\
                            PNN      & $76.0$          & $76.3$         & $68.0$         & $79.5$          & $65.2$          & $73.0$          \\
                    \ourmodel     & $\textbf{83.2}$ & $\textbf{81.8}$ & $\textbf{68.5}$ & $\textbf{82.9}$ & $\textbf{72.7}$ & $\textbf{77.8}$  \\ \hline
\multicolumn{7}{l}{\textit{Growing from} $\text{BERT}_\text{L12\_D768}$ \textit{to} $\text{BERT}_\text{L24\_D1024}$} \\
ER   & $84.7$          & $83.3$          & $68.0$          & $82.7$          & $71.4$          & $78.0$     \\
                    \ourmodel     & $\textbf{86.3}$ & $\textbf{90.4}$ & $\textbf{70.5}$ & $\textbf{84.2}$ & $\textbf{73.8}$ & $\textbf{81.0}$ \\ \bottomrule
\end{tabu}
\caption{Final downstream performance (F1) of BERT on each domain after finishing pre-training on all domains. Experiments of \textsc{Ns} domain are repeated for 10 times with different seeds and others are repeated for 5 times. More detailed results at different pre-training stages are illustrated in \cref{sec:finetune_detail}.}
\label{table:table2}
\end{table}

\subsection{Main Results}
Table~\ref{table:table1} summarizes the pre-training performance each time when the PLM finishes training on a specific domain. Figure~\ref{fig:main_exp} depicts the trend of $\text{AP}$ for BERT w.r.t. train wall time, other trend curves are illustrated in \cref{sec:trend_curve}. We also report the final downstream performance for discriminative PLMs ($\text{BERT}$) on each domain after finishing the whole pre-training in Table~\ref{table:table2}. The intermediate downstream performance each time when the PLM finishes training on one domain is left in \cref{sec:finetune_detail}.

\begin{table*}[!t]
\centering
\small
\begin{tabu}{ccccc|rcrrrrrrrr}
\toprule 
% \textbf{Evaluated After}   & \multicolumn{2}{c}{Domain 1} & \multicolumn{2}{c}{Domain 2}    & \multicolumn{2}{c}{Domain 3}   & \multicolumn{2}{c}{Domain 4}                & \multicolumn{2}{c}{Domain 5}                \\\hline
\multicolumn{5}{c}{\textbf{Domain}}            & \multicolumn{2}{c}{\textsc{WB}}       & \multicolumn{2}{c}{\textsc{Ns}}        & \multicolumn{2}{c}{\textsc{Rev}}     & \multicolumn{2}{c}{\textsc{Bio}}                     & \multicolumn{2}{c}{\textsc{CS}}                      \\\hline
\textbf{WE} & \textbf{DE} & \textbf{FRW} & $\delta_N$ & \textbf{PT}           & AP             & AP+       & AP            & AP+           & AP            & AP+          & AP                  & AP+                 & AP                  & AP+                 \\\hline
% \multicolumn{11}{l}{\textbf{$\text{BERT}_\text{L6\_D384}$}}                 \\
 & & & &    & $7.96$            & -          & $6.85$           & $1.59$           & $6.99$           & $4.09$          & $6.66$           & $3.62$          & $6.39$           & $3.16$          \\ \hline
\Checkmark & &\Checkmark & &       & $7.96$            & -          & $6.23$           & $0.78$           & $5.34$           & $1.42$          & $4.98$                 & $1.20$                 & $4.48$                 & $0.89$                 \\
 &\Checkmark &\Checkmark & &             & $7.96$            & -          & $5.81$           & $0.03$           & $5.49$           & $1.43$          & $5.16$                 & $1.32$                 & $4.79$                 & $0.94$                 \\
\Checkmark &\Checkmark &\Checkmark & & & $7.96$            & -          & $5.78$           & $0.02$           & $4.91$          & $0.76$          & $4.49$                & $0.73$                 & $4.13$                 & $0.52$                 \\
\Checkmark &\Checkmark & & &                 & $7.96$            & -   & $5.79$           & $0.09$           & $5.09$           & $1.13$          & $4.58$                 & $0.88$                 & $4.22$                & $0.65$                 \\
\Checkmark &\Checkmark &\Checkmark &\Checkmark &         & $7.96$            & -          & $5.69$           & $-0.13$          & $4.85$           & $0.67$          & $4.45$                 & $0.69$                 & $4.09$                 & $0.47$                 \\ \hline
\Checkmark &\Checkmark &\Checkmark &\Checkmark &\Checkmark                  & $\textbf{7.92}$   & -          & $\textbf{5.62}$  & $\textbf{-0.20}$ & $\textbf{4.81}$  & $\textbf{0.64}$ & $\textbf{4.41}$        & $\textbf{0.64}$                 & $\textbf{4.06}$                 & $\textbf{0.44}$                 \\ \bottomrule
\end{tabu}
\caption{$\text{AP}$ and $\text{AP}^+$ of different combinations of strategies when growing $\text{BERT}_\text{L6\_D384}$ to $\text{BERT}_\text{L12\_D768}$.}
\label{table:table3}
\end{table*}

\paragraph{Superiority of \ourmodel.}
(1) From the results in Table~\ref{table:table1}, we observe that, compared with all the baselines, \ourmodel achieves the lowest $\text{AP}$ and satisfying $\text{AP}^+$ after finishing training on each domain. This demonstrates that, given limited computational resources, \ourmodel could acquire more knowledge and in the meantime, mitigate the knowledge forgetting problem. (2) We also observe from Figure~\ref{fig:main_exp} that the $\text{AP}$ of \ourmodel descends the fastest, showing the superior training efficiency of \ourmodel over all baselines. (3) Besides, \ourmodel performs the best on all downstream tasks, indicating that the knowledge learned during pre-training could be properly stimulated and leveraged for each downstream task. (4) The superiority of \ourmodel is consistently observed on the larger model size, i.e., $\text{BERT}_\text{L24\_D1024}$ and other model architectures, i.e., $\text{GPT}_\text{L12\_D768}$. This shows that \ourmodel is agnostic to both the model size and the specific PLM model architecture chosen. We expect future work to apply \ourmodel to other PLM architectures and extremely large PLMs.

% To demonstrate that \ourmodel is applicable to other model structures, we experiment on GPT~\citep{radford2018improving} trained with auto-regressive language modeling, and grow it from $\text{GPT}_\text{L6\_D384}$ to $\text{GPT}_\text{L12\_D768}$. Similarly, we choose \textbf{Naive}, \textbf{MAS}, \textbf{ER}, \textbf{Logit-KD} and \textbf{PNN} as the baselines. According to the results shown in Table~\ref{table:table1}, \ourmodel still outperforms all the baselines. This shows that \ourmodel is agnostic to the specific PLM model structure chosen. We expect future work to apply \ourmodel on other PLMs.

\paragraph{Comparisons with Baselines.} (1) First of all, consolidation-based methods (EWC and MAS) perform almost comparable with the naive baseline in either pre-training or downstream tasks. This means that parameter regularization may not be beneficial for PLMs' knowledge acquisition. 
(2) Among memory-based methods, gradient-based replay (A-GEM) exhibits poorer performance in pre-training, on the contrary, data-based replay (ER and Logit-KD) achieve lower $\text{AP}$ and $\text{AP}^+$ than the naive baseline, demonstrating that replaying real data points could more efficiently mitigate the knowledge forgetting problem. Meanwhile, all of the memory-based methods perform comparable or worse than the naive baseline in downstream performance. % Compared with ER, although Logit-KD achieves lower $\text{AP}^+$, its $\text{AP}$ and downstream performance are significantly poorer, which is because it wastes much time to calculate the teacher $\mathcal{M}_{i-1}$'s logits, thus leaving relatively less time for $\mathcal{M}_{i}$ to learn new data.
(3) PNN achieves significantly lower $\text{AP}$ than non-progressive baselines, and is immune to knowledge forgetting ($\text{AP}^+\!\!=\!0$). It also performs better on the downstream tasks than other baselines. This indicates that enlarging the network is an effective way for lifelong pre-training and also benefits downstream tasks. 

% When fine-tuning on $i$-th domain, we add a classifier head on submodel $\mathcal{M}_i$ and optimize $\mathcal{M}_{1-i}$ together, because this is better than freezing previous sub-models and only tuning $\mathcal{M}_i$.  Even so, We find that despite that the algorithm is immune to forgetting and outperforms all non-progressive algorithms during pre-training, PNN performs badly on downstream tasks, especially in the last two domains. A possible reason is that the knowledge transferred from previous sub-models for pre-training is not suitable for downstream tasks. Therefore, appropriate expansion strategy that keeps the consistence of pre-training and downstream fine-tuning is important. Our ELLE framework preserves the overall BERT structure after expanding and thus can achieve lower AP during pre-training and perform better on all downstream tasks.

\section{Analysis}
\label{sec:analysis}

In this section, we conduct analyses to investigate the effect of \ourmodel's components. We follow the setting in \cref{sec:main_exp} by choosing $\text{BERT}_\text{L6\_D384}$ as the initial model and continually growing it to $\text{BERT}_\text{L12\_D768}$. Specifically, we investigate the effect of (1) width expansion (WE), (2) depth expansion (DE), (3) function recovering warmup (FRW), (4) the random noises added into the newly constructed parameters during model expansion ($\delta_N$) and (5) the pre-trained domain prompts (PT). We test \ourmodel under different combinations of the above components and compare the results. The experimental results of pre-training and downstream tasks are summarized in Table~\ref{table:table3} and Table~\ref{table:table4}, respectively. Detailed trend curves for $\text{AP}$ and $\text{AP}^+$ are illustrated in \cref{sec:trend_curve}.%  We also show in \cref{sec:attention_pattern} that the expanded PLM by \ourmodel exhibits similar functionality to the original PLM.

\paragraph{Effect of Width / Depth Expansion.} First, we compare the differences of conducting only width expansion (WE+FRW), only depth expansion (DE+FRW) and expansion on both width and depth (WE+DE+FRW) before function preserving warmup. For a fair comparison, we keep the total number of $\mathcal{M}_i$'s increased parameters for the above three strategies almost the same at each stage $i$. The specific model architectures are listed in \cref{sec:model_arch_model_expansion}. The results show that: (1) compared with the non-expanding baseline, all these three strategies achieve better pre-training and downstream performance, showing that with the growth of model size, the sample efficiency and training efficiency are extensively increased. Therefore, PLMs could gain more knowledge with limited computational resources and perform better in downstream tasks; (2) compared with expanding only width or depth, expanding both of them is more efficient and can also achieve better downstream performance on almost all domains, except the \textsc{Ns} domain. This is also aligned with previous findings that PLM's growth favors compound scaling~\citep{gu-etal-2021-transformer}. We also conclude from the trend curves in \cref{sec:trend_curve} that only expanding depth will make the training process unstable.

\begin{table}[!t]
\centering
\small
\begin{tabu}{c@{~}c@{~}c@{~}c@{~}c@{~}|c@{~~}c@{~~}c@{~~}c@{~~}c@{~~}c@{~}c}
\toprule 
\textbf{WE} & \textbf{DE} & \textbf{FRW} & $\delta_N$ & \textbf{PT}  & \textsc{WB}        & \textsc{Ns}      & \textsc{Rev}    & \textsc{Bio}       & \textsc{CS}        & \textsc{AVG}       \\ \hline
 & & & &   & $77.6$          & $72.2$          & $61.9$          & $78.3$          & $63.5$          & $70.7$           \\ \hline
\Checkmark & &\Checkmark & &        & $81.9$          & $77.5$          & $64.9$          & $80.3$          & $70.7$          & $75.1$                    \\
 &\Checkmark &\Checkmark & &       & $82.4$          & $79.9$          & $66.2$          & $80.4$          & $71.0$          & $75.9$                    \\
\Checkmark &\Checkmark &\Checkmark & &          & $83.4$          & $74.7$          & $67.4$          & $82.4$          & $72.2$ & $76.0$                    \\
\Checkmark &\Checkmark & & &       & $82.6$          & $75.7$          & $67.4$          & $82.3$          & $71.4$          & $75.9$                    \\
\Checkmark &\Checkmark &\Checkmark &\Checkmark &                     & $\textbf{83.5}$ & $77.1$           & $66.9$          & $\textbf{83.3}$ & $71.3$          & $76.4$                   \\ \hline
\Checkmark &\Checkmark &\Checkmark &\Checkmark &\Checkmark                              & $83.2$          & $\textbf{81.8}$ & $\textbf{68.5}$ & $82.9$           & $\textbf{72.7}$          & $\textbf{77.8}$  \\
% \hline
% \multicolumn{5}{l}{\ourmodel}                              & $83.2$          & $\textbf{81.8}$ & $\textbf{68.5}$ & $82.9$           & $\textbf{72.7}$          & $\textbf{77.8}$  \\
% \multicolumn{5}{l}{\ourmodel $-$ $\text{PT}_\text{fine-tune}$}    & $82.9$          & 79.9          & $67.0$          & $82.1$          & $67.7$          & $75.9$           \\
% \multicolumn{5}{l}{\ourmodel $+$ $\lnot\text{PT}_\text{fine-tune}$} &   $83.1$   & $80.6$          &   $68.1$  & $81.7$          & $70.8$          & $76.9$          \\
 \bottomrule
\end{tabu}
\caption{$\text{BERT}_\text{L12\_D768}$'s downstream performance (F1) on each domain after being continually pre-trained on all domains with different combinations of strategies.}
\label{table:table4}
\end{table}

\paragraph{Effect of Function Recovering Warmup.} We compare the performance of the model expansion w/ and w/o FRW, i.e., WE+DE and WE+DE+FRW. For a fair comparison, we keep the total train wall time for either strategy the same, in other words, for WE+DE, PLMs can be trained for more steps on the new domain due to the removal of FRW. However, the results show that WE+DE achieves worse $\text{AP}$ and $\text{AP}^+$, indicating that without FRW, PLM would learn new knowledge slower and also forget more previous knowledge. The trend curve in \cref{sec:trend_curve} also shows that AP and $\text{AP}^+$ decrease faster with FRW. This demonstrates the necessity of the warmup after model expansion, i.e., PLMs could better recover the knowledge lost during model expansion and also get prepared for learning new knowledge. Meanwhile, WE+DE+FRW performs slightly better than WE+DE in most of the downstream tasks, except the \textsc{Ns} domain. % We also found that within a reasonable range, the final performance is not sensitive to the training steps of FRW. % as long as the enlarged PLM achieves comparable PPL than the original PLM on the old domains.

\paragraph{Effect of Random Noises.} Different from the original FPI~\citep{chen2021bert2bert}, \ourmodel additionally adds random noises into the newly copied parameters after expanding the width of PLMs as mentioned in \cref{sec:expand}. By comparing the model performance w/ and w/o this trick, i.e., WE+DE+FRW and WE+DE+FRW+$\delta_N$, we can see that the added noises significantly speed up pre-training and also conduce to improving PLM's overall downstream performance. This validates our hypothesis that random noises are useful for breaking the symmetry of the copied parameters, thus providing a better initialization that further optimization favors.

\paragraph{Effect of Pre-trained Domain Prompts.} 
To investigate the effect of pre-trained domain prompts, we first compare the performance w/ and w/o them, i.e., WE+DE+FRW+$\delta_N$ and WE+DE+FRW+$\delta_N$+PT. From the results we can conclude that when aided with domain prompts, PLMs achieve lower $\text{AP}$ and $\text{AP}^+$ during pre-training, showing that domain prompts could accelerate pre-training and alleviate catastrophic forgetting by disentangling the knowledge from different sources. Furthermore, domain prompts generally improve downstream performance by stimulating the proper knowledge needed for each task.

\begin{table}[!t]
\centering
\small
\begin{tabu}{c@{~~}c@{~~}c@{~~}c@{~~}c@{~~}c@{~~}c@{~~}c}
\toprule 
 \multicolumn{1}{c}{\textbf{Domain}}  & \textsc{WB}        & \textsc{Ns}      & \textsc{Rev}    & \textsc{Bio}       & \textsc{CS}        & \textsc{AVG}       \\ \hline
\ourmodel $-$ $\text{PT}_\text{fine-tune}$    & $82.9$          & 79.9          & $67.0$          & $82.1$          & $67.7$          & $75.9$           \\
\ourmodel $+$ $\lnot\text{PT}_\text{fine-tune}$ &   $83.1$   & $80.6$          &   $68.1$  & $81.7$          & $70.8$          & $76.9$          \\
\ourmodel                              & $\textbf{83.2}$          & $\textbf{81.8}$ & $\textbf{68.5}$ & $\textbf{82.9}$           & $\textbf{72.7}$          & $\textbf{77.8}$  \\
 \bottomrule
\end{tabu}
\caption{$\text{BERT}_\text{L12\_D768}$'s downstream performance (F1) on each domain when no prompt / a wrong prompt is prepended in the input.}
\label{table:table5}
\end{table}

\begin{figure*}[!t]
    \centering
    \includegraphics[width=0.85\textwidth]{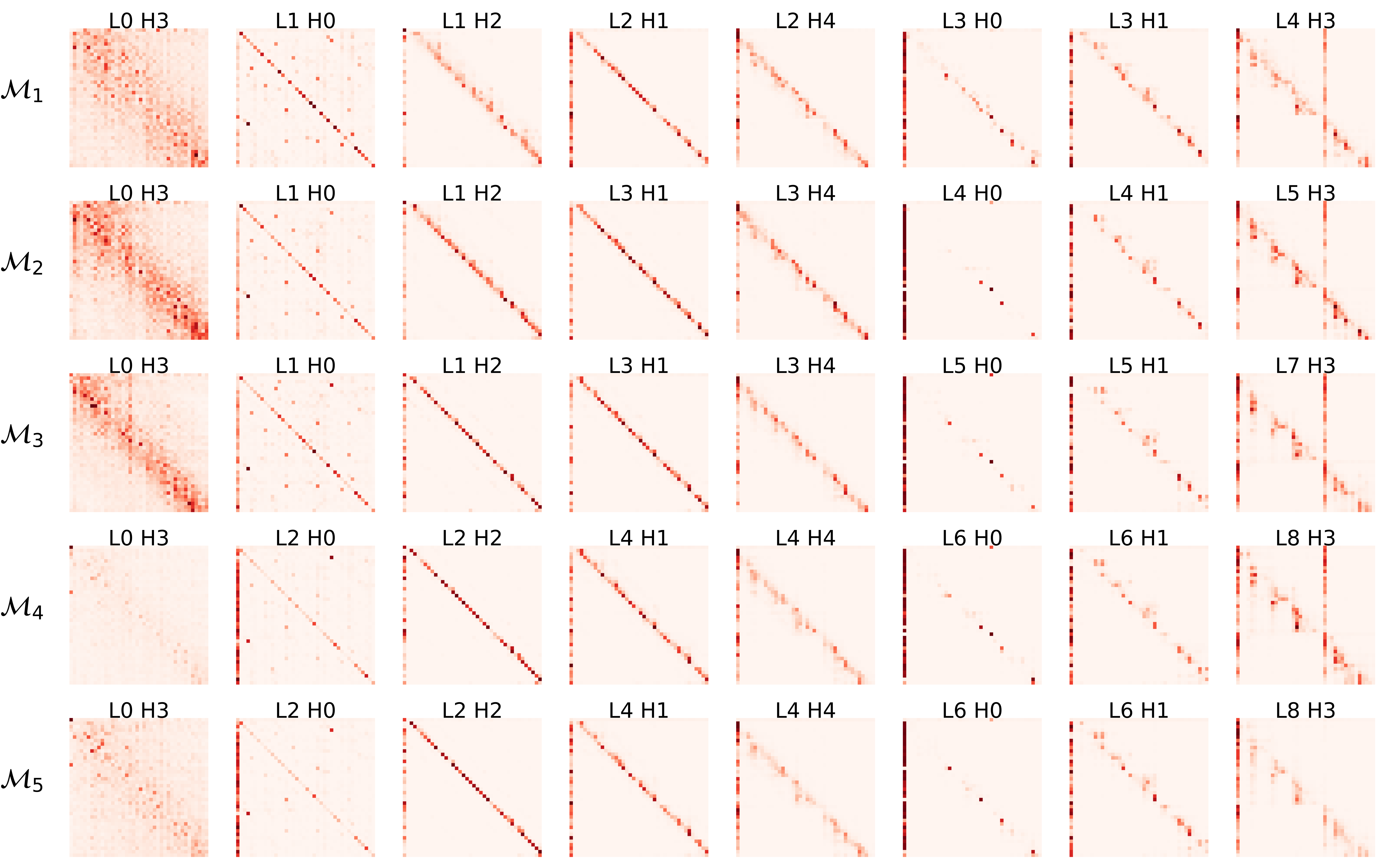}
    \caption{The visualization of the attention patterns of different attention heads in $\mathcal{M}_1$ ($\text{BERT}_\text{L6\_D384}$), $\mathcal{M}_2$ ($\text{BERT}_\text{L8\_D512}$), $\mathcal{M}_3$ ($\text{BERT}_\text{L10\_D640}$), $\mathcal{M}_4$ ($\text{BERT}_\text{L11\_D708}$) and $\mathcal{M}_5$ ($\text{BERT}_\text{L12\_D768}$) after finishing training on the new corpus $\mathcal{D}_i$. Note that in this figure, all the attention heads of a PLM $\mathcal{M}_i$ are expanded from all its ancestors $\{\mathcal{M}_{1}, \dots, \mathcal{M}_{i-1}\}$ in the same column. We observe similar attention patterns between the descendant PLM and the ancestor PLM, demonstrating the descendant PLM successfully preserves the functionality of its ancestors. }
    \label{fig:attn_similar}
\end{figure*}

To rigorously investigate how domain prompts stimulate the knowledge during fine-tuning, for a PLM pre-implanted with prompts during pre-training, we test its downstream performance when (1) no prompt is prepended in the input (i.e., \ourmodel - $\text{PT}_\text{fine-tune}$) during fine-tuning and (2) a prompt from a random wrong domain is prepended in the input (i.e., \ourmodel + $\lnot\text{PT}_\text{fine-tune}$). The results in Table~\ref{table:table5} show that both of the above strategies have lower downstream performance than prepending the right prompt (\ourmodel). We hypothesize the reasons are two-fold: (1) firstly, for \ourmodel  - $\text{PT}_\text{fine-tune}$, there exists a great gap between the formats of input during pre-training and fine-tuning, and such a gap would hinder the successful knowledge transfer; (2) secondly, for \ourmodel + $\lnot\text{PT}_\text{fine-tune}$, although the above gap disappears, the PLM is primed with a wrong domain prompt, and thus cannot properly stimulate the knowledge that is most relevant to the downstream task. Although manually deciding the most relevant domain prompt for a specific downstream task is relatively easy and fast, such a process can also be automated by training a domain discriminator, which is left as future work.
% Specifically, not inserting any prompt leads to even worse performance than inserting a wrong prompt.
% The above experiments indicate that prompts can remind PLMs of proper knowledge during downstream fine-tuning. Hence it's important to maintain the consistance of pre-training and fine-tuning. 

% Showing that these prompts could be manually inserted to help PLMs stimulate the proper knowledge needed for each downstream task.

% According our hypothesis, different pre-trained domain prompts (PT) help disentangle the knowledge from different sources and benefits lifelong pre-training.

\paragraph{Attention Pattern Visualization of a Stream of PLMs.}
\label{sec:attention_pattern}
Through the function preserved model expansion, PLMs inherit the knowledge of their ``ancestors'' contained in the parameters. Intuitively, the descendant PLM (the expanded larger PLM) should have similar functionalities to the ancestor PLM (the original PLM before model expansion). We thus investigate such functionality similarity through the lens of attention patterns of each attention head in the Transformer layer. 

Specifically, we visualize the attention patterns of a stream of PLMs ($\{\mathcal{M}_1, \ldots, \mathcal{M}_5\}$) trained by \ourmodel when growing from $\text{BERT}_\text{L6\_D384}$ to $\text{BERT}_\text{L12\_D768}$. We checkpoint each PLM $\mathcal{M}_i$ when it finishes training on the emerging data $\mathcal{D}_i$. We input the same data into these checkpoints to derive the attention patterns. The results are illustrated in Figure~\ref{fig:attn_similar}, from which we observe that the attention patterns of a head in a descendant PLM are surprisingly similar to those of its ``ancestors'', even if the descendant PLM is further trained on the new data and enlarged many times. This indicates that the expanded PLM by \ourmodel successfully inherits the knowledge from its ``ancestor'', and thus exhibits similar functionality to some extent.
\section{Conclusion}
In this paper, we present the efficient lifelong pre-training problem, which requires PLMs to continually integrate the information from emerging data efficiently. To achieve our goal, we propose \ourmodel and progressively expand PLMs to acquire knowledge efficiently and mitigate the knowledge forgetting. We also pre-implant domain prompts during pre-training and use them to stimulate the needed knowledge for downstream tasks. The experimental results show the superiority of \ourmodel over various lifelong learning baselines in both pre-training efficiency and downstream performances.

\section*{Acknowledgments}
This work is supported by the National Key R\&D Program of China (No. 2020AAA0106502), NExT++ project from the National Research Foundation, Prime Minister’s Office, Singapore under its IRC@Singapore Funding Initiative, Beijing Academy of Artificial Intelligence (BAAI), and International Innovation Center of Tsinghua University, Shanghai, China. This work is also supported by the Pattern Recognition Center, WeChat AI, Tencent Inc. Yujia Qin, Jiajie Zhang and Yankai Lin designed the methods and the experiments. Jiajie Zhang conducted the experiments. Yujia Qin, Jiajie Zhang and Yankai Lin wrote the paper. Zhiyuan Liu, Peng Li, Maosong Sun and Jie Zhou advised the project and participated in the discussion. The authors would like to thank Yichun Yin and Cheng Chen for their constructive advice.

% Zhengyan Zhang conducted the experiments. Zhengyan Zhang, Yankai Lin, Zhiyuan Liu, and Peng Li wrote the paper. Maosong Sun and Jie Zhou provided valuable advices to the research.
\bibliography{anthology,custom}
\bibliographystyle{acl_natbib}

\clearpage
\appendix
\section*{Appendices}
\label{apdx}

\section{Additional Analysis on Function Preserved Model Expansion}
\label{sec:model_expand}
In addition to the analyses of function preserved model expansion conducted in our main paper, in this section, we further analyze the effect of (1) the expanded model size at each training stage and (2) the choice of copied layer during depth expansion. We experiment on the combination of WE+DE+FRW as mentioned in \cref{sec:analysis} and choose $\text{BERT}_\text{L6\_D384}$ as the initial PLM $\mathcal{M}_1$. Other settings are kept the same as \cref{sec:analysis}.

\begin{table*}[!t]
\centering
\small
\begin{tabu}{lrcrrrr}
\toprule 
% \textbf{Evaluated After}   & \multicolumn{2}{c}{Domain 1} & \multicolumn{2}{c}{Domain 2}    & \multicolumn{2}{c}{Domain 3}   & \multicolumn{2}{c}{Domain 4}                & \multicolumn{2}{c}{Domain 5}                \\\hline
\textbf{Domain}            & \multicolumn{2}{c}{\textsc{WB}}       & \multicolumn{2}{c}{\textsc{News}}        & \multicolumn{2}{c}{\textsc{Review}}                           \\\hline
\textbf{Metrics}           & AP             & $\text{AP}^+$       & AP            & $\text{AP}^+$           & AP            & $\text{AP}^+$         \\\hline
Expand $0$ layers and heads per domain      & $\textbf{13.09}$           & -          & $8.99$           & $-0.49$          & $8.24$           & $2.80$          \\
Expand $2$ layers and heads per domain      & $\textbf{13.09}$  & - & $\textbf{8.28}$  & $\textbf{-1.44}$ & $\textbf{7.25}$  & $\textbf{1.11}$ \\
Expand $4$ layers and heads per domain      & $\textbf{13.09}$           & -          & $8.62$          & $-0.95$          & $7.53$           & $1.30$          \\
Expand $6$ layers and heads per domain      & $\textbf{13.09}$           & -          & $9.08$           & $-0.24$          & $7.92$           & $1.49$     \\ \bottomrule
\end{tabu}
\caption{AP and $\text{AP}^+$ of PLMs trained with \ourmodel that expands $0$, $2$, $4$ and $6$ layers and heads during model expansion, respectively. AP and $\text{AP}^+$ are evaluated when each PLM finishes training on each domain.}
\label{table:expand_size}
\end{table*}

\paragraph{Effect of Expanded Model Size.} In our main experiments, we assume that the data size of each emerging corpus is the same and linearly enlarge the model size when conducting model expansion. In this section, we explore the effect of expanded model size given limited computational resources. We conduct experiments on a stream of data from $3$ domains, i.e., \textsc{WB}, \textsc{Ns} and \textsc{Rev} domain. We start from the initial PLM $\text{BERT}_\text{L6\_D384}$ and continually adapt it to new corpora. Under the same training environment, we control the computational costs (train wall time) of each domain to be $7200$ seconds. We compare the performances when the PLM expands $0$, $2$, $4$, and $6$ layers and heads for each domain, respectively. Note the PLMs expanded with a larger size would be trained with fewer steps to control the train wall time.

The results are shown in Table~\ref{table:expand_size}, from which we can conclude that the best performance is obtained when the model expands $2$ layers and heads at each expansion stage, and expanding more or fewer parameters leads to a performance drop. The reasons are two-fold: (1) firstly, as mentioned before, expanding the model size improves the sample efficiency~\citep{kaplan2020scaling,li2020train}, which is beneficial for PLMs' knowledge acquisition; (2) secondly, when increasing the expanded model size, the benefits from inheriting the knowledge of a small PLM would become less and less evident. To sum up, expanding with an intermediate size strikes the best trade-off between the above two reasons, and there may exist an optimal expanded size when performing model expansion.

Intuitively, the optimal expanded model size may be influenced by many factors, e.g., the computational budgets, the amount of emerging data, the PLM's model architecture, etc. And systematically analyzing the effects of all these factors is beyond the scope of this paper, thus we expect future works to design algorithms to accurately estimate the optimal expanded size for model expansion.

\paragraph{Choice of Copied Layer.} As mentioned in \cref{sec:expand}, each time when we conduct width expansion, we choose those layers that have not been copied before. To demonstrate the benefit of this trick, we compare three expansion strategies: (1) always replicating those layers that have not been copied before (WE+DE+FRW); (2) always replicating the first layer (WE+$\text{DE}_\text{first}$+FRW) and (3) always replicating the last layer (WE+$\text{DE}_\text{last}$+FRW).

\begin{figure*}[!t]
    \centering
    \includegraphics[width=0.45\textwidth]{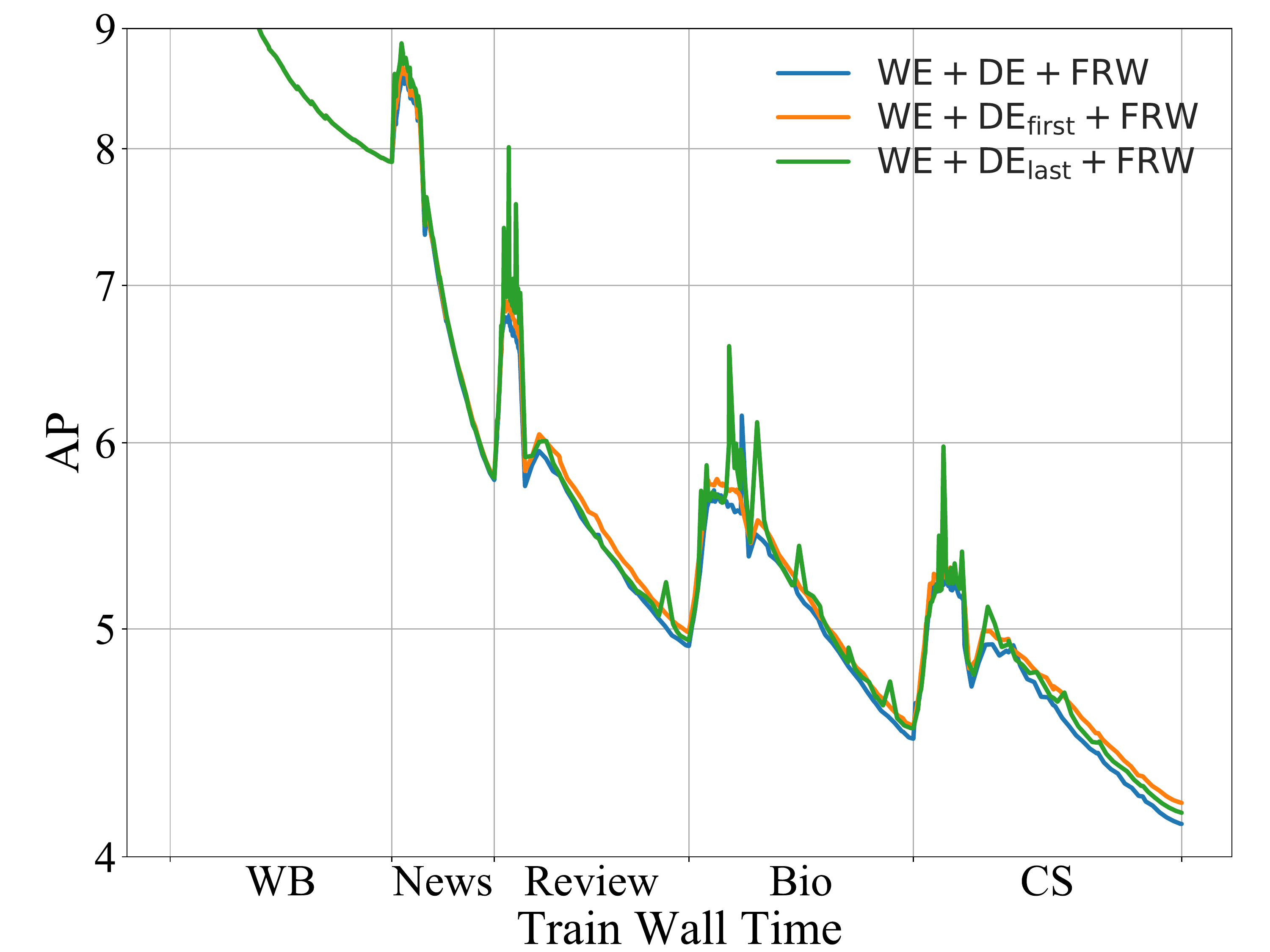}
    \includegraphics[width=0.45\textwidth]{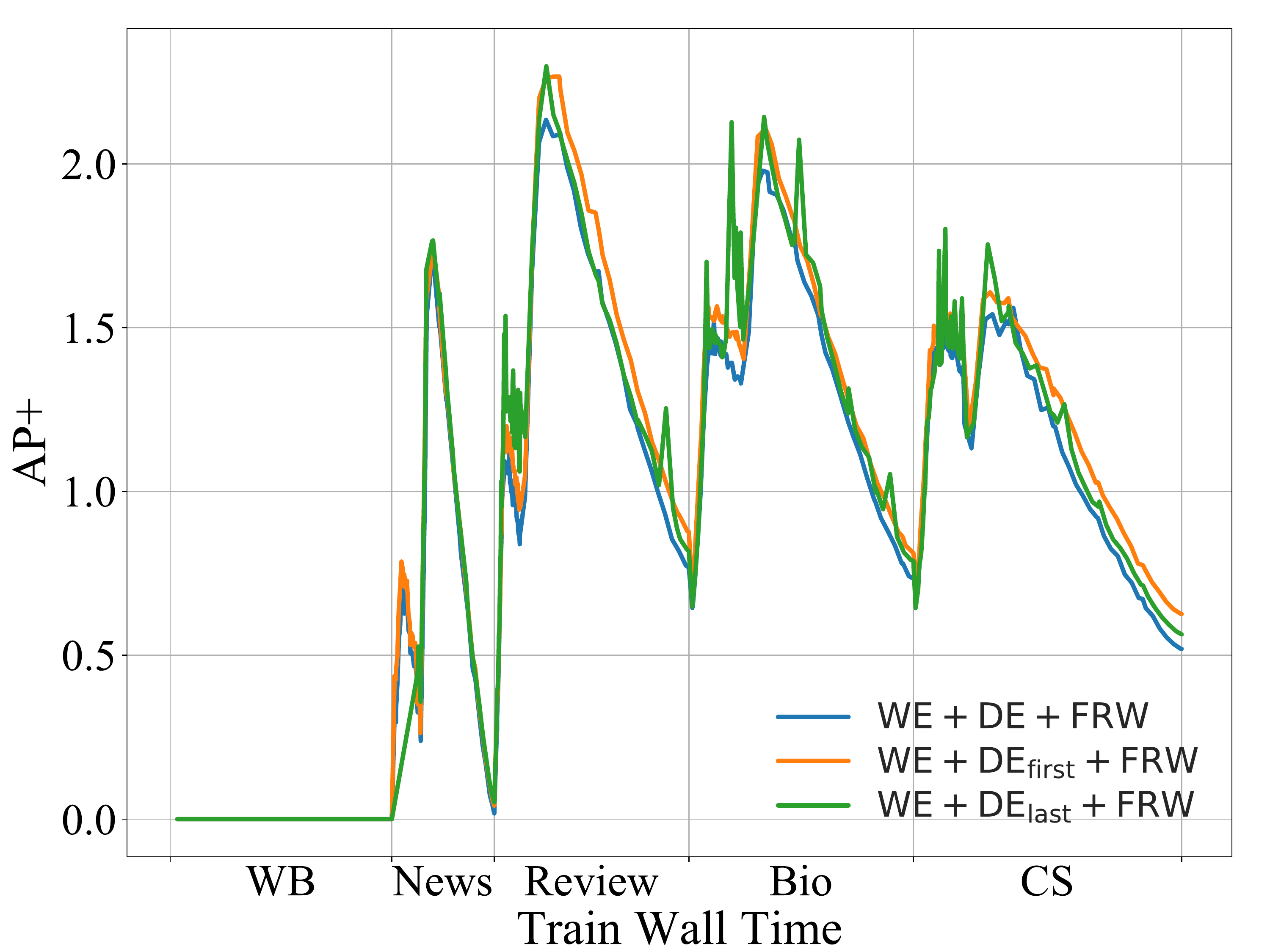}
    \caption{AP and $\text{AP}^+$ of PLMs trained by \ourmodel using different depth expansion strategies: WE+DE+FRW, WE+$\text{DE}_\text{first}$+FRW and WE+$\text{DE}_\text{last}$+FRW w.r.t train wall time.}
    \label{fig:same_layer}
\end{figure*}

The results in Figure~\ref{fig:same_layer} show that $\text{AP}$ and $\text{AP}^+$ descend the fastest when we always replicate those layers that have not been copied before (i.e., WE+DE+FRW). This demonstrates that, since different layers have different functionalities, choosing those layers that have not been expanded before would help PLMs develop in an all-around way, instead of just developing a certain kind of functionality. Furthermore, we find empirically that when pre-training PLMs continually on multiple domains, if we always choose those layers that have not been expanded before at each depth expansion stage, then the final performance is not sensitive to choosing which layers to expand first.

% The expanded size cannot be too large so that given limited training computations, the newly constructed large PLM could at least converge during the function recovering warmup and recover comparable performance than the original PLM.

% We also empirically found that given more computational resources, adding more parameters would gradually outperforms adding less parameters. 

% Too little expansion is not effecient for pre-training, while too much expansion may make optimization difficult. 

% \begin{figure*}[!t]
%     \centering
%     \includegraphics[width=0.45\textwidth]{figs/enlarge_avg_ppl.pdf}
%     \includegraphics[width=0.45\textwidth]{figs/enlarge_avg_ppl_increased.pdf}
%     \caption{Trend curves of AP and $\text{AP}^+$ w.r.t train wall time. The evaluated PLMs are trained with \ourmodel, and expand $0$, $2$, $4$ and $6$ layers and heads during model expansion.}
%     \label{fig:expand_size}
% \end{figure*}

\begin{table*}[!t]
  \centering
  \small
    \begin{tabu}{cccccccccccc}
    \toprule 
            \multicolumn{1}{c}{$\textbf{Model}$}
          & \multicolumn{1}{c}{$n_{\text{params}}$} & \multicolumn{1}{c}{$n_{\text{layers}}$} & \multicolumn{1}{c}{$d_{\text{model}}$} & \multicolumn{1}{c}{$n_{\text{heads}}$} & \multicolumn{1}{c}{$d_{\text{FFN}}$} & \multicolumn{1}{c}{\text{lr}} & SF & STF & RW & TWT(s)
          \\
    \hline
    \multicolumn{11}{l}{\textit{Growing from} $\text{BERT}_\text{L6\_D384}$ \textit{to} $\text{BERT}_\text{L12\_D768}$} \\
    $\mathcal{M}_{1}$ & $30.3$M      & $6$      & $384$      &  $6$     &   $1536$    &  $5.0\times10^{-4}$ & - & $62.5$k & $ 8\%$ & $6.0\times10^{4}$\\
    $\mathcal{M}_{2}$ & $51.5$M      & $8$      & $512$      &  $8$     &   $2048$    &  $5.0\times10^{-4}$  & $5$k & $20$k & $ 8\%$ & $2.4\times10^{4}$\\
    $\mathcal{M}_{3}$ & $82.2$M      & $10$      & $640$      &  $10$     &   $2560$    &  $5.0\times10^{-4}$  & $5$k & $20$k  & $ 8\%$ & $5.0\times10^{4}$\\
    $\mathcal{M}_{4}$ &  $102$M     & $11$   &  $704$ &  $11$ &  $2816$  &  $5.0\times10^{-4}$ & $5$k & $20$k & $ 8\%$ & $5.8\times10^{4}$\\
    $\mathcal{M}_{5}$ &  $125$M     & $12$   &  $768$ &  $12$ &  $3072$  &  $5.0\times10^{-4}$  & $5$k & $20$k & $ 8\%$ & $6.8\times10^{4}$\\
    \hline
    \multicolumn{11}{l}{\textit{Growing from} $\text{BERT}_\text{L12\_D768}$ \textit{to} $\text{BERT}_\text{L24\_D1024}$} \\
    $\mathcal{M}_{1}$ & $125$M      & $12$   &  $768$ &  $12$ &  $3072$  &  $5.0\times10^{-4}$ & - & $62.5$k & $ 8\%$ & $1.9\times10^{5}$\\
    $\mathcal{M}_{2}$ & $216$M      & $15$      & $960$      &  $15$     &   $3840$    &  $2.5\times10^{-4}$  & $1$k & $20$k  & $ 20\%$ & $6.5\times10^{4}$\\
    $\mathcal{M}_{3}$ & $280$M      & $18$      & $1024$      &  $16$     &   $4096$    &  $2.5\times10^{-4}$  & $1$k & $20$k  & $ 20\%$ & $1.4\times10^{5}$\\
    $\mathcal{M}_{4}$ &  $318$M     & $21$   &  $1024$ &  $16$ &  $4096$  &  $2.5\times10^{-4}$  &  $1$k & $20$k & $ 20\%$ & $1.7\times10^{5}$\\
    $\mathcal{M}_{5}$ &  $355$M     & $24$   &  $1024$ &  $16$ &  $4096$  &  $2.5\times10^{-4}$  &  $1$k & $20$k & $ 20\%$ & $2.2\times10^{5}$  \\
    \hline
    \multicolumn{11}{l}{\textit{Growing from} $\text{GPT}_\text{L6\_D384}$ \textit{to} $\text{GPT}_\text{L12\_D768}$} \\
    $\mathcal{M}_{1}$ & $29.9$M      & $6$      & $384$      &  $6$     &   $1536$    &  $5.0\times10^{-4}$  &  - & $62.5$k  & $ 16\%$ & $6.7\times10^{4}$\\
    $\mathcal{M}_{2}$ & $51.0$M      & $8$      & $512$      &  $8$     &   $2048$    &  $5.0\times10^{-4}$  & $5$k & $20$k  & $ 16\%$ & $3.9\times10^{4}$\\
    $\mathcal{M}_{3}$ & $81.4$M      & $10$      & $640$      &  $10$     &   $2560$    &  $5.0\times10^{-4}$  & $5$k & $20$k  & $ 16\%$ & $5.6\times10^{4}$\\
    $\mathcal{M}_{4}$ &  $101$M     & $11$   &  $704$ &  $11$ &  $2816$  &  $5.0\times10^{-4}$ &  $5$k & $20$k & $ 16\%$ & $ 6.8\times10^{4}$\\
    $\mathcal{M}_{5}$ &  $124$M     & $12$   &  $768$ &  $12$ &  $3072$  &  $5.0\times10^{-4}$ &  $5$k & $20$k & $ 16\%$ & $7.8\times10^{4}$ \\
    \bottomrule
    \end{tabu}%
  \caption{Model architectures, learning rate (lr), steps of FRW (SF), steps of training after FRW (STF), the ratio of steps for learning rate warmup (for both FRW and pre-training) (RW), and train wall time (TWT) for all the models pre-trained with ELLE in this paper. We list the details when growing $\text{BERT}_\text{L6\_D384}$ to $\text{BERT}_\text{L12\_D768}$, $\text{BERT}_\text{L12\_D768}$ to $\text{BERT}_\text{L24\_D1024}$ and $\text{GPT}_\text{L6\_D384}$ to $\text{GPT}_\text{L12\_D768}$, respectively. The total train wall time consumed by the above three settings is $2.57 \times 10^5$ seconds, $7.79 \times 10^5$ seconds, and $3.08 \times 10^5$ seconds, respectively.}
  \label{tab:arch}%
\end{table*}%

\section{Pre-training Hyper-parameters}
\label{sec:pretrain_hyper}
In Table \ref{tab:arch}, we list the architectures and the hyper-parameters for the PLMs we pre-trained with ELLE in this paper, including the total number of trainable parameters ($n_{\text{params}}$), the number of layers ($n_{\text{layers}}$), the number of units in each bottleneck layer ($d_{\text{model}}$), the number of attention heads ($n_{\text{heads}}$), the inner hidden size of FFN layer ($d_{\text{FFN}}$), the learning rate (lr), the training steps of FRW (SF), the training steps of adaptation after FRW (STF) when learning the new corpus, the ratio of learning rate warmup (RW), and the total train wall time (TWT). We set the dropout rate for each model to $0.1$, weight decay to $0.01$ and use linear learning rate decay for BERT and inverse square root decay for GPT. We adopt Adam~\citep{kingma2014adam} as the optimizer. The hyper-parameters for the optimizer is set to $1\times10^{-6}, 0.9, 0.98$ for $\epsilon, \beta_1, \beta_2$, respectively. We reset the optimizer and the learning rate scheduler each time when the PLM finishes FRW or the training a on new corpus. All experiments are conducted under the same computation environment with $8$ NVIDIA 32GB V100 GPUs. All the pre-training implementations are based on \texttt{fairseq}\footnote{\url{https://github.com/pytorch/fairseq}}~\cite{ott2019fairseq} (MIT-license).

\section{Implementation Details and Additional Experiments for Downstream Fine-tuning}
\label{sec:finetune_detail}
\begin{table*}[thbp]
  \centering
  \small
    \begin{tabular}{lccccc}
    \toprule
    \textbf{HyperParam} & \textsc{MNLI} & \textsc{HyperPartisan} & \textsc{Helpfulness} & \textsc{ChemProt}& \textsc{ACL-ARC}   \\
    \midrule
    Learning Rate &  $1\times10^{-5}$     & $2\times10^{-5}$     & $2\times10^{-5}$     & $2\times10^{-5}$     & $2\times10^{-5}$      \\
    Batch Size &   $32$    & $256$    & $256$    & $256$    &   $256$ \\
    Weight Decay &  $0.1$     & $0.1$     & $0.1$     & $0.1$     & $0.1$ \\
    Max Epochs &    $10$   & $10$   & $10$   & $10$   &   $10$ \\
    Learning Rate Decay &  Linear     & Linear     & Linear     & Linear     &  Linear \\
    Warmup Ratio &  $0.06$     &  $0.06$     &  $0.06$     &  $0.06$     &   $0.06$ \\
    \bottomrule
    \end{tabular}%
  \caption{Hyper-parameters for fine-tuning on downstream tasks of each domain. As mentioned in the main paper, for each domain, we select a representative task that is relatively stable, i.e., \textsc{MNLI}~\citep{williams2017broad}, \textsc{HyperPartisan}~\citep{kiesel2019semeval}, \textsc{Helpfullness}~\citep{mcauley2015image}, \textsc{ChemProt}~\citep{kringelum2016chemprot} and \textsc{ACL-ARC}~\citep{jurgens2018measuring} for \textsc{WB}, \textsc{Ns}, \textsc{Rev}, \textsc{Bio} and \textsc{CS}, respectively.}
  \label{tab:finetune}%
\end{table*}%

\paragraph{Implementation Details.}
Table \ref{tab:finetune} describes the hyper-parameters for fine-tuning PLMs on downstream tasks of each domain. The implementations of MNLI are based on \texttt{fairseq}\footnote{\url{https://github.com/pytorch/fairseq}}~\cite{ott2019fairseq} (MIT-license). The implementations of \textsc{HyperPartisan}, \textsc{Helpfulness} \textsc{ChemProt}, and ACL-ARC  are based on \cite{gururangan2020don}\footnote{\url{https://github.com/allenai/dont-stop-pretraining}}. 

\begin{figure}[!t]
    \centering
    \includegraphics[width=0.48\textwidth]{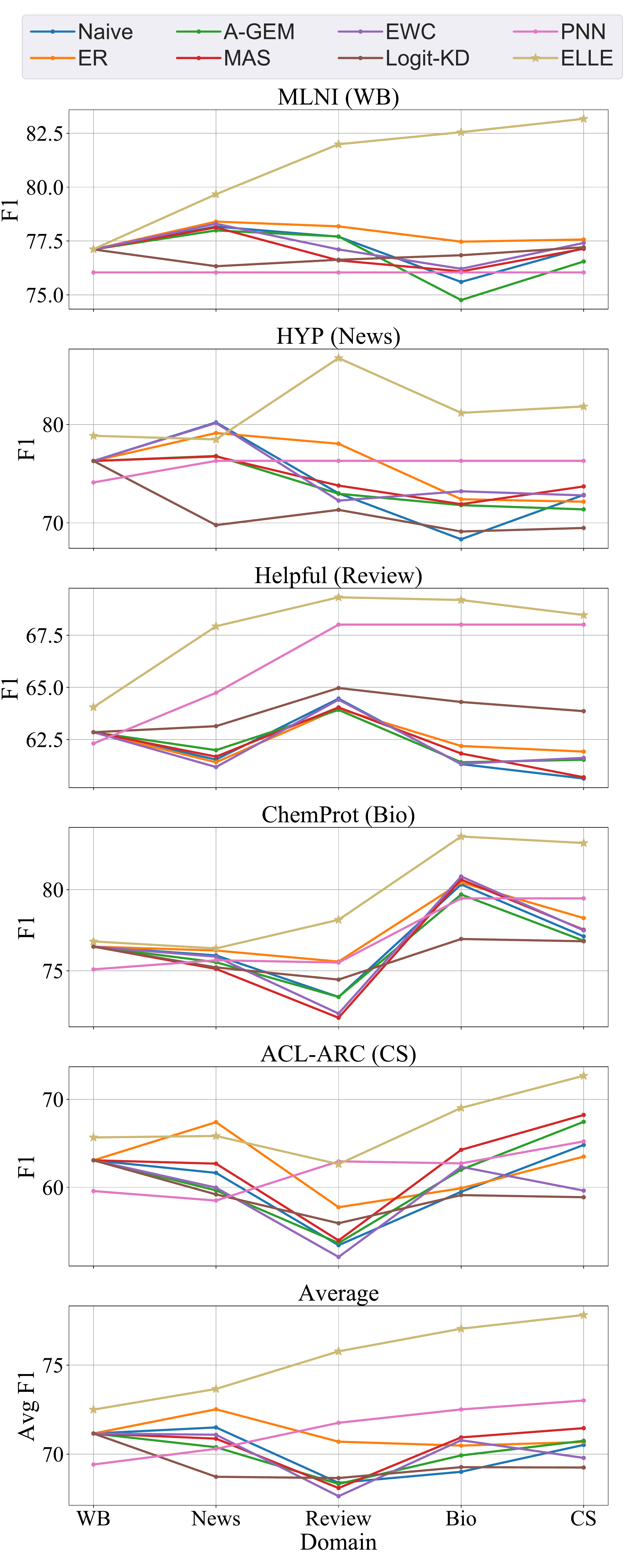}
    \caption{Specific and average F1 on downstream tasks of all domains of different lifelong learning methods. The initial PLM is chosen as $\text{BERT}_\text{L6\_D384}$. The score is evaluated after each model finishes training on each domain.}
    \label{fig:avg_f1}
\end{figure}

\paragraph{Additional Experiments.}
Figure~\ref{fig:avg_f1} visualizes the specific F1 on each downstream task and the average F1 of PLMs trained with Naive, A-GEM, EWC, MAS, ER, Logit-KD, PNN and \ourmodel after finishing training on each domain when we choose $\text{BERT}_\text{L6\_D384}$ as the initial PLM $\mathcal{M}_1$. The average F1 when finishing training on the $i$-th domain is calculated as follows:
\begin{equation}
    \text{F1}_{avg}^i = \frac{1}{N}\sum_{j=1}^N \text{F1}^j_{\mathcal{M}_i}
\end{equation}
where $\text{F1}^j_{\mathcal{M}_i}$ is the F1 score of $\mathcal{M}_i$ evaluated on the downstream task of the $j$-th domain. We also list the detailed numerical results for each task in Table~\ref{table:f1}, covering all PLMs trained by each lifelong learning method.

The results show that \ourmodel outperforms all the lifelong learning baselines after finishing training on each domain, demonstrating that \ourmodel could properly stimulate the learned knowledge during pre-training and boost the performance in downstream tasks.

\section{Trend Curves for AP and $\text{AP}^{+}$}
\label{sec:trend_curve}
\begin{figure*}[!t]
    \centering
    \includegraphics[width=0.45\textwidth]{figs/main_avg_ppl.pdf}
    \includegraphics[width=0.45\textwidth]{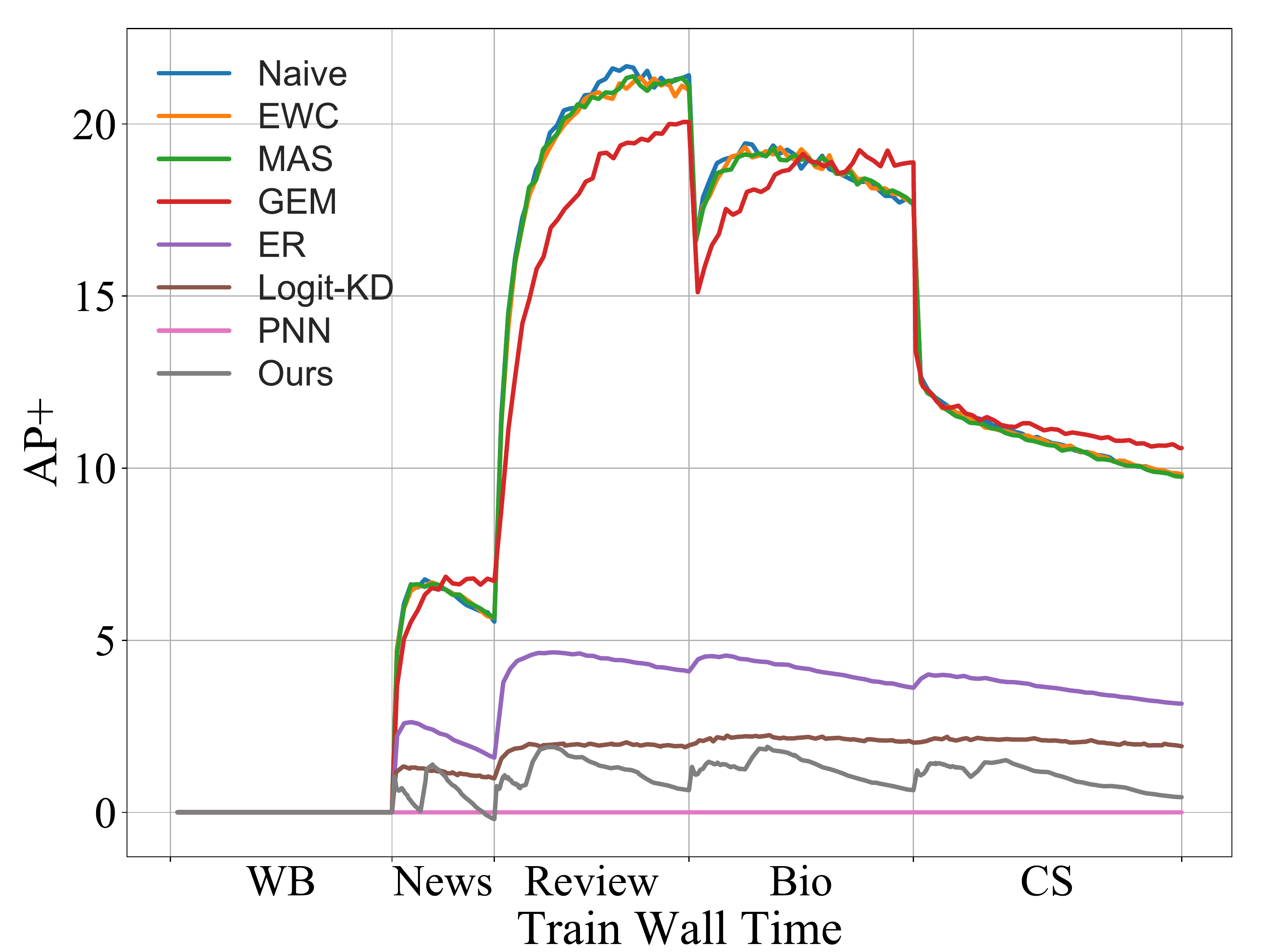}
    \caption{AP and $\text{AP}^+$ of different lifelong learning methods with $\text{BERT}_\text{L6\_D384}$ as the initial PLM w.r.t train wall time. \ourmodel continually grows $\text{BERT}_\text{L6\_D384}$ to $\text{BERT}_\text{L12\_D768}$.}
    \label{fig:bert_small}
\end{figure*}

\begin{figure*}[!t]
    \centering
    \includegraphics[width=0.45\textwidth]{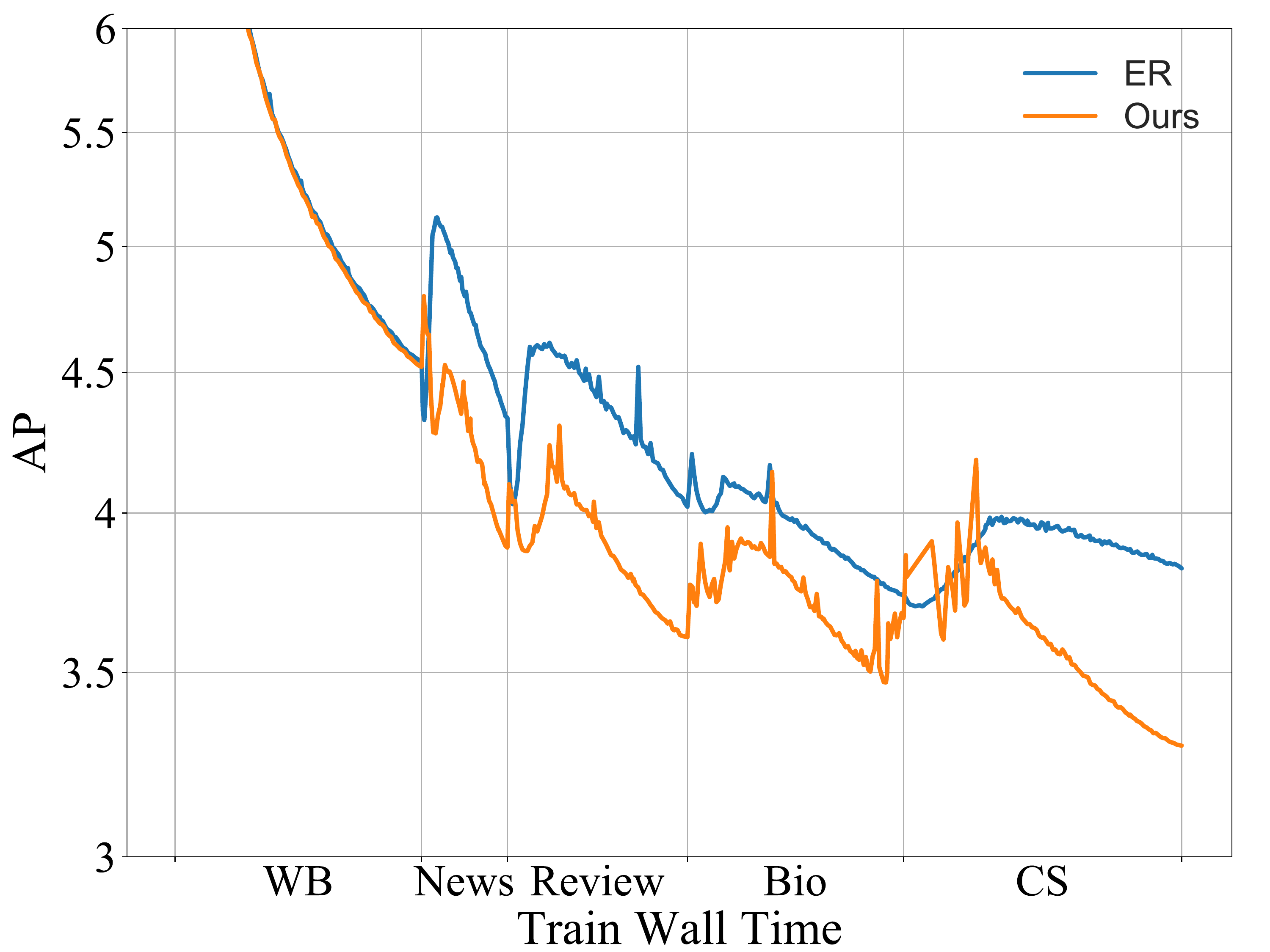}
    \includegraphics[width=0.45\textwidth]{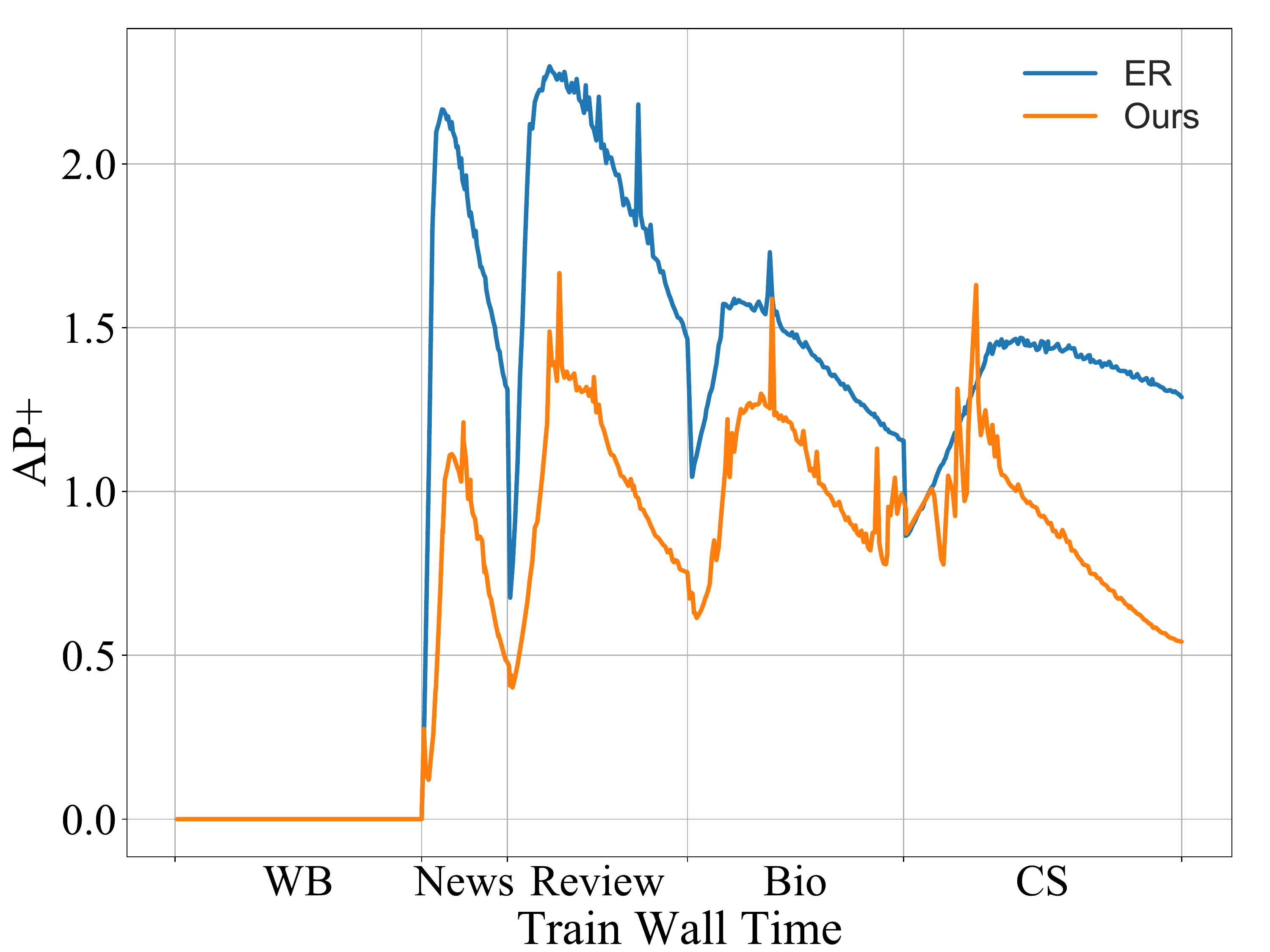}
    \caption{AP and $\text{AP}^+$ of \ourmodel when growing $\text{BERT}_\text{L12\_D768}$ to $\text{BERT}_\text{L24\_D1024}$.}
    \label{fig:bert_large}
\end{figure*}

\begin{figure*}[!t]
    \centering
    \includegraphics[width=0.45\textwidth]{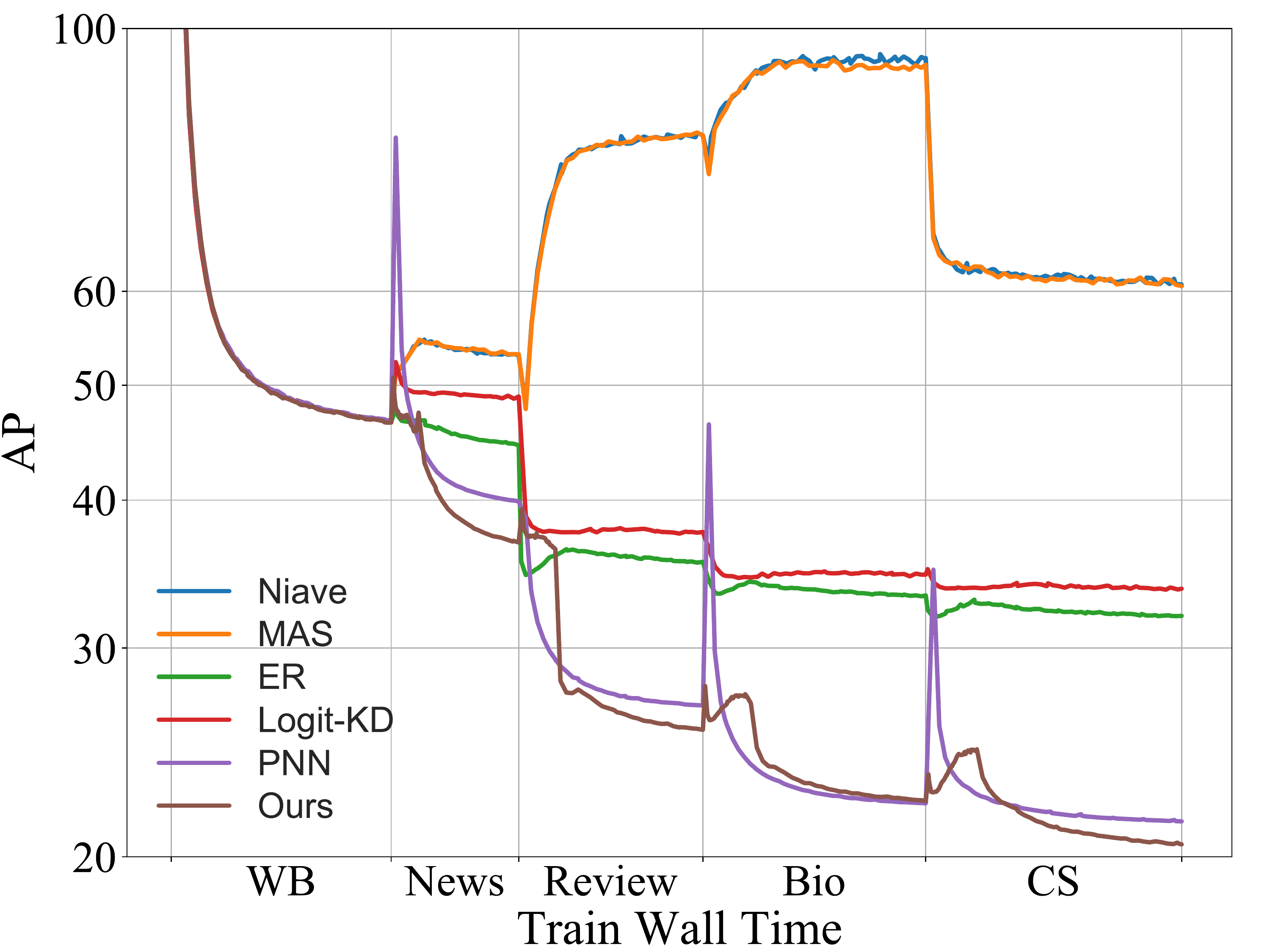}
    \includegraphics[width=0.45\textwidth]{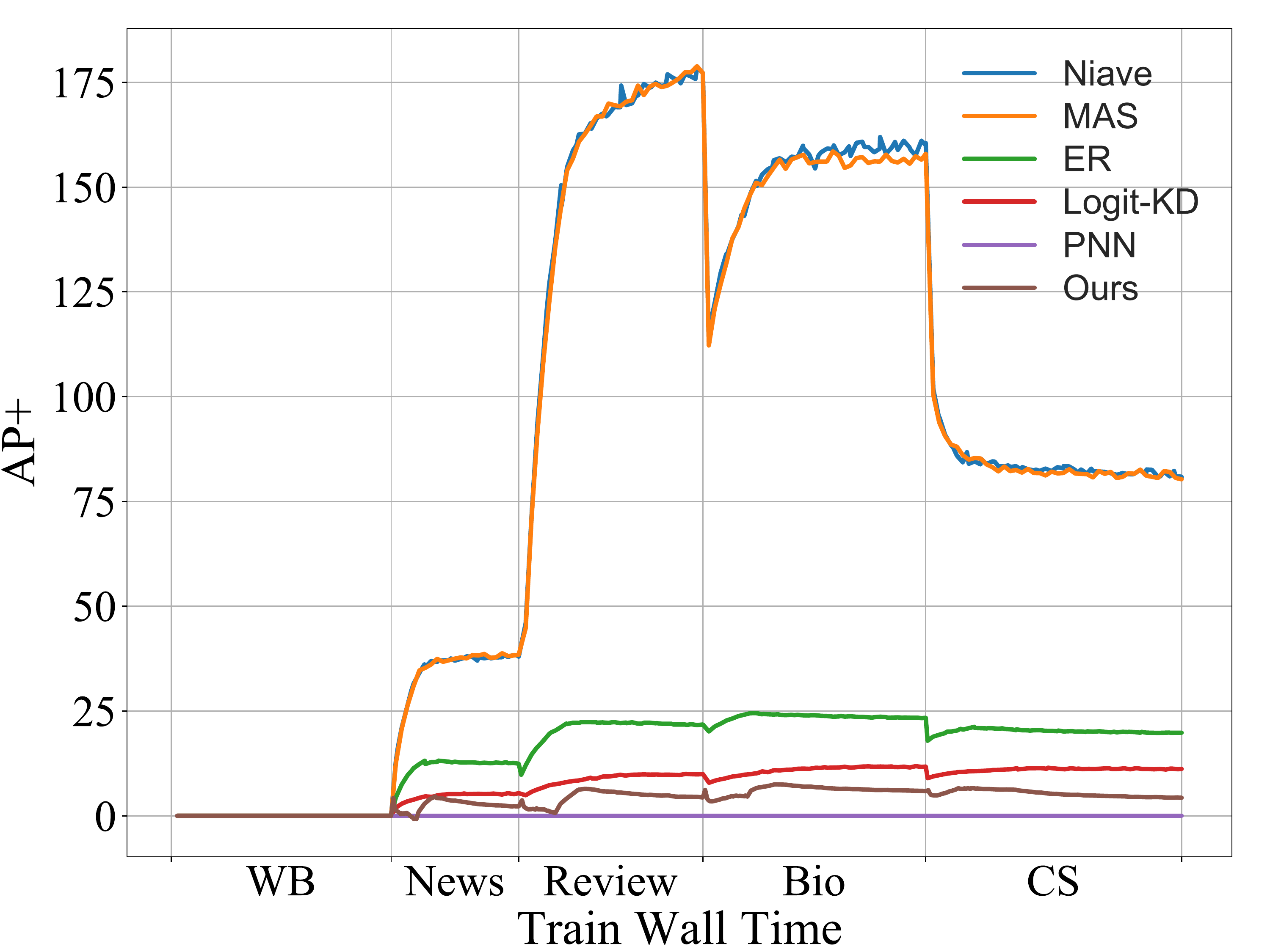}
    \caption{AP and $\text{AP}^+$ of different lifelong learning methods with $\text{GPT}_\text{L6\_D384}$ as the initial PLM w.r.t train wall time. \ourmodel continually grows $\text{GPT}_\text{L6\_D384}$ to $\text{GPT}_\text{L12\_D768}$.}
    \label{fig:gpt}
\end{figure*}

For the experiments in \cref{sec:main_exp}, the trend curves of average perplexity (AP) and average increased perplexity ($\text{AP}^+$) w.r.t train wall time are shown in Figure~\ref{fig:bert_small} (growing from $\text{BERT}_\text{L6\_D384}$ to $\text{BERT}_\text{L12\_D768}$), Figure~\ref{fig:bert_large} (growing from $\text{BERT}_\text{L12\_D768}$ to $\text{BERT}_\text{L24\_D1024}$), and Figure~\ref{fig:gpt} (growing from $\text{GPT}_\text{L6\_D384}$ to $\text{GPT}_\text{L12\_D768}$). Each figure illustrates the performance of different lifelong learning methods. The above results reflect that, compared with all the baselines, $\text{AP}$ and $\text{AP}^+$ of \ourmodel descend with the fastest speed, demonstrating that \ourmodel could acquire knowledge and mitigate the knowledge forgetting on previous domains more efficiently. Thus given limited computational resources, PLMs trained by \ourmodel could integrate more information from different domains.

\begin{figure*}[!t]
    \centering
    \includegraphics[width=0.45\textwidth]{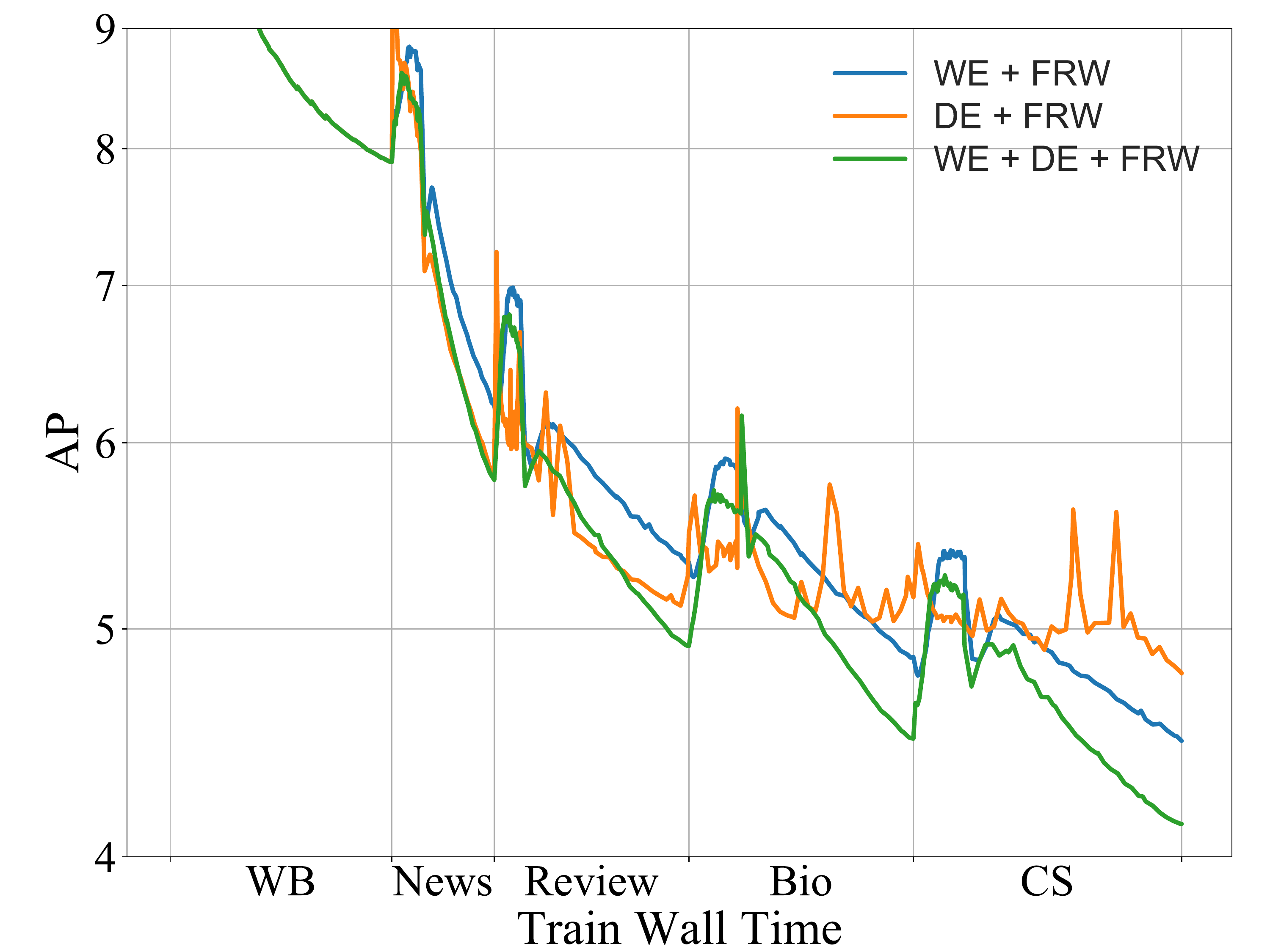}
    \includegraphics[width=0.45\textwidth]{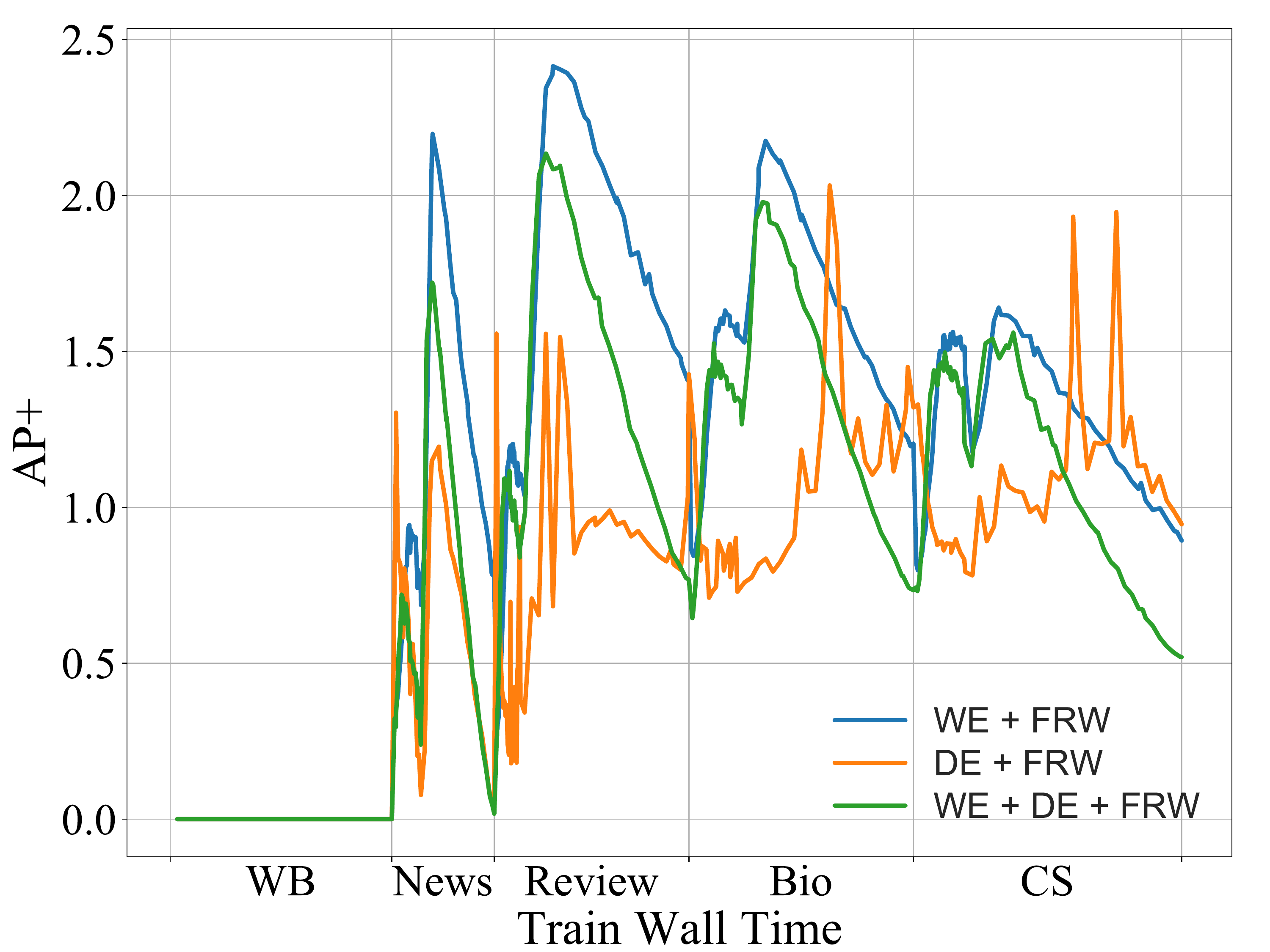}
    \caption{AP and $\text{AP}^+$ of PLMs trained with different model expansion strategies: expanding width only (WE+FRW), expanding depth only (DE+FRW) and expanding width and depth together (WE+DE+FRW) w.r.t train wall time.}
    \label{fig:expansion}
\end{figure*}

\begin{figure*}[!t]
    \centering
    \includegraphics[width=0.45\textwidth]{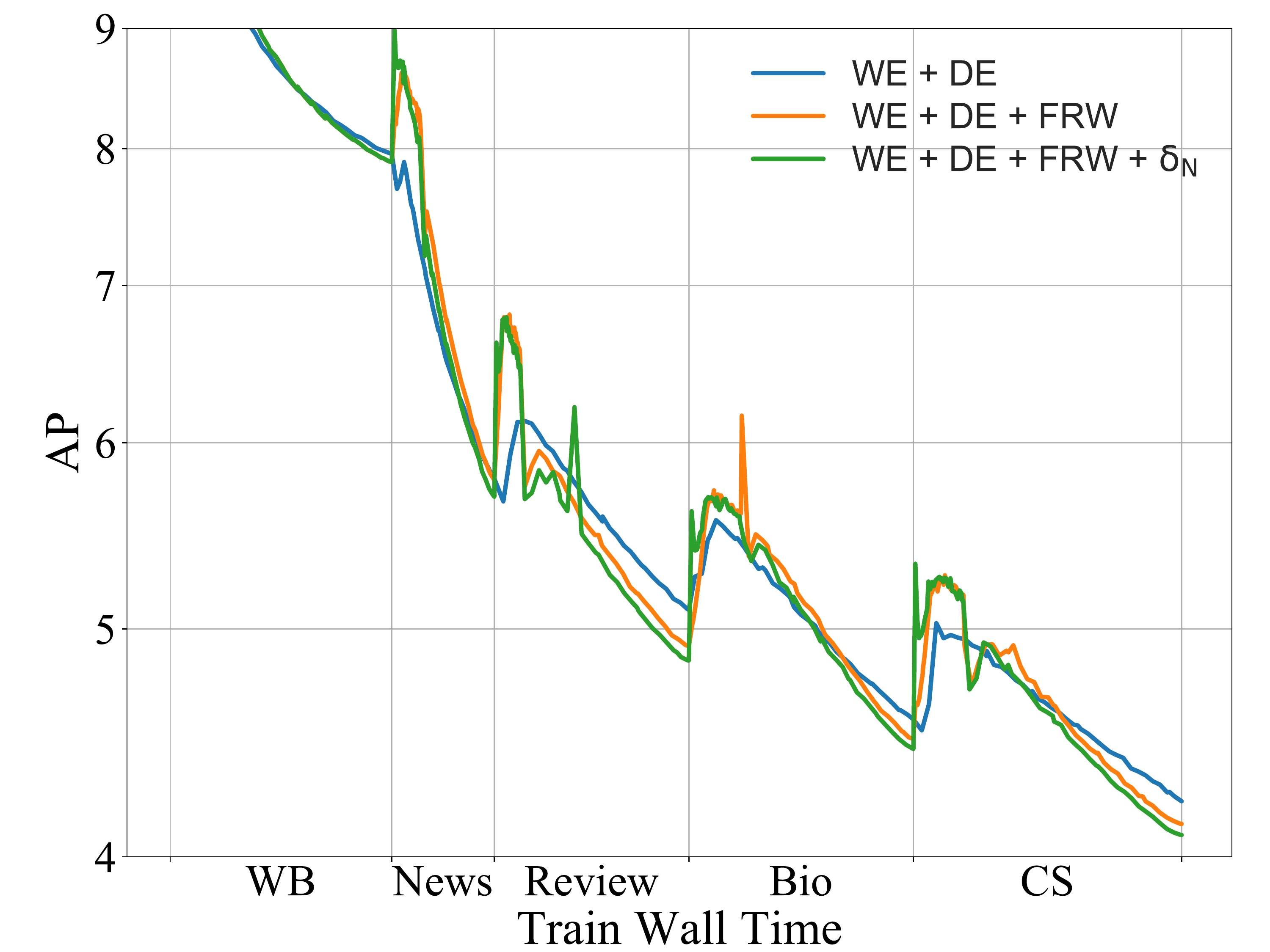}
    \includegraphics[width=0.45\textwidth]{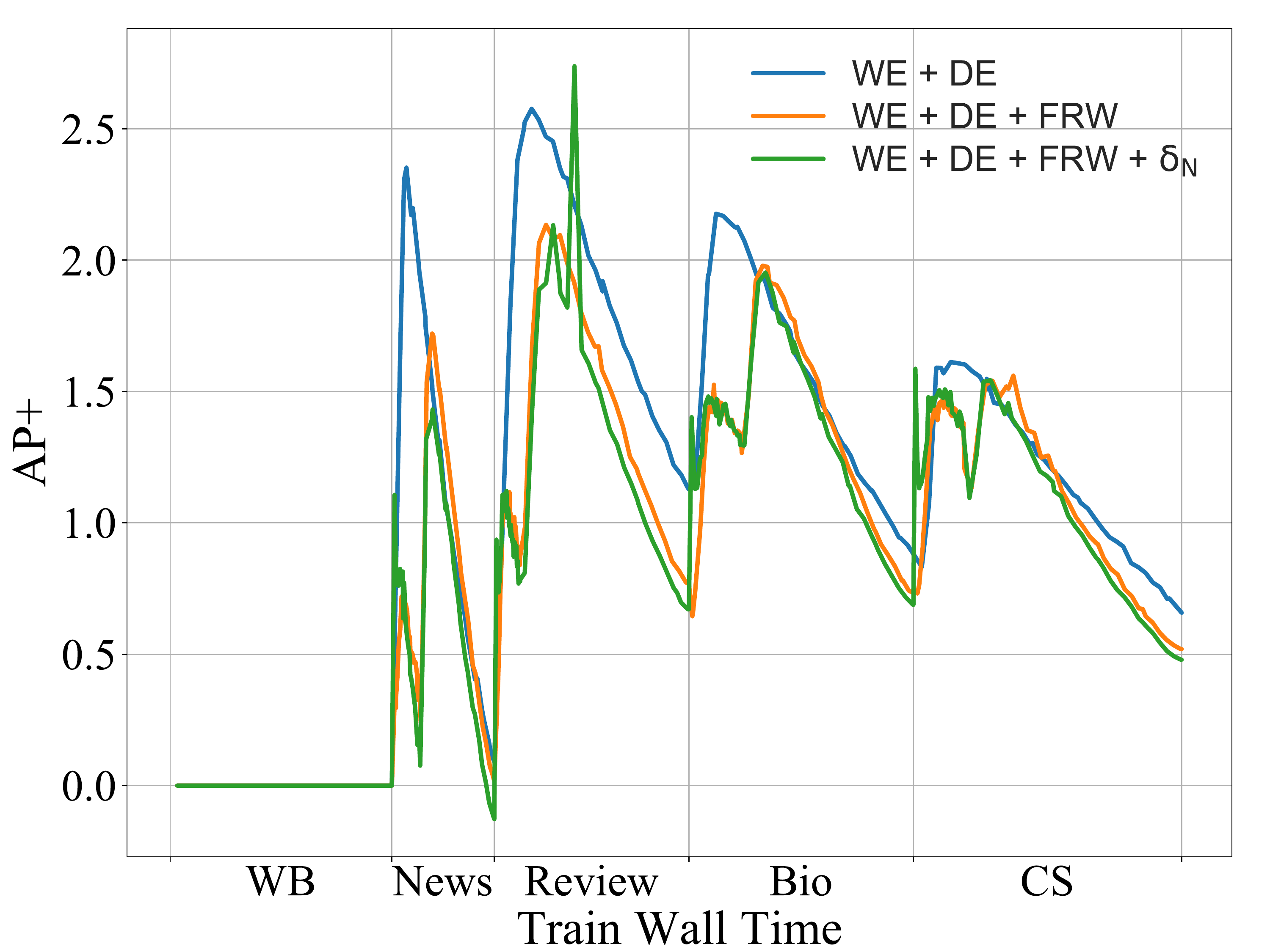}
    \caption{AP and $\text{AP}^+$ of PLMs trained by WE+DE, WE+DE+FRW, WE+DE+FRW+$\delta_N$ w.r.t train wall time.}
    \label{fig:FRW}
\end{figure*}

\begin{figure*}[!t]
    \centering
    \includegraphics[width=0.45\textwidth]{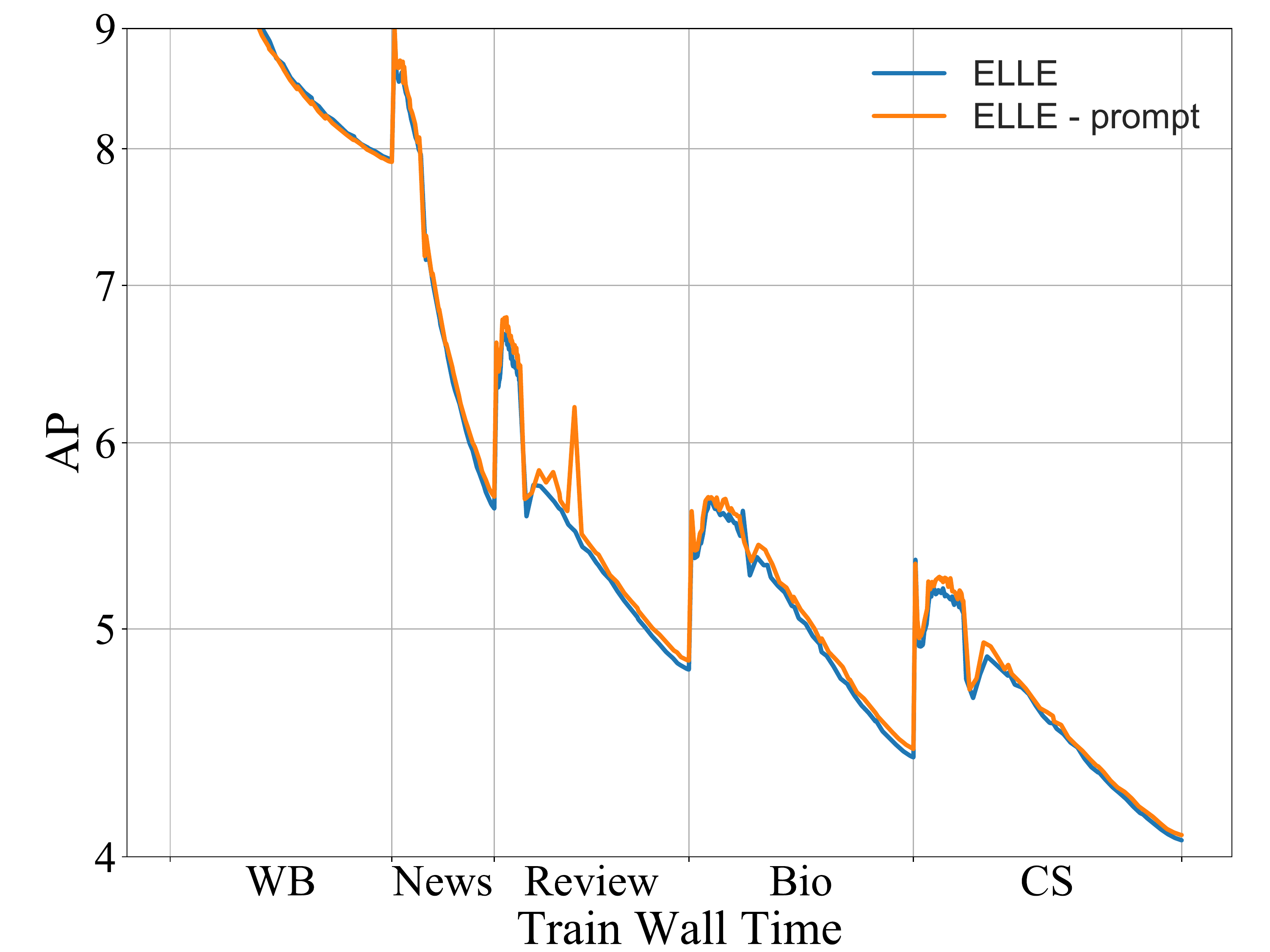}
    \includegraphics[width=0.45\textwidth]{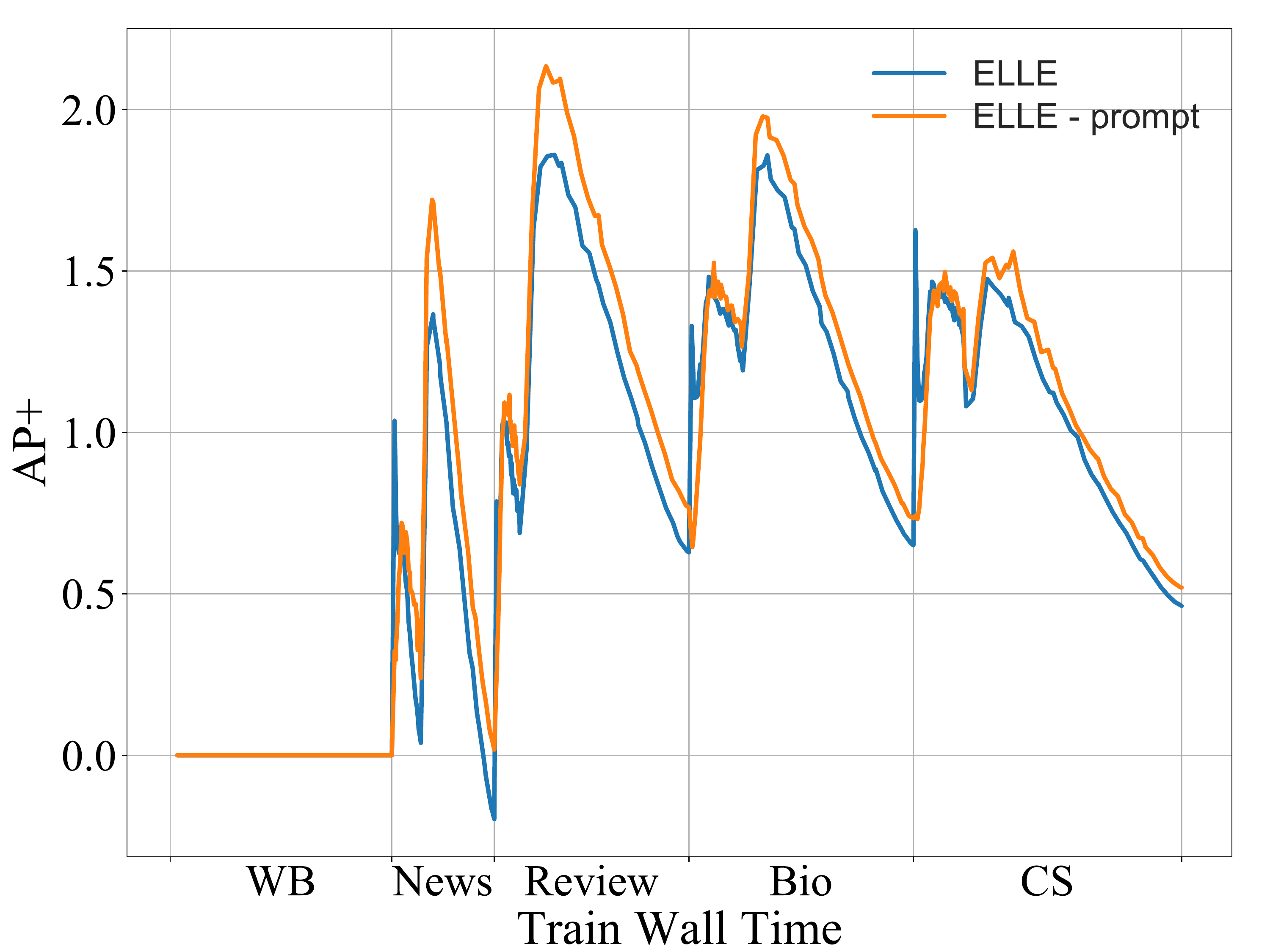}
    \caption{AP and $\text{AP}^+$ of PLMs trained by \ourmodel with and without domain prompts w.r.t train wall time.}
    \label{fig:prompt}
\end{figure*}

For the analysis in \cref{sec:analysis}, we visualize the trend curves of AP and $\text{AP}^{+}$ when choosing different combinations of strategies. Specifically, we investigate (1) the effect of width / depth expansion in Figure~\ref{fig:expansion} (comparing WE+FRW, DE+FRW and WE+DE+FRW); (2) the effect of function recovering warmup in Figure~\ref{fig:FRW} (comparing WE+DE and WE+DE+FRW); (3) the effect of random noises added into the newly initialized parameters during model expansion in Figure~\ref{fig:FRW} (comparing WE+DE+FRW and WE+DE+FRW+$\delta_N$) and (4) the effect of pre-trained domain prompts in Figure~\ref{fig:prompt} (comparing \ourmodel and \ourmodel-PT). All of the above results again demonstrate the effectiveness of \ourmodel's each component.

% \section{Comparison between \ourmodel and \citet{jin2021lifelong}}
% \label{sec:comparison}
% Since for PLMs, pre-training with more computations almost always results in better performance~\citep{clark2020electra,li2020train,kaplan2020scaling}, a simple yet effective method to integrate the information from all domains is to continually pre-train existing PLMs on all the existing data exhaustively. In this regard, the most important consideration for lifelong pre-training should be the training efficiency. Therefore, when comparing different lifelong learning methods, it is important to equalize the computational costs consumed by each method. Conforming to this rule, we control the computational costs (estimated by train wall time~\citep{li2020train}) for all the methods in our experiments the same, and find that \ourmodel tends to be the most training efficient and could help PLMs acquire more knowledge.

\begin{table}[!t]
\centering
\small
\begin{tabu}{l@{~~~~}c@{~~~~}c@{~~~~}c@{~~~~}c@{~~~~}c@{~~~~}c}
\toprule 
\multicolumn{1}{l}{\textbf{Domain}}   & \textsc{WB}    & \textsc{Ns}  & \textsc{Rev} & \textsc{Bio}   & \textsc{CS}    & \textsc{AVG}    \\\hline
\multicolumn{6}{l}{Naive}\\
$\mathcal{M}_1$                              & $77.11$ &  $76.29$ & $62.85$       &  $76.49$     &   $63.07$    & $71.16$ \\
$\mathcal{M}_2$                              & $78.17$  & $80.21$  &    $61.54$    & $75.95$      &    $61.64$   & $71.50$   \\
$\mathcal{M}_3$                              & $77.70$   & $73.00$  & $64.46$   &   $73.39$    &  $53.41$     & $68.39$   \\
$\mathcal{M}_4$                              & $75.60$  & $68.33$  & $61.32$   & $80.32$ &   $59.49$    & $69.01$  \\
$\mathcal{M}_5$                              & $77.18$  & $72.84$   & $60.63$  & $77.12$  & $64.82$  & $70.52$   \\\hline
\multicolumn{6}{l}{A-GEM}\\
$\mathcal{M}_1$                              & $77.11$ &  $76.29$ & $62.85$       &  $76.49$     &   $63.07$    & $71.16$ \\
$\mathcal{M}_2$                              & $77.99$  & $76.80$  &    $61.99$    & $75.53$      &    $59.65$   & $71.50$   \\
$\mathcal{M}_3$                              & $77.71$   & $72.96$  & $63.92$   &   $73.39$    &  $53.66$     & $68.39$   \\
$\mathcal{M}_4$                              & $74.76$  & $71.80$  & $61.41$   & $79.70$ &   $62.00$    & $69.93$  \\
$\mathcal{M}_5$                              & $76.55$  & $71.37$   & $61.53$  & $76.85$  & $64.82$  & $70.75$   \\\hline
\multicolumn{6}{l}{MAS}\\
$\mathcal{M}_1$                              & $77.11$ &  $76.29$ & $62.85$       &  $76.49$     &   $63.07$    & $71.16$ \\
$\mathcal{M}_2$                              & $78.13$  & $76.75$  &   $61.68$     &   $75.12$    &   $62.69$    & $70.87$   \\
$\mathcal{M}_3$                              & $76.60$   & $73.79$  & $64.04$   &   $72.11$    &  $53.95$     & $70.87$   \\
$\mathcal{M}_4$                              & $76.09$  & $71.90$  & $61.83$   & $80.62$ &  $64.26$     & $70.94$  \\
$\mathcal{M}_5$                              & $77.14$  & $73.70$   & $60.69$  & $77.53$  & $68.23$  & $71.46$   \\\hline
\multicolumn{6}{l}{MAS}\\
$\mathcal{M}_1$                              & $77.11$ &  $76.29$ & $62.85$       &  $76.49$     &   $63.07$    & $71.16$ \\
$\mathcal{M}_2$                              & $78.30$  & $80.15$  &   $61.18$     &   $75.87$    &   $59.96$    & $71.09$   \\
$\mathcal{M}_3$                              & $77,11$   & $72.26$  & $64.41$   &   $72.37$    &  $52.07$     & $67.64$   \\
$\mathcal{M}_4$                              & $76.21$  & $73.21$  & $61.34$   & $80.81$ &  $62.33$     & $70.78$  \\
$\mathcal{M}_5$                              & $77.41$  & $72.79$   & $61.62$  & $77.49$  & $59.62$  & $69.79$   \\\hline
\multicolumn{6}{l}{ER}      \\
$\mathcal{M}_1$                              & $77.11$ &    $76.29$   &   $62.85$     &     $76.49$     &   $63.07$   & $71.16$  \\
$\mathcal{M}_2$                              & $78.40$   & $79.13$  &   $61.41$     &   $76.25$    &    $67.41$   & $72.52$   \\
$\mathcal{M}_3$                              & $78.18$  & $78.04$  & $63.98$   &   $75.57$    &  $57.53$      & $70.70$  \\
$\mathcal{M}_4$                              & $77.47$  & $72.40$   & $62.19$  & $80.44$  &   $59.89$    & $73.13$   \\
$\mathcal{M}_5$                              & $77.57$  & $72.15$  & $61.92$  & $78.25$  & $63.49$  & $70.68$   \\\hline
\multicolumn{6}{l}{Logit-KD} \\
$\mathcal{M}_1$                              & $77.11$ &   $79.29$    &   $62.85$     &   $76.49$    &      $64.07$ & $71.16$  \\
$\mathcal{M}_2$                              & $76.33$  & $69.77$  &   $63.14$     &   $75.21$    &    $59.19$   & $68.73$   \\
$\mathcal{M}_3$                              & $76.63$  & $71.32$ & $64.97$   &   $74.46$    &    $55.91$   & $68.66$   \\
$\mathcal{M}_4$                              & $76.84$  & $69.12$ & $64.30$    & $76.96$  &   $59.11$    & $69.27$  \\
$\mathcal{M}_5$                              & $77.21$ & $69.48$  & $63.86$   & $76.82$ & $58.87$  & $69.25$   \\\hline
\multicolumn{6}{l}{PNN} \\
$\mathcal{M}_1$                              & $76.04$ &   $74.11$    &   $62.31$     &   $75.09$    &      $59.57$ & $69.42$  \\
$\mathcal{M}_2$                              & $76.04$  & $76.30$  &   $64.74$     &   $75.65$    &    $59.19$   & $70.24$   \\
$\mathcal{M}_3$                              & $76.04$  & $76.30$ & $68.01$   &   $75.51$    &    $55.91$   & $71.76$   \\
$\mathcal{M}_4$                              & $76.04$  & $76.30$ & $68.01$    & $79.46$  &   $59.11$    & $72.51$  \\
$\mathcal{M}_5$                              & $76.04$ & $76.30$  & $68.01$   & $79.46$ & $58.87$  & $73.01$   \\\hline
\multicolumn{6}{l}{ELLE}     \\
$\mathcal{M}_1$                              & $77.12$ &   $78.85$    &   $64.05$     &   $76.81$    &    $65.67$   & $72.50$  \\
$\mathcal{M}_2$                              & $79.67$  & $78.48$  &   $67.93$     &    $76.38$   &   $65.84$    & $73.66$   \\
$\mathcal{M}_3$                              & $81.99$  & $86.75$  & $69.32$  &   $78.14$    &   $62.63$    & $75.77$   \\
$\mathcal{M}_4$                              & $82.55$  & $81.18$  & $69.19$  & $83.27$  &   $69.03$    & $77.04$   \\
$\mathcal{M}_5$                              & $83.17$  & $81.83$  & $68.47$   & $82.87$  & $72.69$  & $77.81$ \\\bottomrule
\end{tabu}
\caption{Specific and average F1 scores on downstream tasks from each domain after the PLM finishes training on each domain. We evaluate PLMs trained with different lifelong learning methods that choose $\text{BERT}_\text{\L6\_D384}$ as the initial model $\mathcal{M}_1$. }
\label{table:f1}
\end{table}

\begin{figure}[!t]
    \centering
    \includegraphics[width=0.45\textwidth]{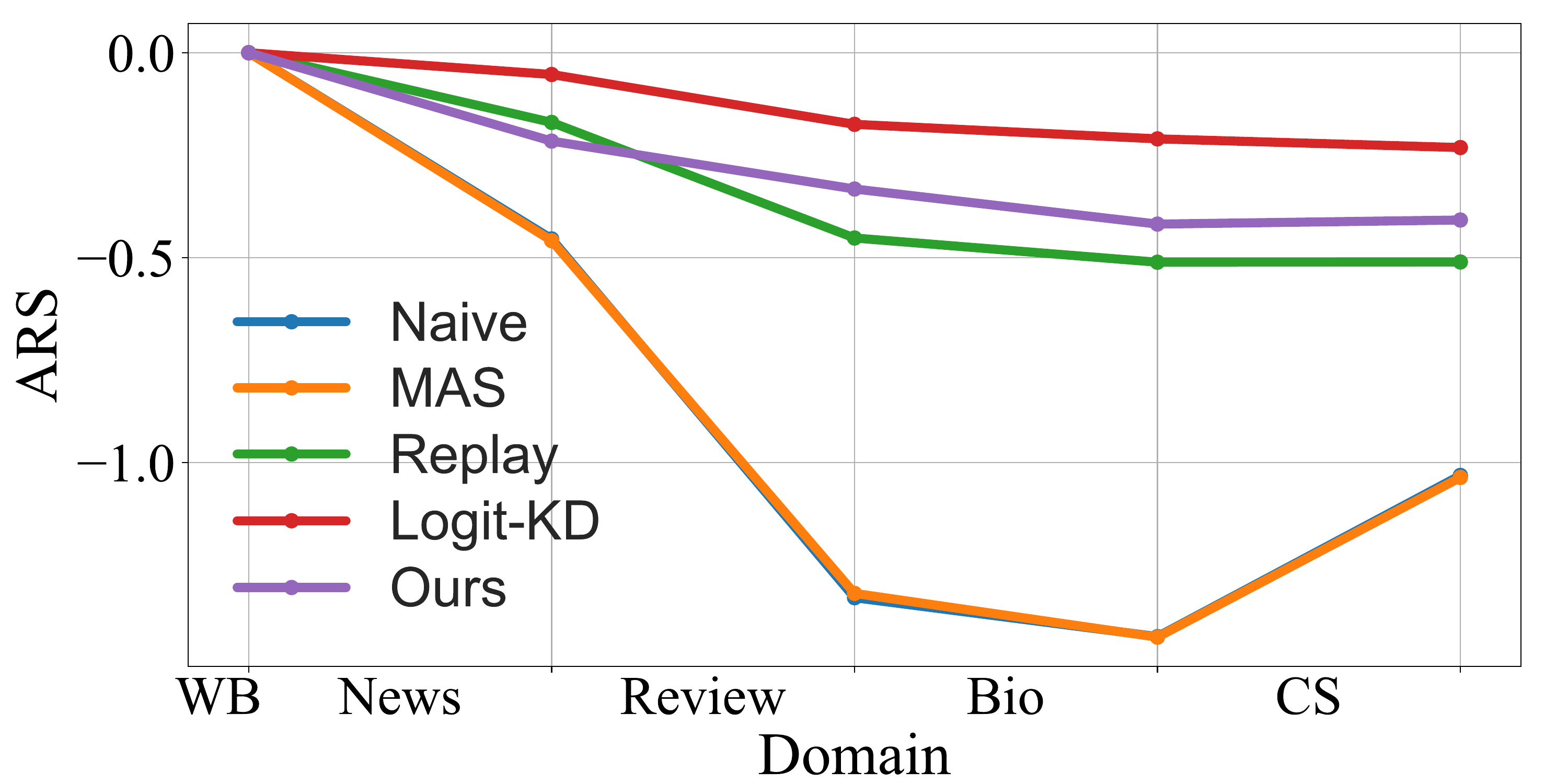}
    \caption{Average representational similarity (ARS) of a stream of PLMs comparing different lifelong learning algorithms. We choose $\text{BERT}_\text{L6\_D384}$ as the initial PLM $\mathcal{M}_1$.}
    \label{fig:similarity}
\end{figure}

\begin{table*}[!t]
\small
\centering
\begin{tabu}{l@{~~~~}r@{~~~~}c@{~~~~}r@{~~~~}r@{~~~~}r@{~~~~}r@{~~~~}r@{~~~~}r@{~~~~}r@{~~~~}r}
\toprule
% \multicolumn{2}{c}{\textbf{Evaluated After}} & \multicolumn{2}{c}{Domain $1} & \multicolumn{2}{c}{Domain $2}    & \multicolumn{2}{c}{Domain $3}   & \multicolumn{2}{c}{Domain $4}   & \multicolumn{2}{c}{Domain $5}   \\ \hline
\multicolumn{1}{c}{\textbf{Domain}}          & \multicolumn{2}{c}{\textsc{WB}}       & \multicolumn{2}{c}{\textsc{Ns}}        & \multicolumn{2}{c}{\textsc{Rev}}     & \multicolumn{2}{c}{\textsc{Bio}}        & \multicolumn{2}{c}{\textsc{CS}}         \\ \hline
\multicolumn{1}{c}{\textbf{Metrics}}         & AP             & AP+       & AP            & AP+           & AP            & AP+          & AP            & AP+          & AP            & AP+          \\ \hline
\multicolumn{11}{l}{\textit{Half train wall time}} \\
MAS      & $7.96$          & - & $8.50$          & $6.22$          & $12.85$         & $18.88$         & $13.99$         & $17.52$         & $10.31$         & $10.22$         \\
ER       & $7.96$          & - & $7.12$          & $1.98$          & $7.11$          & $4.14$          & $6.83$          & $3.77$          & $6.53$          & $3.78$          \\
Logit-KD & $7.96$          & - & $7.72$          & $1.12$          & $7.27$          & $1.94$          & $7.17$          & $2.08$          & $7.06$          & $1.99$          \\
PNN      & $7.96$          & - & $6.75$          & $\textbf{0.00}$ & $5.53$          & $\textbf{0.00}$ & $5.09$          & $\textbf{0.00}$ & $5.03$          & $\textbf{0.00}$ \\
\ourmodel(ours)     & $\textbf{7.92}$ & - & $\textbf{6.05}$ & $0.26$          & $\textbf{5.21}$ & $1.04$          & $\textbf{4.83}$ & $0.96$          & $\textbf{4.42}$ & $0.68$   \\\hline
\multicolumn{11}{l}{\textit{Smaller memory}} \\
MAS      & $7.96$            & -    & $8.08$          & $5.65$          & $13.44$         & $21.17$         & $13.87$         & $17.67$         & $9.91$          & $9.75$          \\
ER       & $7.96$            & -    & $6.99$          & $2.09$          & $7.15$          & $4.53$          & $6.86$          & $4.09$          & $6.49$          & $3.42$          \\
Logit-KD & $7.96$            & -    & $7.68$          & $1.15$          & $7.24$          & $2.06$          & $7.21$          & $2.27$          & $7.05$          & $2.16$          \\
PNN      & $7.96$            & -    & $6.52$          & $\textbf{0.00}$ & $5.29$          & $\textbf{0.00}$ & $4.84$          & $\textbf{0.00}$ & $4.76$          & $\textbf{0.00}$ \\
\ourmodel(ours)     & $\textbf{7.92}$   & -    & $\textbf{5.85}$ & $0.39$          & $\textbf{5.04}$ & $1.13$          & $\textbf{4.58}$ & $0.98$          & $\textbf{4.20}$ & $0.70$                 \\ \hline
\multicolumn{11}{l}{\textit{Full train wall time \& memory (the main results in \cref{sec:main_exp})}} \\
\ourmodel(ours)      & $7.92$   & -          & $5.62$  & $-0.20$ & $4.81$  & $0.64$          & $4.41$  & $0.64$          & $4.06$  & $0.44$          \\
\bottomrule       
\end{tabu}
\caption{Average perplexity (AP) and average increased perplexity ($\text{AP}^+$) of PLMs trained by different lifelong learning methods with half train wall time on Ns, Rev, Bio, CS domains and smaller memory containing $34$M tokens for each domain.  We evaluate the performance each time when PLMs finish training on one domain.}
\label{table:few_time_memory}
\end{table*}

\begin{table}[!t]
\centering
\small
\begin{tabu}{l@{~~~~}c@{~~~~}c@{~~~~}c@{~~~~}c@{~~~~}c@{~~~~}c}
\toprule
\textbf{Domain}        & \textsc{WB}        & \textsc{Ns}      & \textsc{Rev}    & \textsc{Bio}       & \textsc{CS}        & \textsc{AVG}       \\ \hline
\multicolumn{7}{l}{\textit{Half train wall time}} \\
MAS      & $76.7$          & $72.3$          & $61.6$          & $77.4$          & $64.3$          & $70.5$          \\
ER       & $78.0$          & $71.0$          & $61.1$          & $77.4$          & $65.8$          & $70.7$          \\
Logit-KD & $77.0$          & $72.6$          & $63.8$          & $76.2$          & $58.4$          & $69.6$          \\
PNN      & $76.0$          & $55.9$          & $62.6$          & $53.1$          & $28.0$          & $55.1$          \\
\ourmodel     & $\textbf{82.0}$ & $\textbf{78.4}$ & $\textbf{68.7}$ & $\textbf{81.7}$ & $\textbf{74.0}$ & $\textbf{77.0}$ \\ \hline
\multicolumn{7}{l}{\textit{Smaller memory}} \\
MAS      & $77.1$                   & $73.7$                     & $60.7$                       & $77.5$                    & $68.2$                   & $71.5$                    \\
ER       & $77.9$                   & $72.0$                     & $61.5$                       & $76.3$                    & $63.6$                   & $70.3$                    \\
Logit-KD & $77.0$                   & $73.1$                     & $63.3$                       & $75.9$                    & $57.4$                   & $69.3$                    \\
PNN      & $76.0$                   & $64.9$                     & $64.2$                       & $55.1$                    & $30.5$                   & $58.1$                    \\
\ourmodel     & $\textbf{82.9}$          & $\textbf{80.5}$            & $\textbf{68.9}$              & $\textbf{82.6}$           & $\textbf{74.2}$          & $\textbf{77.8}$   \\ \hline
\multicolumn{7}{l}{\textit{Full train wall time \& memory (the main results in \cref{sec:main_exp})}} \\
\ourmodel     & $83.2$ & $81.8$ & $68.5$ & $82.9$ & $72.7$ & $77.8$  \\
\bottomrule     
\end{tabu}
\caption{Final downstream performance (F1) of BERT on each domain after finishing pre-training on all domains with half train wall time on Ns, Rev, Bio, CS domains and smaller memory containing $34$M tokens for each domain. Experiments of \textsc{Ns} domain are repeated for $10$ times with different seeds and others are repeated for $5$ times. }
\label{table:downstream_few_time_memory}
\end{table}

\begin{table*}[!t]
  \centering
  \small
    \begin{tabu}{cccccccc}
    \toprule 
            \multicolumn{1}{c}{$\textbf{Model}$}
          & \multicolumn{1}{c}{$n_{\text{params}}$} & \multicolumn{1}{c}{$n_{\text{layers}}$} & \multicolumn{1}{c}{$d_{\text{model}}$} & \multicolumn{1}{c}{$n_{\text{heads}}$} & \multicolumn{1}{c}{$d_{\text{FFN}}$} & \multicolumn{1}{c}{\text{lr}}
          \\
    \hline
    \multicolumn{7}{l}{WE + FRW} \\
    $\mathcal{M}_{1}$ & $30.3$M      & $6$      & $384$      &  $6$     &   $1536$    &  $5.0\times10^{-4}$ \\
    $\mathcal{M}_{2}$ & $53.6$M      & $6$      & $576$      &  $9$     &   $2304$    &  $5.0\times10^{-4}$  \\
    $\mathcal{M}_{3}$ & $82.2$M      & $6$      & $768$      &  $12$     &   $3072$    &  $5.0\times10^{-4}$  \\
    $\mathcal{M}_{4}$ &  $104$M     & $6$   &  $896$ &  $14$ &  $3584$  &  $5.0\times10^{-4}$ \\
    $\mathcal{M}_{5}$ &  $129$M     & $6$   &  $1024$ &  $16$ &  $4096$  &  $5.0\times10^{-4}$  \\
    \hline
    \multicolumn{7}{l}{DE + FRW} \\
    $\mathcal{M}_{1}$ & $30.3$M      & $12$   &  $768$ &  $12$ &  $3072$  &  $5.0\times10^{-4}$ \\
    $\mathcal{M}_{2}$ & $51.6$M      & $18$   &  $768$ &  $12$ &  $3072$  &  $2.5\times10^{-4}$ \\
    $\mathcal{M}_{3}$ & $83.6$M      & $36$   &  $768$ &  $12$ &  $3072$  &  $2.5\times10^{-4}$ \\
    $\mathcal{M}_{4}$ & $105$M      & $48$   &  $768$ &  $12$ &  $3072$  &  $2.5\times10^{-4}$ \\
    $\mathcal{M}_{5}$ &  $126$M      & $60$   &  $768$ &  $12$ &  $3072$  &  $2.5\times10^{-4}$ \\
    \hline
    \multicolumn{7}{l}{WE + DE + FRW} \\
    $\mathcal{M}_{1}$ & $30.3$M      & $6$      & $384$      &  $6$     &   $1536$    &  $5.0\times10^{-4}$ \\
    $\mathcal{M}_{2}$ & $51.5$M      & $8$      & $512$      &  $8$     &   $2048$    &  $5.0\times10^{-4}$  \\
    $\mathcal{M}_{3}$ & $82.2$M      & $10$      & $640$      &  $10$     &   $2560$    &  $5.0\times10^{-4}$  \\
    $\mathcal{M}_{4}$ &  $102$M     & $11$   &  $704$ &  $11$ &  $2816$  &  $5.0\times10^{-4}$ \\
    $\mathcal{M}_{5}$ &  $125$M     & $12$   &  $768$ &  $12$ &  $3072$  &  $5.0\times10^{-4}$  \\
    \bottomrule
    \end{tabu}%
  \caption{Model architectures the investigated PLMs of WE+FRW, DE+FRW, WE+DE+FRW. We keep the total number of $\mathcal{M}_i$’s increased parameters for the above three strategies almost the same at each stage $i$.}
  \label{tab:arch_WE_DE}%
\end{table*}%

\section{Representational Similarity of a Stream of PLMs}
We investigate the representational similarity~\citep{abnar2019blackbox} of a descendant PLM and its ancestors. Representational similarity measures how similar two PLMs represent the data. Specifically, we experiment on a stream of PLMs when growing $\text{BERT}_\text{L6\_D384}$ to $\text{BERT}_\text{L12\_D768}$. For a model $\mathcal{M}_j$ and its ancestor $\mathcal{M}_i$ ($1 \le i \le j-1$), we randomly sample $n$ \texttt{[MASK]} tokens from the raw corpus $\mathcal{D}_j$, and get the probability distributions $\bm{p}_k^i$ and $\bm{p}_k^j$ output by the LM head of $\mathcal{M}_i$ and $\mathcal{M}_j$, respectively for each \texttt{[MASK]} token $k$, where $ 1 \le k \le n$. We calculate the average representational similarity (ARS) between $\mathcal{M}_j$ and all its ancestors $\{\mathcal{M}_1, \cdots, \mathcal{M}_{j-1}\}$ as follows:
\begin{equation}
\small
    \begin{aligned}
    \text{ARS}_{j} &= \frac{-1}{(j-1) \times n}\sum_{i=1}^{j-1}\sum_{k=1}^n \text{KL}(\bm{p}_k^i, \bm{p}_k^j),
\end{aligned}
\end{equation}
where KL denotes the Kullback-Leibler divergence between two probability distributions. Higher $\text{ARS}_{j}$ means the representations of $\mathcal{M}_j$ and its ancestors are more similar. To some extent, $\text{ARS}_{j}$ could reflect how much knowledge / functionality of the ancestors is preserved by $\mathcal{M}_j$.

We compare ARS of PLMs trained by Naive, MAS, ER, Logit-KD and \ourmodel and illustrate the results in Figure~\ref{fig:similarity}, from which we observe that Logit-KD has the highest ARS. This is because the training objective of knowledge distillation in Logit-KD is highly correlated with ARS. In addition, \ourmodel takes second place. We also find that, with PLMs continually absorbing new knowledge, the ASR generally decreases.

% Table \ref{table:few_time_memory} and Table \ref{table:downstream_few_time_memory} summarize the pre-training and downstream fine-tuning results and 

\section{Model Architectures for the Analysis of Model Expansion}
\label{sec:model_arch_model_expansion}
In Table~\ref{tab:arch_WE_DE}, we list the model architectures of all the investigated PLMs when conducting the analysis of model expansion in \cref{sec:analysis}. Specifically, three strategies are investigated, including WE+FRW, DE+FRW and WE+DE+FRW. As mentioned in our main paper, for a fair comparison, we keep the total number of $\mathcal{M}_i$'s increased parameters for the above three strategies almost the same at each stage $i$.

\section{Performance of \ourmodel with Fewer Computational Budgets and Storage Budgets}
\label{sec:budgets}
To investigate the performance of \ourmodel under limited (1) computational budgets and (2) storage budgets, in this section, we take an initial step to investigate the effect of (1) training resources (train wall time) and (2) memory size for \ourmodel. Following the experimental setting in \cref{sec:main_exp}, we continually grow $\text{BERT}_\text{L6\_D384}$ to $\text{BERT}_\text{L12\_D768}$ on a stream of data from $5$ domains. We test the performance of \ourmodel and a series of lifelong learning baselines (MAS, ER, Logit-KD and PNN), by (1) reducing the train wall time by half (for \textsc{Ns}, \textsc{Rev}, \textsc{Bio} and \textsc{CS} domain) and (2) randomly sample only $34$M tokens ($1$\% of the full corpus) as the memory $\mathcal{D}_i^{sub}$ for each corpus $i$, compared with the memory size $200$M in \cref{sec:main_exp}. 

\begin{figure*}[!t]
    \centering
    \includegraphics[width=0.45\textwidth]{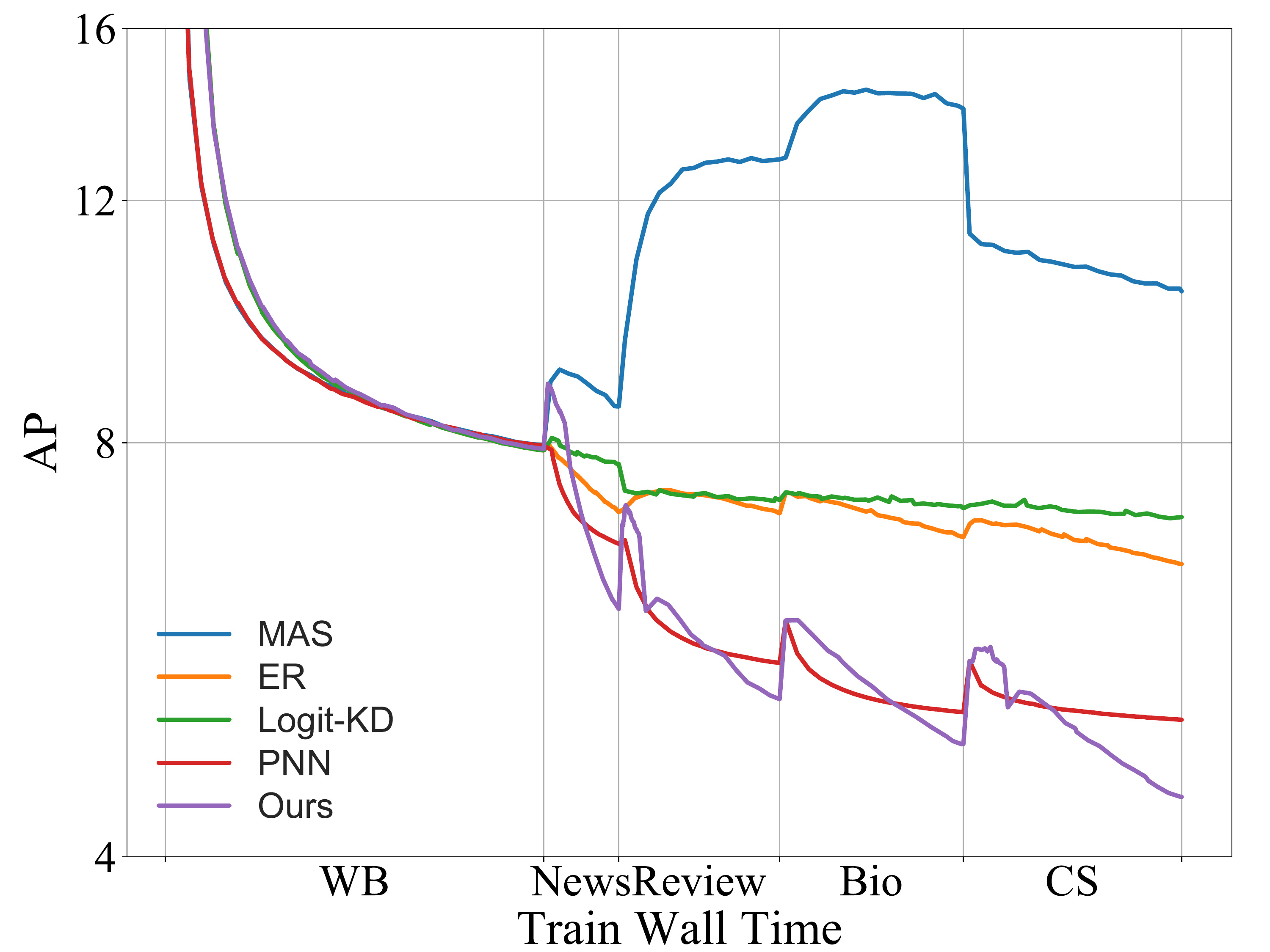}
    \includegraphics[width=0.45\textwidth]{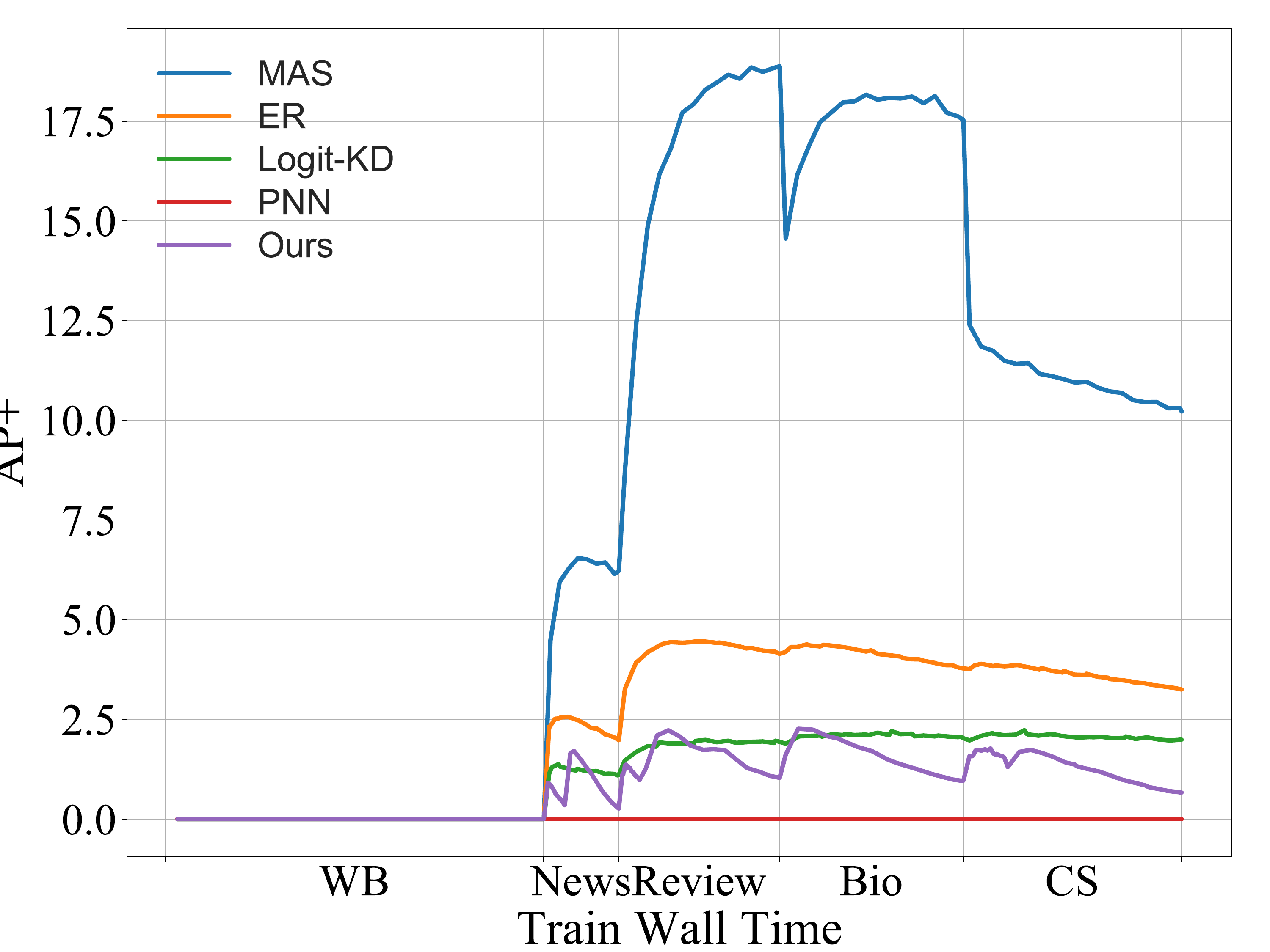}
    \caption{AP and $\text{AP}^+$ of different lifelong learning methods with $\text{BERT}_\text{L6\_D384}$ as the initial model w.r.t train wall time. The train wall time on News, Review, Bio, CS domains is half of the original experiment in Section \ref{sec:main_exp}. \ourmodel continually grows $\text{BERT}_\text{L6\_D384}$ to $\text{BERT}_\text{L12\_D768}$.}
    \label{fig:few_time}
\end{figure*}

\begin{figure*}[!t]
    \centering
    \includegraphics[width=0.45\textwidth]{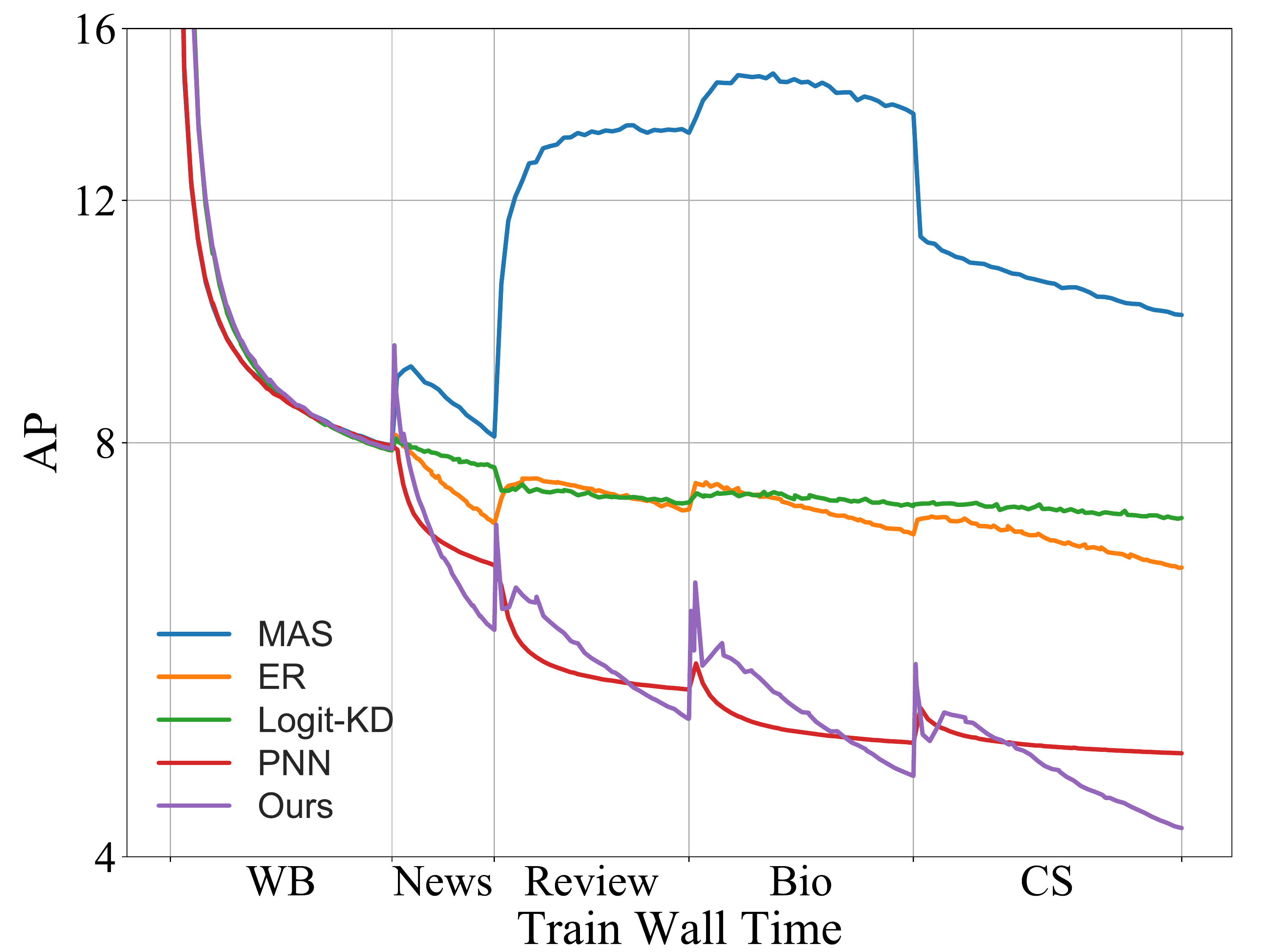}
    \includegraphics[width=0.45\textwidth]{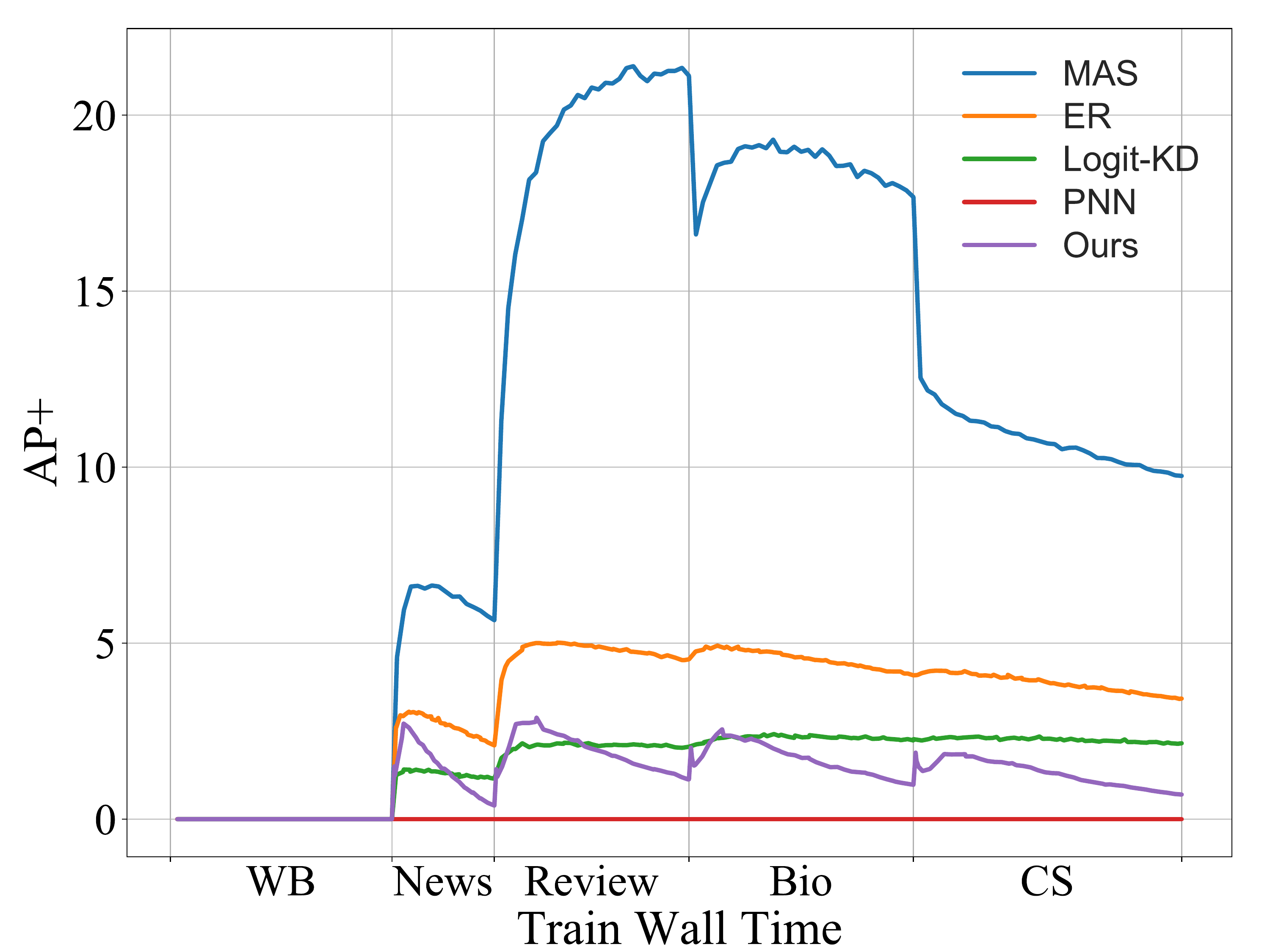}
    \caption{AP and $\text{AP}^+$ of different lifelong learning methods with $\text{BERT}_\text{L6\_D384}$ as the initial model with smaller memory  w.r.t train wall time. For domain $i$, we randomly sample only about $34$M tokens as memory $\mathcal{D}_i^{sub}$, which is $1\%$ of training corpus $\mathcal{D}_i$ . \ourmodel continually grows $\text{BERT}_\text{L6\_D384}$ to $\text{BERT}_\text{L12\_D768}$.}
    \label{fig:few_memory}
\end{figure*}

The experimental results for the above two settings are listed in Table~\ref{table:few_time_memory} (pre-training) and Table~\ref{table:downstream_few_time_memory} (fine-tuning), respectively. We also illustrate the trend curves of $\text{AP}$ and $\text{AP}^+$ in Figure~\ref{fig:few_time} and Figure~\ref{fig:few_memory}. From the above results, we find that: (1) when given fewer computational budgets and storage budgets, \ourmodel still outperforms all the lifelong learning baselines in both pre-training and downstream performance, which demonstrates the superiority of \ourmodel; (2) for \ourmodel, when PLMs are trained with fewer computational budgets, we observe significant performance drops in both pre-training (higher $\text{AP}$ and $\text{AP}^+$) and downstream tasks (lower average F1). This shows that pre-training with fewer computations would harm PLMs' knowledge acquisition; (3) for \ourmodel, when there are fewer memory budgets, although we also observe slight performance drops in pre-training (higher $\text{AP}$ and $\text{AP}^+$), the performance in downstream tasks is generally not influenced, with the average F1 score keeping almost the same ($77.8$). This shows the \textbf{data-efficiency} of PLMs, i.e., PLMs could easily recall the learned knowledge by reviewing small-scale data conserved in the memory (as few as $1\%$). As mentioned before, considering that for pre-training, the expense of storage (e.g., hard disks) is far cheaper than the computational resources (e.g., GPUs), the storage space problem for memory seldom needs to be considered.

% In the experiment with less train wall time, we only use half train wall time of original experiment in Section \ref{sec:main_exp} on News, Reviews, Bio and CS domains. In the experiment with smaller memory,  for each corpus $\mathcal{D}_i$ that contains about 3400M tokens,  we only randomly sample about 34M tokens as memory $\mathcal{D}_i^{sub}$, which is $1\%$ of $\mathcal{D}_i$. The results show that ELLE still outperforms all the baselines in both pre-training and downstream performance.
% The trend curves of average perplexity (AP) and average increased perplexity ($\text{AP}^+$) w.r.t train wall time are shown in Figure~\ref{fig:few_time} (less train wall time) and . 

\section{Details of Baselines}
\begin{figure*}[!t]
    \centering
    \includegraphics[width=0.45\textwidth]{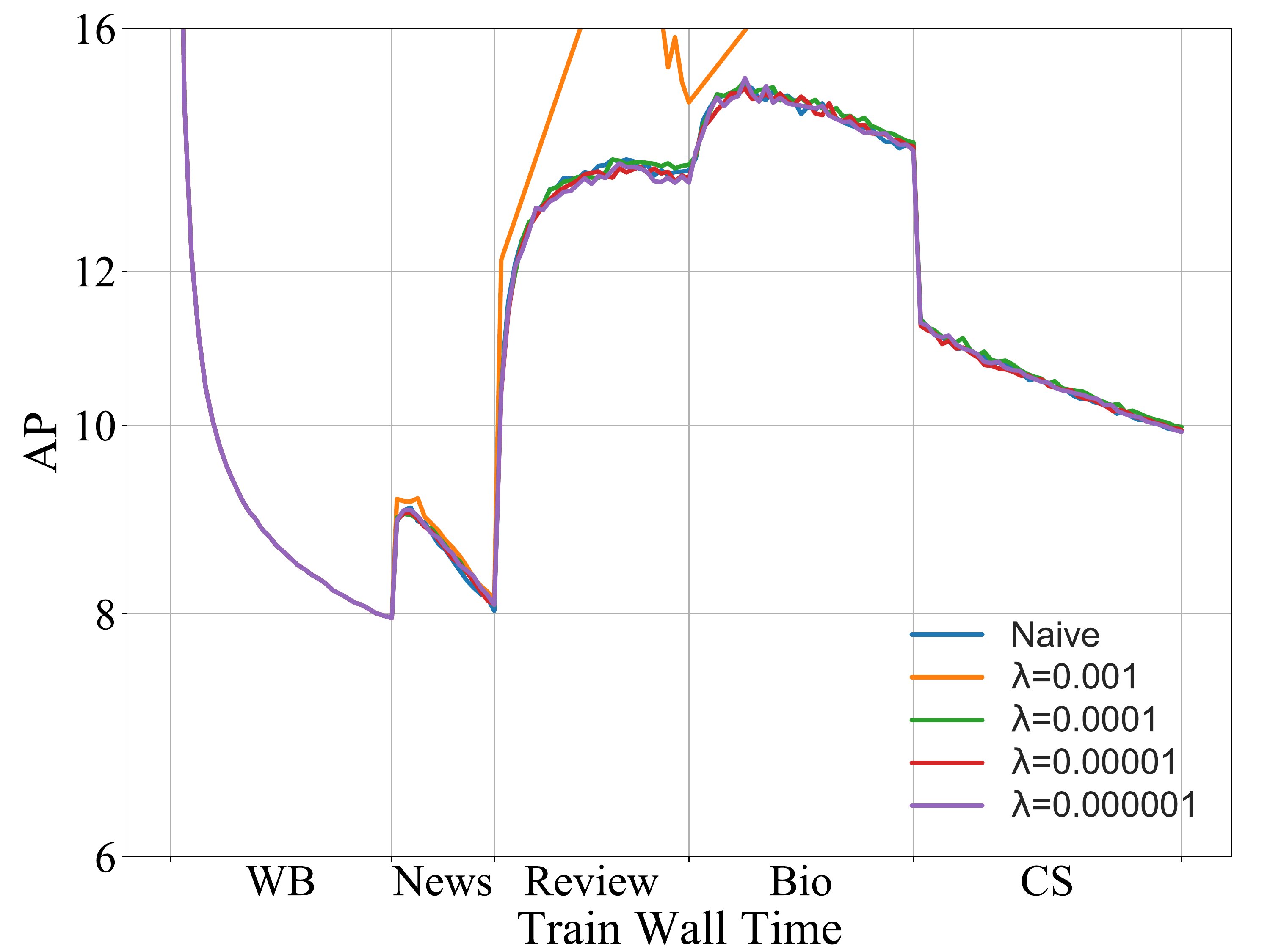}
    \includegraphics[width=0.45\textwidth]{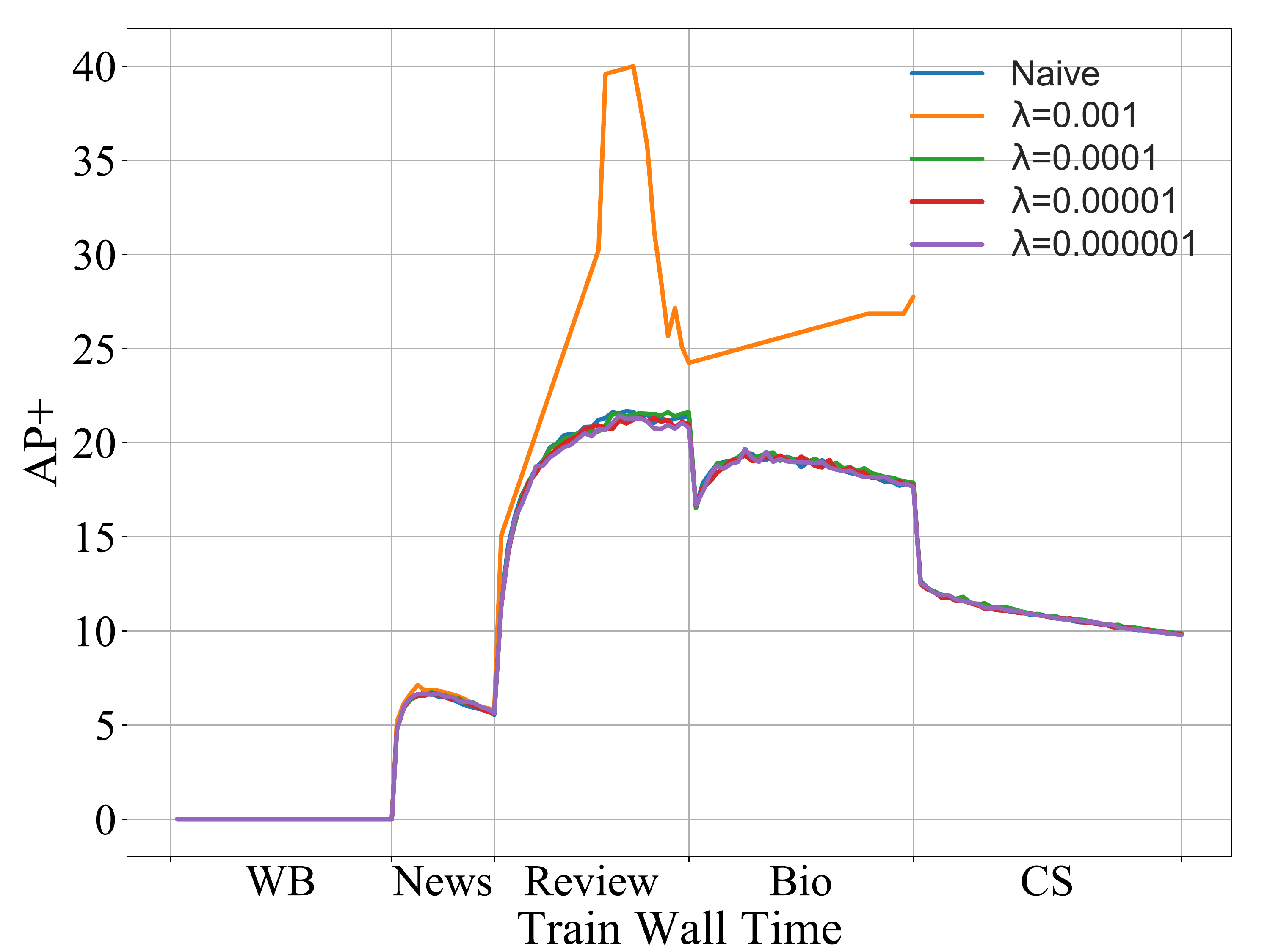}
    \caption{AP and $\text{AP}^+$ of EWC with $\text{BERT}_\text{L6\_D384}$ as the initial model and with different regularization parameter $\lambda$ w.r.t train wall time. }
    \label{fig:hyper_ewc}
\end{figure*}

\begin{figure*}[!t]
    \centering
    \includegraphics[width=0.45\textwidth]{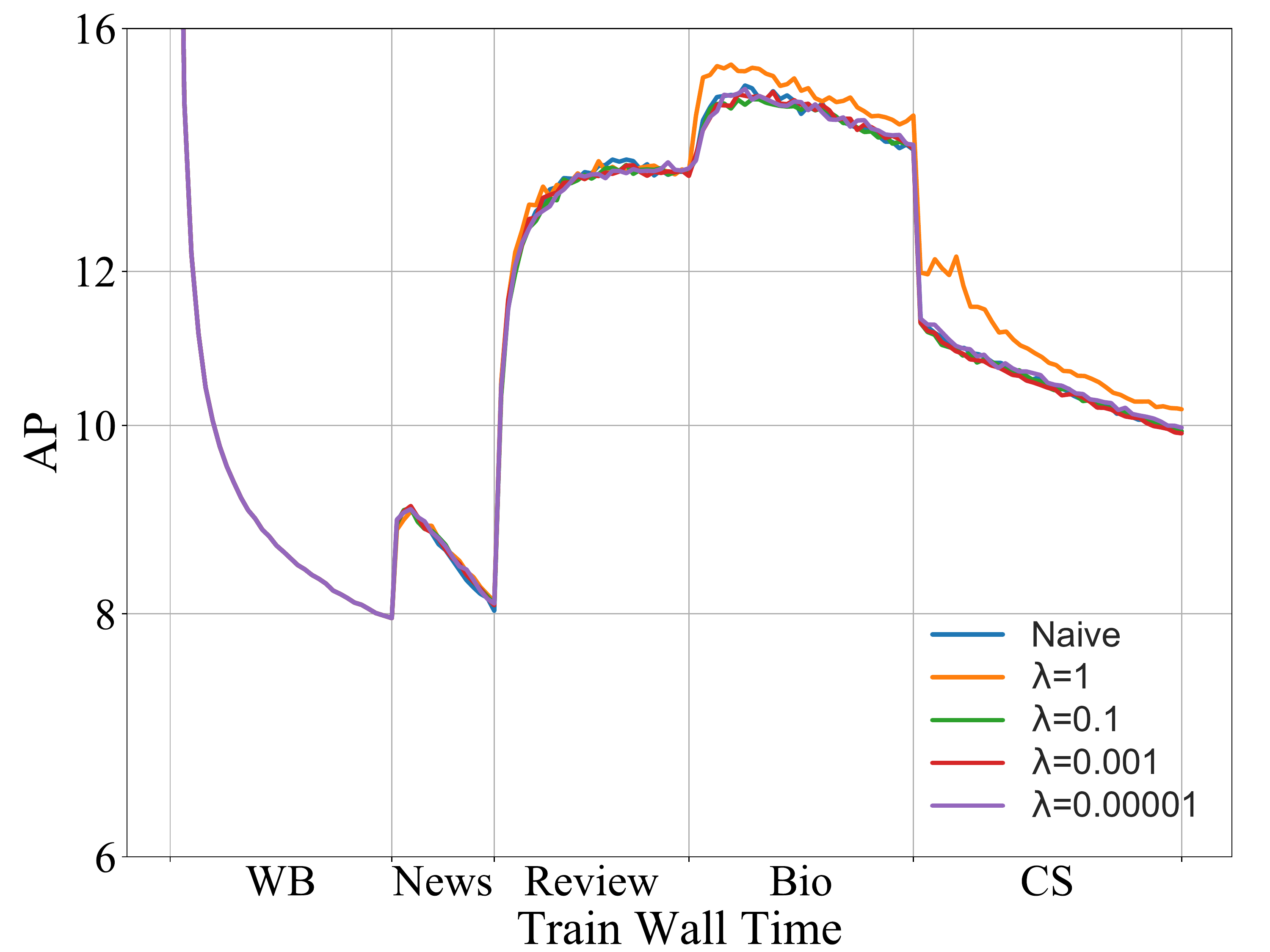}
    \includegraphics[width=0.45\textwidth]{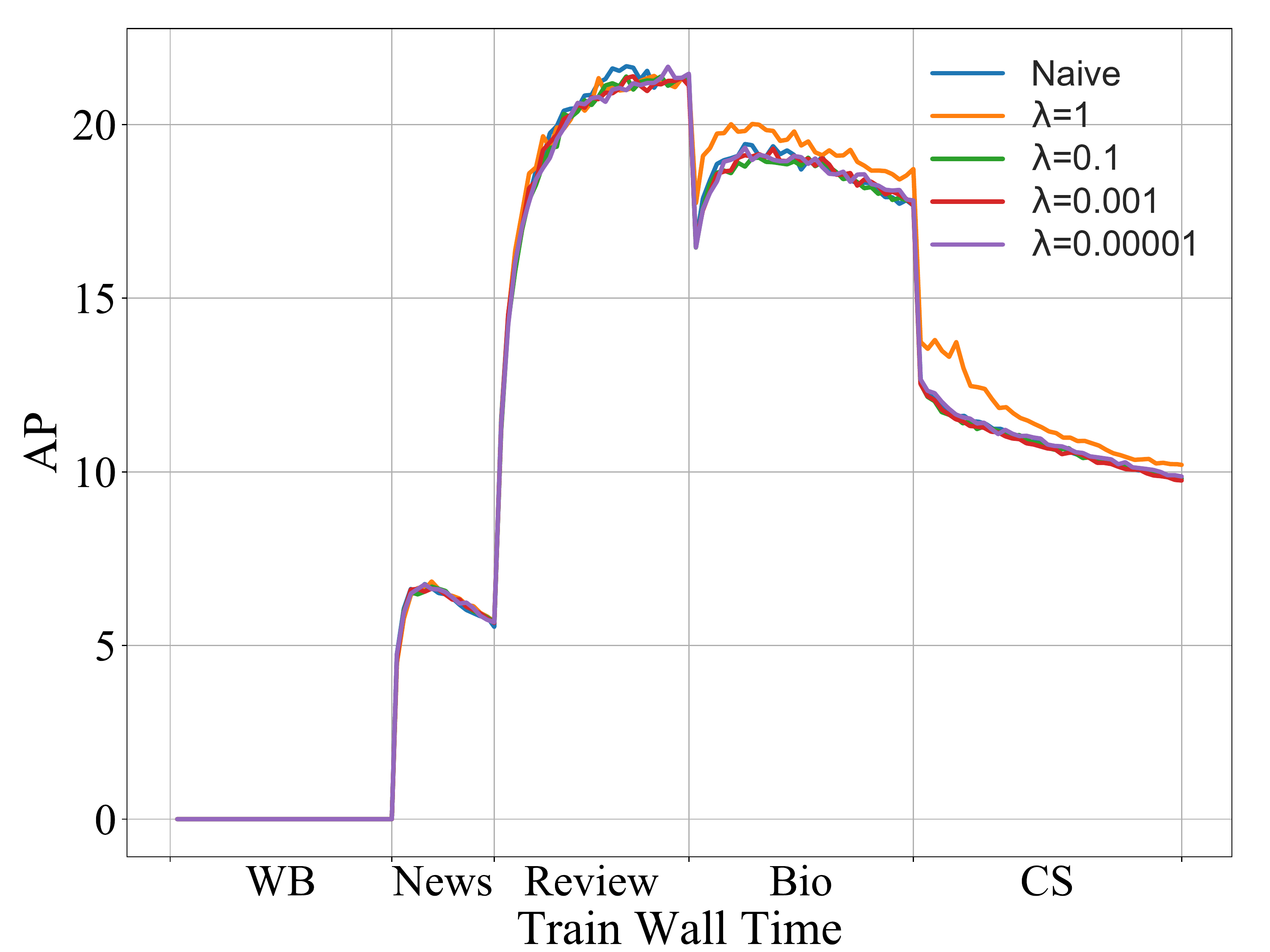}
    \caption{AP and $\text{AP}^+$ of MAS with $\text{BERT}_\text{L6\_D384}$ as the initial model and with different regularization parameter $\lambda$ w.r.t train wall time. }
    \label{fig:hyper_mas}
\end{figure*}

\begin{figure*}[!t]
    \centering
    \includegraphics[width=0.45\textwidth]{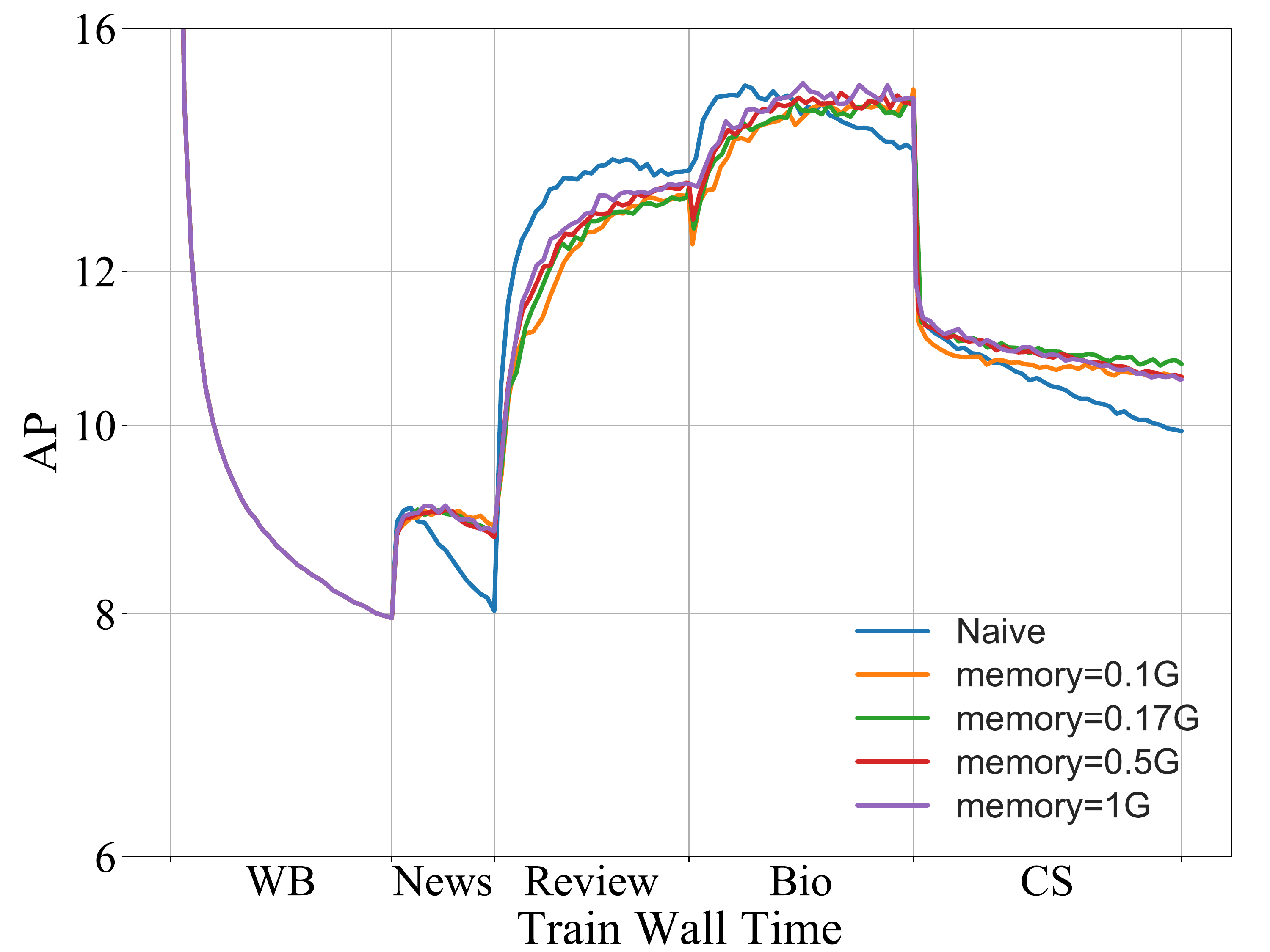}
    \includegraphics[width=0.45\textwidth]{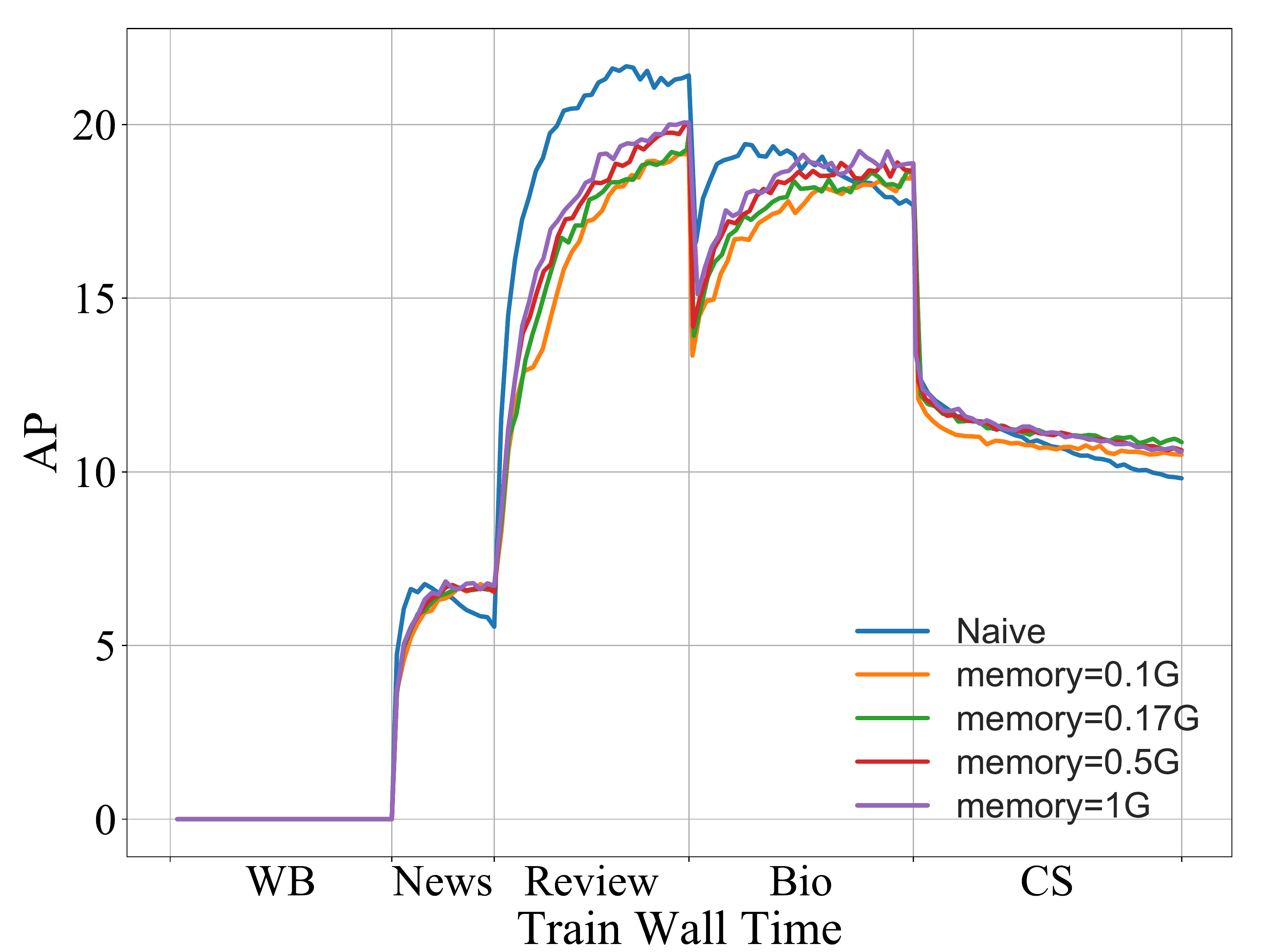}
    \caption{AP and $\text{AP}^+$ of GEM with $\text{BERT}_\text{L6\_D384}$ as the initial model and with different memory size w.r.t train wall time. }
    \label{fig:hyper_gem}
\end{figure*}

We tried different hyper-parameters for baselines, including the regularization parameter $\lambda$ for EWC and MAS, and the memory size for A-GEM, to derive and report their best performance. Their AP and $\text{AP}^+$ curves are shown in Figure \ref{fig:hyper_ewc}, \ref{fig:hyper_mas} and \ref{fig:hyper_gem}. From the results we can see that none of these hyperparameters works well. For EWC and MAS, when the regularization parameter $\lambda$ is small, the pre-training performance is not better than that of naive method. However, if we slightly increase $\lambda$, the performance would become worse than baseline. For A-GEM, the case with bigger memory also doesn't outperform cases with smaller memory and naive case. Specially, we observed that during A-GEM pre-training, $99.9$\% of the inter-products of current gradient and replay gradient are positive, implying that pre-training on different domains is similar to each other to a large extent. This might indicate that EWC, MAS, and A-GEM cannot deal with the subtle difference of various domains.

\end{document}